\documentclass[11pt,openright,a4paper]{report}
\usepackage{algorithmic}
\usepackage{algorithm}
\usepackage{multirow}
\usepackage{enumerate}
\usepackage{pdfpages}

\usepackage{hyperref}   
\usepackage{harvard}    
\usepackage{graphicx}   
\usepackage{amsmath}
\usepackage{mathtools}
\usepackage{pdflscape}  
\usepackage{multicol}   
\usepackage{listings}   

\usepackage{calc}
\citationstyle{dcu}
\fussy

\newenvironment{spaced}[1]
  {\begin{minipage}[c]{\textwidth}\vspace{#1}}
  {\end{minipage}}

\newenvironment{centrespaced}[2]
  {\begin{center}\begin{minipage}[c]{#1}\vspace{#2}}
  {\end{minipage}\end{center}}

\newcommand{\declaration}[2]{
  \thispagestyle{empty}
  \begin{spaced}{4em}
    \begin{center}
      \LARGE\textbf{#1}
    \end{center}
  \end{spaced}
  \begin{spaced}{3em}
    \begin{center}
      Submitted by: #2
    \end{center}
  \end{spaced}
  \begin{spaced}{5em}
    \section*{COPYRIGHT}

    Attention is drawn to the fact that copyright of this dissertation rests
    with its author. The Intellectual Property Rights of the products
    produced as part of the project belong to the author unless otherwise specified
    below, in accordance with the University of Bath's policy on intellectual property 
   (see http://www.bath.ac.uk/ordinances/22.pdf).

    This copy of the dissertation has been supplied on condition that anyone
    who consults it is understood to recognise that its copyright rests with its
    author and that no quotation from the dissertation and no information
    derived from it may be published without the prior written consent of
    the author.

    \section*{Declaration}
    This dissertation is submitted to the University of Bath in accordance
    with the requirements of the degree of Bachelor of Science in the
    Department of Computer Science. No portion of the work in this dissertation
    has been submitted in support of an application for any other degree
    or qualification of this or any other university or institution of learning.
    Except where specifically acknowledged, it is the work of the author.
  \end{spaced}

  \begin{spaced}{5em}
    Signed: Eugene Yuta Bann
  \end{spaced}
  }

\newcommand{\consultation}[1]{%
\thispagestyle{empty}
\begin{centrespaced}{0.8\textwidth}{0.4\textheight}
\ifnum #1 = 0
This dissertation may be made available for consultation within the
University Library and may be photocopied or lent to other libraries
for the purposes of consultation.
\else
This dissertation may not be consulted, photocopied or lent to other
libraries without the permission of the author for #1 
\ifnum #1 = 1
year
\else
years
\fi
from the date of submission of the dissertation.
\fi
\vspace{4em}

Signed: Eugene Yuta Bann
\end{centrespaced}
}

\title{Discovering Basic Emotion Sets via\\Semantic Clustering on a Twitter Corpus}
\author{Eugene Yuta Bann\\eugene@aeir.co.uk}
\date{Bachelor of Science in Computer Science with Honours\\The University of Bath\\May 2012}

\begin{document}

\lstset{breaklines,breakatwhitespace,basicstyle=\small}

\setcounter{page}{0}
\pagenumbering{roman}

\maketitle
\newpage

\consultation{0}
\newpage

\declaration{Discovering Basic Emotion Sets via\\Semantic Clustering on a Twitter Corpus}{Eugene Yuta Bann}
\newpage

\abstract

A plethora of words are used to describe the spectrum of human emotions, but how many emotions are there really, and how do they interact? Over the past few decades, several theories of emotion have been proposed, each based around the existence of a set of ‘basic emotions’, and each supported by an extensive variety of research including studies in facial expression, ethology, neurology and physiology. Here we present research based on a theory that people transmit their understanding of emotions through the language they use surrounding emotion keywords. Using a labelled corpus of over 21,000 tweets, six of the basic emotion sets proposed in existing literature were analysed using Latent Semantic Clustering (LSC), evaluating the \textit{distinctiveness} of the semantic meaning attached to the emotional label. We hypothesise that the more distinct the language is used to express a certain emotion, then the more distinct the perception (including proprioception) of that emotion is, and thus more ‘basic’. This allows us to select the dimensions best representing the entire spectrum of emotion. We find that Ekman's set, arguably the most frequently used for classifying emotions, is in fact the most semantically distinct overall. Next, taking all analysed (that is, previously proposed) emotion terms into account, we determine the optimal \textit{semantically irreducible} basic emotion set using an iterative LSC algorithm. Our newly-derived set (\textsc{Accepting, Ashamed, Contempt, Interested, Joyful, Pleased, Sleepy, Stressed}) generates a 6.1\% increase in distinctiveness over Ekman's set (\textsc{Angry, Disgusted, Joyful, Sad, Scared}). We also demonstrate how using LSC data can help visualise emotions. We introduce the concept of an \textit{Emotion Profile} and briefly analyse compound emotions both visually and mathematically.

\newpage

\tableofcontents
\newpage
\listoffigures
\newpage
\listoftables
\newpage
\listofalgorithms
\newpage

\chapter*{Acknowledgements}
I would like to thank Mr Alan Hayes and Professor James Davenport for their commitment during my studies. It has certainly been a roller-coaster ride! Also thanks to Dr Radim {\v R}eh{\r u}{\v r}ek for his helpful comments on using Gensim, and Professor Philip Johnson-Laird, Professor Keith Oatley and Professor James Russell for their encouraging comments on the project. I am especially grateful to Miss Stela Pashankova for her insightful discussions, Miss Vici Williamson for her rigorous proof reading and advice, and Miss Sarah Beckett for planting the seed to start this research. An extra special thanks to Dr Joanna Bryson for her help, hard work and dedication to this project --- her extensive and diverse experience in technical writing, coupled with her interests aligned with this project has proved invaluable.

\newpage

\setcounter{page}{1}
\pagenumbering{arabic}


\chapter{Introduction}

There are a plethora of words to describe the spectrum of human emotion. Many theories are based on the existence of a set of `basic emotions' that are seemingly hardwired into our brain as individual neurological circuits \cite{watson,izard,plutchik,panksepp,gray}, and that all other emotions are derived from these `biological primitives' as either a combination or specific valence of these neural circuits \cite{basic}. Recently, however, the notion that emotion is a conceptualised act has been proposed \cite{conceptual-emo}, and experimental results have been shown to support this hypothesis \cite{conceptual-act}. Emotion in this sense can be regarded in the same way as colour, insofar we categorise and communicate discrete colours within the confines of language, even though colour itself is in fact a spectrum of visible light. Thus, discerning the psychological primaries of emotion is as fundamental as discerning the unique hues of red, blue, green and yellow in Hering's Opponent Process Theory of Color \cite{colour}.

There is a long-standing view that language intrinsically shapes how people perceive and categorise their world known as the Linguistic Relativity Hypothesis \cite{lrh}. How one categorises certain concepts can be captured by the language used; findings show that language drives the acquisition of colour categories, consistent with newer evidence showing that emotion language influences the acquisition of emotion concepts \cite{perception}. While linguistic \textit{determinism} is ubiquitously thought to have been disproven, empirical evidence appears to be consistent with the idea that there is indeed linguistic \textit{relativity} in the perception of emotion in others (i.e. language influences thought and behaviour as opposed to determining it). Specifically, it has been shown that people of different cultures divide the affective world into different basic emotion categories such that emotion concepts differ across cultures \cite{cultures}.

\section{Emotion Theories}

In contrast to the basic emotions model, the dimensional model, or \textit{core affect} \cite{russ03} represents the entire emotion spectrum by mapping discrete emotion terms in a multi-dimensional space, typically using the dimensions of \textit{valence} (the polarity of the emotion) and \textit{activation} (the level of engagement) \citeaffixed{russell,wt}{for example}. Core affect, however, assumes that each person's conceptualisation of each emotion is universal. Conceptualisation does not refer to the individual cognitive differences of \textit{what} makes one feel a particular emotion, but instead refers to individual thresholds for labelling specific emotion \textit{qualia} as a certain emotion term, and thus its relative placement in the space.

There are several other alternative models of emotion. \citeasnoun{roseman} specified event appraisals that elicit 16 discrete emotions, dependent on cognition. It has been long thought that emotions follow a cognitive structure, specifically that an emotion is elicited as a result of a valenced action, event or object \citeaffixed{occ,occ-revisit,comp-occ}{see for example}. \citeasnoun{occ} distinguishes between emotional and affective-not-emotional words, i.e. words referring to an affective state that are not explicit emotions, for example, moods (`animosity'), traits (`competitiveness'), sensations (`coldness'), cognitive states (`dazed'), and attitude (`defensive'), building up an ontology, or a cognition-based hierarchical structure, of emotion. Recent work in this area involves Mutual Action Histograms (MAH) \cite{mah}, in which knowing the MAH between a subject and an object in a specific event would allow a system to reasonably estimate the emotion evoked. However, these models do not attempt to understand the \textit{meaning} of individual emotions, and are used according to their theoretical definitions within the literature.

\section{Project Objectives}

The primary objective of this project is to evaluate existing basic emotion sets to find out which contain the most emotions expressed in the most distinct language, testing the hypothesis that the more distinct an emotion is (that is, unlike any other emotion), the more distinct the language is used to express the experience of that emotion. \textit{Semantics} refers to the meaning of an expression; in particular, we consider co-occurring words to measure similarities of meaning. We attempt to show such semantic changes in emotion language from a corpus of explicitly expressed emotions extracted from the micro-blogging website Twitter, and evaluate six basic emotion sets on a scale of \textit{semantic distinctiveness}, based on the theory that the more distinct the language is used to express a certain emotion, then conceptually (i.e. what we understand that emotion keyword to mean), the more psychologically irreducible that emotion is. The less semantically accurate a set of emotions is, the more similar these emotions are to each other, or in other words, if similar words are used when expressing two different emotions, then these emotions are, in theory, conceptually, and thus psychologically, similar. A large majority of computer scientists tend to use Ekman's basic emotion set for emotion categorisation, and it appears that, semantically, it is the most distinct set. The secondary objective of this project is to identify a set of basic emotions by clustering underlying semantic features of each expression within the corpus. We also aim to discover to what extent do the semantics of emotion language vary according to geographical regions, thus providing empirical support for the Linguistic Relativity Hypothesis.


\chapter{The Psychology of Emotion}

Emotion is that which leads the subject's condition to become so transformed that their judgement is affected \cite{aristotle-def}, triggered by a subconscious appraisal process about something that matters to the person experiencing it \cite{ekman-def}. It is characterised by behavioral, expressive, cognitive, and physiological changes \cite{panksepp-def} and can be started and executed unconsciously \cite{damasio-def}. The desire to experience or not experience an emotion largely determines the contents and focus of consciousness throughout the life span \cite{izard-09}.

The above definition of emotion is not in the least a conclusive definition of emotion, but takes the most important aspects from notable theorists' definitions. Attention is drawn to Aristotle's wording, stating that emotion ``is \textit{that} which...", implying that emotion is in fact a type of \textit{quale}, that is, a subjective conscious experience that cannot be communicated, or apprehended by any other means other than direct experience \cite{qualia}. Qualia refers to subjective `raw feels', for example, the taste of red wine, or the experience of seeing the colour red. Emotion qualia thus refers to the raw feel of an emotion; the actual phenomenon of a particular emotion experienced may actually differ according to each person's perception of that emotion, with perception being the result of the individual's past and hypothesised responses, unique to each human being. A useful representation of particular emotion qualia involves the use of \textit{scripts} \cite{cultures}, in which prototypical features of each emotion are described as a list of sub-events, or in other words using an example-based definition, although this too is subject to individualisation based on past experiences.

Emotion could be thought of as a form of internal communication between organisms within an environment. Indeed, \citeasnoun{emo-lang} has proposed the theory that emotions are a reorganisation of the organism-environment system, and that emotion and knowledge are in fact only different aspects of the same process. Emotion as an explicit language is different to communicative languages as it is arguably difficult to explicitly transfer an exact emotion to another person without any language taking place (i.e. remote transfer of feelings), although awareness of emotional contagion allows for it to be explicitly transferred via body language, tone of language, facial expressions and so on. Consequently, in addition to gestures, expressions and the like, people use labels, or emotion keywords, to approximate the combination of qualitative sensations that best describe emotion qualia to communicate their feelings, limited to the complexity of language. Language enables people to communicate with each other \textit{without} the need for emotion, which, although enabling people to explicitly hide their true emotion, has enabled the evolution of cognitive thinking.

\section{Basic Emotions}\label{basic-section}

The dominant theory of emotion postulates the existence of a small set of hardwired, or `basic', emotions, and consequently the majority of textual emotion recognition research has been based on such, with \possessivecite{ekman-face} set: 

\begin{center}
\textsc{anger disgust fear joy sadness surprise}
\end{center}

arguably being the most popular within the field of computer science for emotion mining and classification. However, not only do the emotions comprising each basic emotion set vary amongst theorists, they do not always agree about what emotions \textit{are}, thus adding to the confusion of what exactly are the basic emotions, or whether they exist at all. To illustrate this point, consider the emotion of \textit{surprise}: can it be considered an emotion if it can take the form of a negative, neutral or positive valence? Moreover, consider the emotion \textit{disgust}: the same label is used to describe both the feeling of moral disgust and visceral disgust. This can be viewed as a problem regarding the vagueness of language, suggesting that there is a general problem about how to talk about the objects (emotion qualia) one wishes to study \cite{basic}. Language is the most readily available non-phenomenal access we have to emotions, although it must be noted that a theory of emotion must not be confused with a theory of the language of emotion \cite{occ}. There are two viewpoints concerning the advocation of basic emotions: they are either biologically primitive or psychologically irreducible.

\subsection{Biologically Primitive Emotions}

Biologically primitive emotions have arisen largely from affective research with animals \citeaffixed{pani}{for example,}. Animal researchers have created taxonomies of the basic emotions and proposed specific neural pathways associated with each one, although human studies confirming these findings have proved elusive. It is argued that encephalization of the brain during human evolution has not relocated emotional core processes from subcortical to cortical structures, but has in fact \textit{expanded} the penumbra of emotional core processing circuitry into the upper layers of the brain, thus not replacing the basic emotional brain that we share with other animals \cite{berridge}. In other words, our emotional brains are greater in size, but not structurally different to similar animals. Developmental studies assume a similar empirical perspective to the study of emotion as that taken by animal models of affect insofar the behaviors of nonverbal subjects are observed and interpreted by researchers into specific emotional categories. These affective behaviors, however, are not identical to subjective feelings, or in other words, eliciting an emotional response is not equivalent to experiencing an emotion \cite{circumplex}.

Biologically primitive emotions are also referred to as \textit{universal} emotions, as one would expect to find neurophysiological or anatomical evidence of these hardwired emotions in all members of the species \cite{basic}. The most widely used methodology in this field establishes basic emotions by identifying their associated characteristic facial expressions \cite{darwin-face,ekman-face}. This proposal, however, has been largely discredited due to the overlap of expressive characteristics amongst seemingly basic emotions, resulting in the taxonomy of facial expressions not adequately describing the taxonomy of emotions. Recently, \citeasnoun{context} suggest that facial expressions are perceived relatively to the context they appear in, such as language. Highlighting \possessivecite{context} example, an identical facial expression expressing excitement could be mistaken for an expression of anger when taken out of context. This lack of a universal signature could have consequences for how clinicians are trained and also for the security industry (for example, classification of behaviours that determine who to `spot-check'). Moreover, it could be argued that many people can experience an emotion without the need to express the associated facial expression, so while these people register their feelings, they do not necessarily or identifiably alter their facial expression. Facial expressions can help identify an emotion, but an emotion cannot be classified or explained by facial expressions --- \citeasnoun{robots} calls this a perceptual stance; we shouldn't be concerned with how emotions create facial expressions, but what information an observer can determine from a facial expression.

\subsection{Psychologically Irreducible Emotions}

\begin{figure*}
\centering
\includegraphics[width=400px]{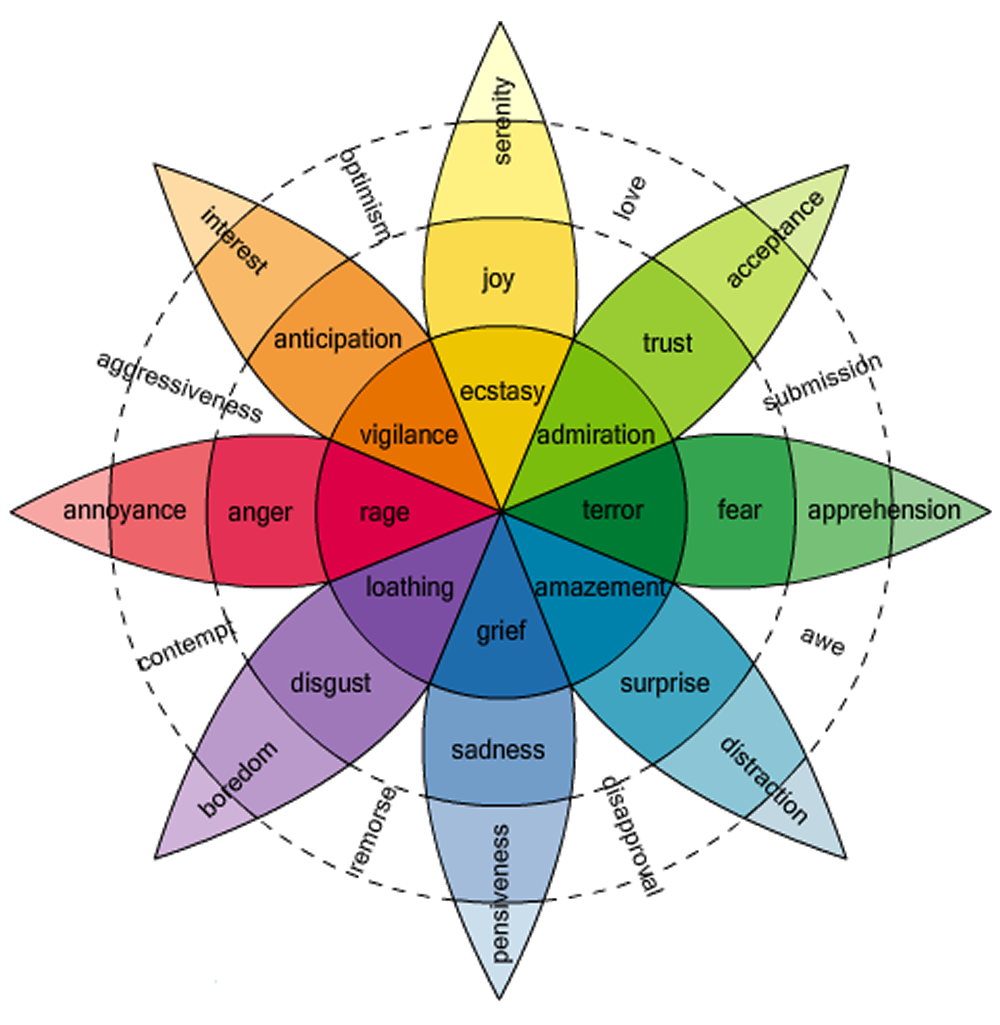}
\caption[Plutchik's Wheel of Emotions]{\possessivecite{plutchik} Wheel of Emotions.}
\label{plutchik}
\end{figure*}

Theories of psychologically irreducible emotions postulate that all other `non-basic' emotions can be created by fusing, blending, mixing or compounding basic emotions, or in other words, basic emotions do not have other emotions as constituents \cite{basic}. The most notable model in this field is \possessivecite{plutchik} Wheel of Emotions (see Figure~\ref{plutchik}) in which eight primary emotions, each with three levels of activation, and eight secondary emotions that are the fusion of their two adjacent primary emotions are defined and mapped. While this is a popular model, no general principles of combination are presented, and no details are offered about the kinds of mechanisms that might be involved in the creation of such combinations \cite{basic}.

\begin{figure*}
\centering
\includegraphics[width=400px]{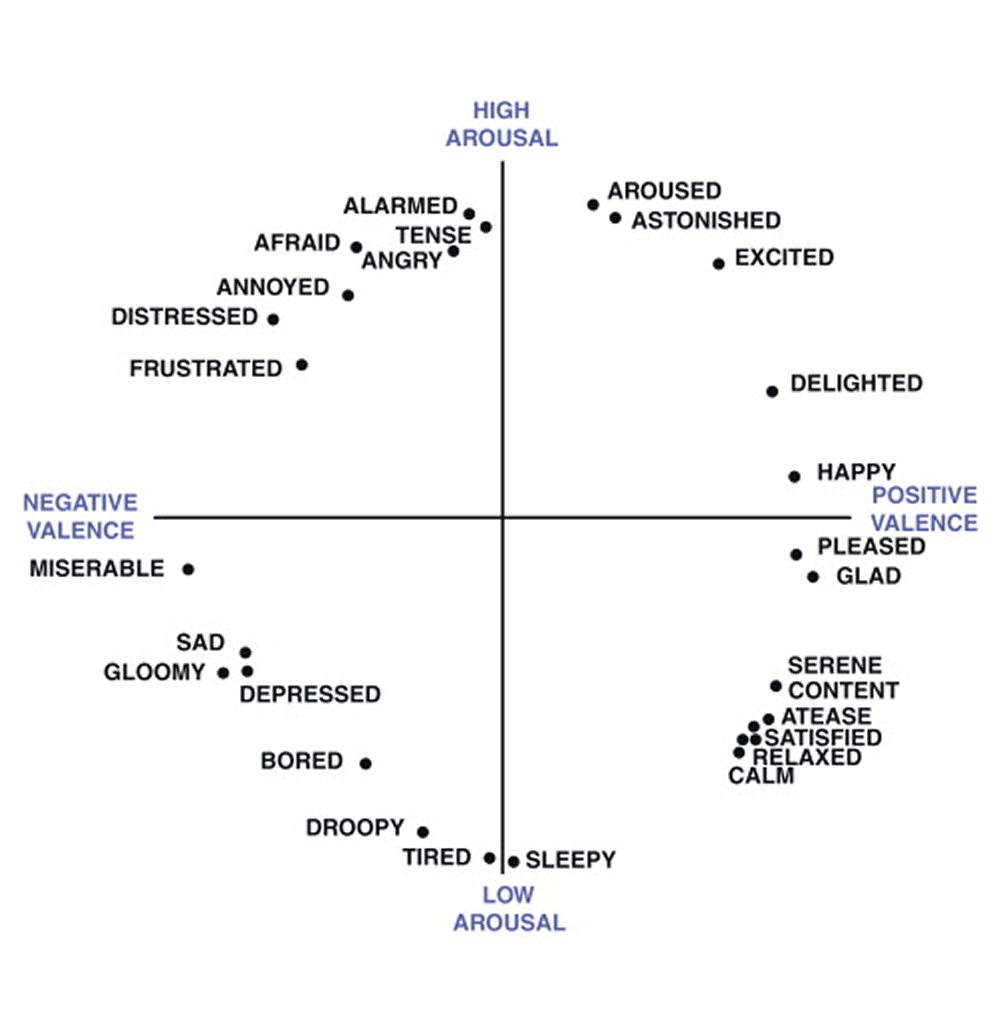}
\caption[Russell's Circumplex Model of Affect]{\possessivecite{russell} Circumplex Model of Affect.}
\label{russell}
\end{figure*}

\begin{figure*}
\centering
\includegraphics[width=400px]{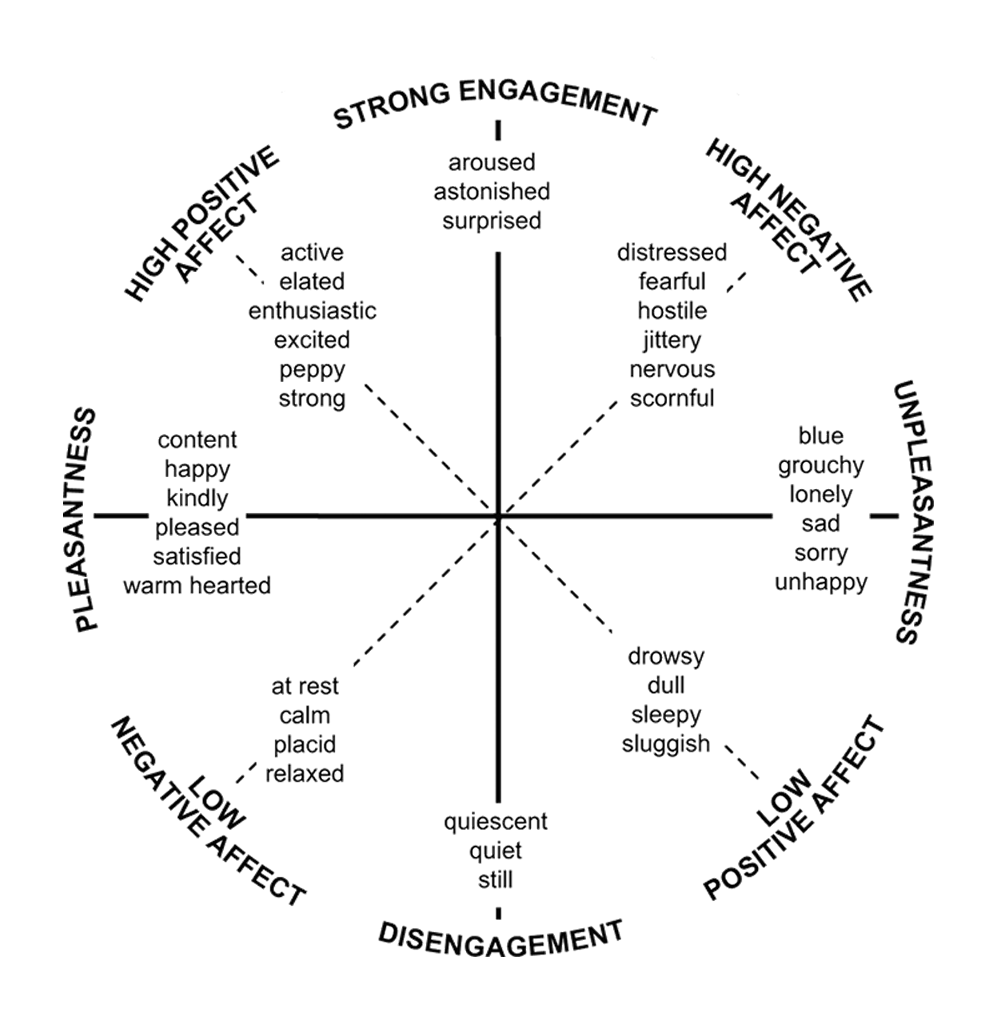}
\caption[Watson and Tellegen's Circumplex Theory of Affect]{\possessivecite{wt} Circumplex Theory of Affect.}
\label{watson}
\end{figure*}

\section{Core Affect}\label{coreaffect}

In contrast to the basic emotions model which treats emotions as discrete, labelled phenomena, core affect --- an emerging paradigm in affective neuroscience --- considers a continuous approach to defining emotions. One of the very first empirical models of core affect is \possessivecite{russell} Circumplex Model of Affect (CMA) (see Figure~\ref{russell}), in which 28 discrete emotions are mapped around a circumplex according to a study that involved each of the emotions being classified as one of eight emotion categories, each constituting an octant of a circumplex. The eight categories were chosen theoretically according to two dimensions --- arousal and pleasure --- and emotions were scaled around the circumplex using Principal Component Analysis. Another popular model of core affect is \possessivecite{wt} Circumplex Theory of Affect (CTA) (see Figure~\ref{watson}), in which 38 emotions are distributed amongst four bipolar dimensions: positive affect, negative affect, pleasantness and engagement. Another is \possessivecite{ESM} Evaluative Space Model, which is similar to CTA but posits that co-activation can occur, that is, that bipolar dimensions are not negatively correlated.

All models of core affect have one thing in common: they represent emotion qualia as a single integral blend of two dimensions, represented as a single point in the space. Core affect is universal, primitive, and irreducible to anything else psychological --- similar to the subjective experience of temperature \cite{russ03}. Core affect can, to some extent, be represented in a three dimensional space of pleasure, arousal, and dominance (PAD), typically using the Self Assessment Manikin (SAM) which asks people to rate an item (for example, text) with respect to pleasure, arousal and dominance using an intuitive interface \citeaffixed{pad-sam}{see for example}.

\possessivecite{russell} work also presents the three properties of the cognitive representation for affect:

\begin{enumerate}[I]
\item The pleasentness-unpleasentness and arousal-sleep dimensions account for the major proportion of varience;
\item The dimensions descriptive of affect are bipolar;
\item Any affect word could be described as some combination of the pleasure and arousal components.
\end{enumerate}

If basic emotions are meant to be as distinct from each other as possible, then we could understand basic emotions to be the emotion keywords used to describe the dimensions of the emotion spectrum. We can modify the properties above to generate the three laws of basic emotion sets:

\begin{enumerate}[I]
\item Both positive and negative emotion subsets must cover a range of arousal levels;
\item Emotions must be able to be paired in such a way that each pair is bipolar;
\item An emotion that can be described as a combination of two or more emotions \textit{within the set} cannot be considered a basic emotion.
\end{enumerate}

Armed with these laws, we are better able to evaluate potential basic emotion sets, although we still face the problem of assuming a universal conceptualisation of emotions.

\section{The Conceptualisation of Emotion}

\possessivecite{conceptual-emo} work studies the act of conceptualising core affect, or in other words, why people attach emotion labels to the experience of emotion qualia. Barrett proposes the hypothesis that emotion is a psychological event constructed from the more basic elements of core affect and conceptual knowledge. In a study focusing on the conceptualisation of \textit{fear}, it was found that neither the presence of accessible emotion concept knowledge nor core affect alone was sufficient to produce the world-focused experience of fear \cite{conceptual-act}. As emotions are constructed from conceptual knowledge about the world, we can see that emotions themselves are in fact concepts that humans begin learning in infancy and continuously extended and revised throughout life \cite{conceptual-act}. This repeated experience of labelling a combination of core affect and the context in which it occurs as an emotion provides ``training" in how to recognise and respond to that emotion --- in this sense, Barrett describes emotions as ``simulations". This ``skill" of conceptualising core affect as an emotion could be a core aspect of emotional intelligence --- in much the same way as conceptual thinking is core to cognitive intelligence  --- defining how humans deal with their internal state, but more importantly, defining the emotion labels used as a combination of specific experiences. Each person's conceptualisation of their emotion spectrum is thus unique; it is this conceptualisation that we attempt to aggregate and analyse in this research.

\section{The Negativity Bias}

By mapping discrete emotion terms around a circumplex, \citeasnoun{russell} shows that basic emotion sets are in fact a skewed representation of the entire emotion spectrum, in favour of negatively-valanced emotions. This implicitly supports the \textit{negativity bias} in psychology \cite{neg-bias}: the tendency to use negative emotions much more than positive emotions. Particularly, it has been indicated that negative events elicit a more rapid and more prominent response than positive events \cite{neg-events}. Recently, it has been shown that expressed negative emotions ``boosts activity" on BBC forums \cite{bbc}, suggesting that perhaps people's individual emotion circumplexes \textit{are} negatively skewed. It is thought that experiencing negative emotion serves as a call for mental or behavioral adjustment \cite{neg-bias}, and as a result is more likely to be communicated as a way of searching for a solution. In contrast, positive emotion serves as a cue to stay on course or as a cue to explore the environment \cite{neg-bias} and consequently is communicated less. The communication of emotion does appear to be most beneficial when expressing negative emotions. However, the negativity bias will have a negligible effect on this study due to our focus towards the words surrounding a \textit{specific} emotion keyword. The English language as a whole has in fact a positive bias, with a recent study showing that 72\%, 79\% and 78\% of words used in Twitter (which we form our corpus from), Google Books and New York Times articles respectively are positive words \cite{positive}.

\section{Discussion}

There exist many models that have been devised over the years that aim to explain why humans experience emotions. Emotions play a key part in human consciousness, influencing the thoughts and actions of every human being. Antonio Damasio points out that ``people who lack emotions because of brain injuries often have difficulty making decisions at all"\footnote{\texttt{http://www.usatoday.com/tech/science/discoveries/2006-08-06-brain-study\_x.htm}}; indeed, it could be said that the combination of emotional and cognitive information underpins ``rational" behaviour \cite{underpin}. Rational decisions are dependent on the amount of knowledge that is available, altering perceptions that could significantly change the decision process. However, the actual process of making a decision with a given amount of knowledge is not unemotional: making a completely rational decision is ultimately a positive emotion determining the course taken, be it, for example, \textit{interest}, \textit{assurance}, \textit{certainty}, the absence of negative emotions or the `lesser of two evils' (a `more positive' negative emotion).

Basic emotion theorists suppose that a small set of basic emotions are in fact inherently within all humans regardless of individualisation. A theory of basic emotions is a discrete theory in which each basic emotion maps to one neural system. Taken individually, biologically primitive emotion models have a strong grounding for inclusion of constituent emotion terms. However, there is no consensus on the constituents of universal basic sets. One way to define a basic emotion set could be to populate it with labels of emotion that are present within an array of languages spanning several cultures, with the premise that the emotions omitted are specific to culture, and thus could be `created' from other, more basic emotions. Psychologically irreducible models of emotion describe the interplay of basic and non-basic emotions, but although making psychological sense, no empirical evidence has been provided to confirm the composition of these models. 

The act of conceptualising emotion treats emotions as a combination of core affect and conceptual knowledge, built using the same mechanisms as in cognitive learning. Thus, we can only understand and label emotions that we are aware of, although when we experience new emotions in varying contexts, we update our emotion knowledge by extrapolating to our past experience of a similar emotion. With language, we have conceptualised our view of emotion by the use of discrete emotion labels that represent easy to understand definitions of emotion qualia. This enables humans to talk about emotions in order to share and understand them to the extent of linguistic limitations imposed by the language used.

Emotions can be expressed in a variety of ways including facial expressions, body language, tone of voice, and the language used in speech and text. This project focuses on the most explicit of these examples --- the language used in communication --- with the proposition that how humans communicate to one another can reveal individual conceptualisations of specific emotions, given that the specific emotion keyword is used within the communication. Defining \textit{basic emotions} as emotions that are conceptually distinct from any other emotion, we explore the hypothesis that the language used in communicating basic emotions should be significantly different for each one, as each basic emotion should describe a sufficiently distinct concept. In order to test this hypothesis, we must first collect a sample of `real' emotion data --- expressions of basic emotions --- by conducting Lexical Emotion Extraction on the Internet.


\chapter{Lexical Emotion Extraction}\label{emo-extract}

Plato, Aristotle, Darwin, James, and even Ekman did not have access to the wealth of emotional data being constantly published all over the Internet, in particular within social networking sites. We tap into this emotionally rich source of information to create a relatively large corpus of emotional experiences to perform our analysis on. As we will discuss at greater depth in section~\ref{sec-moredata}, an increase in data quantity leads to an increase in the accuracy of results so, in theory, a corpus of over 21,000 tweets should produce a much more accurate overall result compared to experimental studies using a sample of, say 50 people.

Lexical Emotion Extraction, or \textit{Emotion Mining}, is a subfield of data mining that aims to extract emotions from text using an array of techniques. There are numerous systems that automatically or semi-automatically extract emotions from text for analysis, enabling researchers to track how people are feeling with respect to a variety of indicators. Emotional indicators have recently been demonstrated to perform more accurately than some existing market-based predictors \cite{stock}, suggesting a strong indication that emotion greatly influences the decisions people will make in the future, whether as a consequence of self-experience or the experiences of others. Many corpora have been analysed including customer reviews \cite{reviews}, news headlines \cite{semeval,headlines}, speech samples \cite{basic-ext-2}, literature \cite{alm} and emails \cite{email-buddy}. Online data is increasingly being used; indeed, people's emotional and cognitive mechanisms are being revealed through what they type online on social networking sites. One such social networking site that has received notable attention in the domain of emotion mining is Twitter.

\section{Twitter}\label{twitter-section}

Twitter is a public micro-blogging system that allows users to share short messages of up to 140 characters. Its user base has grown exponentially since its conception in 2006 and currently has around about 100 million active users\footnote{\texttt{http://blog.twitter.com/2011/09/one-hundred-million-voices.html}} with 200 million micro-blogs, or tweets, posted each day\footnote{\texttt{http://blog.twitter.com/2011/06/200-million-tweets-per-day.html}}. Tweets can be posted `at home' via a computer or `on the go' via a mobile device, and, as they are publicly available, provides us with an ethical way of collecting a diverse range of public expressions. Coupled with the fact that a good proportion of tweets project the user's emotion, we are able to assume that Twitter is a valid sample of human emotive expression and thus a suitable corpus for this project. \citeasnoun{matthis} sees emotions as \textit{preconscious} \cite{precon}; due to their limited character size of 140 characters, tweets could be seen as a form of, albeit weak, automatic appraisal, insofar a large majority of users take a considerably less amount time of time or thought in publishing a tweet compared to forum posts, descriptions of in-depth experiences and emails. Although we cannot assume that Twitter is a measure of preconscious expressions, there is somewhat of an explicit impulse to communicate emotions on Twitter and although the underlying cause is not always explicitly mentioned, it is this factor that we attempt to capture. For the purpose of this project, we neglect the suggestion that there may be some bias in expressions due to public image considerations, the widely fluctuating level of which should be further investigated (see Chapter~\ref{further-research}). Although a few systems have mined emotion from Twitter (see section~\ref{sec-emo-recog}) the majority of systems have focused on sentiment analysis.

\section{Sentiment Analysis}

Twitter has been used extensively to measure and classify sentiment \citeaffixed{sentiment-analysis,twitter2,twitter3,twitter1}{e.g.}. Sentiment, at least in the computer science sense, refers to the opinions of individuals, specifically regarding the emotional polarity of a document, labelling it either positive, negative or neutral. It is essentially a uni-dimensional, valence-focused variation of the core affect model.

The majority of sentiment mining from the web is based on the extraction of emoticons (for example :-) for happy and :-( for sad) and then training a classifier from these results \citeaffixed{sentiment-analysis,twitter2,twitter3}{see for example}. Using this technique assumes that (a) emoticons are an indication of an elicited emotion, (b) the polarity of the emotion is defined by the type of emoticon, and (c) that the emotional polarity applies to all words within the document. Traditionally, machine learning techniques only work well when there is a good match between training and test data with respect to topic, however, emoticon labels have the potential of being independent of domain, topic and time \cite{read}. Emoticons are also a more intuitive representation of an elicited emotion, being pictorial rather than lexical, although they are not very specific and thus can only effectively be applied to sentiment analysis, or to augment emotion extraction. Other methods to extract sentiment have been used, including using Amazon's Mechanical Turk to manually label a Twitter corpus \cite{pred-twitter} before training classifiers. \citeasnoun{twitter1}, to discern target-dependent features, used a combination of emoticons, related tweets and \textsc{General Inquirer} to build up a more complex and accurate sentiment score for a tweet. Emotion lexicons, such as \textsc{SentiWordNet} have also been used to generate sentiment scores based on individual emotional scores of each word \cite{twitter4}.

\textsc{LingPipe} (Java) and Natural Language Toolkit (NLTK) (Python) integrate machine learning techniques which can be used to train a sentiment classifier. In terms of the performance of classifiers, \citeasnoun{read} has shown that neither Support Vector Machines (SVM) or Na\"{i}ve Bayes classifiers outperforms the other. A Na\"{i}ve Bayes classifier is less complicated to set up and works best for unigrams (single keywords) although when taking both unigrams and bigrams (two-word phrases) into account, a Maximum Entropy classifier has been found to produce slightly better results \cite{sentiment-analysis}.

There are many online tools available that aim to measure sentiment by incorporating these various methods of classification, including \textsc{Lexalytics}, \textsc{Tweetfeel}, \textsc{Twitter Sentiment} \cite{sentiment-analysis}, \textsc{twitrratr}, Linguistic Inquiry and Word Count (LIWC), \textsc{OpinionFinder} and \textsc{General Inquirer}. However, these systems only measure textual sentiment and do not provide a detailed insight into the emotional nature of each context. Emotions are much more specific than sentiment, so while emotions can infer sentiment based on the theoretical polarity of emotional keywords, it is harder to classify due to increased constraints, such as engagement and dominance. As a result, different techniques are required to extract emotion, rather than sentiment, from text.

\section{Emotion Recognition}\label{sec-emo-recog}

The past five to ten years has seen a rapid increase in emotion-recognition research. The most common method is identifying text as one of the basic emotions, treating basic emotions as emotion \textit{types} \citeaffixed{basic-ext-2,emo-cause2}{see for example}. The predominant technique is to use some sort of ontology or dataset to seed each basic emotion type with synonyms and related words to produce a hierarchical emotion lexicon, and then to categorise text using these classifications. Several ontologies and lexicons have been used including \textsc{WordNet} \cite{basic-ext-1}, \textsc{WordNet-Affect} \cite{headlines}, \textsc{SentiWordNet} \citeaffixed{semeval}{the \textsc{UPAR7} system in} and the Open Mind Common Sense (OPCS) database \cite{email-buddy}. Additional methods such as part of speech tagging, named entity extraction, head noun extraction, Porter stemming and word stopping are also used to categorise and normalise the data. Another popular method is to assign the emotion to the word with the highest affective weight. A Vector Space Model has been used to classify emotions based on the International Survey on Emotion Antecedents and Reactions (ISEAR) dataset labelled with basic emotions, which performed significantly better than Na\"{i}ve Bayes and SVM classifiers and also the \textsc{ConceptNet} lexicon \cite{feeler}. However, the underlying assumption remains that the emotions being seeded are in fact universal basic emotion types that best represent the entirety of the emotion spectrum. 

As previously mentioned, \citeasnoun{twitter4} used the emotions ontology \textsc{SentiWordNet} to create a sentiment score based on the emotional score of each word. \possessivecite{semeval} \textsc{CLaC} system used the reverse of this technique by using \textsc{General Inquirer}, similar to \citeasnoun{twitter1}, to define each emotion and seeded term as a ratio of valence. \textsc{CLaC} took a more dimensional (indeed, \textit{fuzzy}) approach to the nature of emotion recognition and consequently had better accuracy and precision over \textsc{UPAR7}. This is similar to \possessivecite{kim} work who used the Affective Norms for English Words (ANEW) dataset to define words in terms of valence, arousal and dominance (VAD --- akin to PAD, see Section~\ref{coreaffect}), mapping documents in a 3D emotional space, which proved most effective on the ISEAR dataset. This 3D mapping technique was also used to map basic emotions to calculate reactions of the sociable robot Kismet \cite{robots}.

\subsection{Mood Analysis}

The work of \citeasnoun{gpoms1} involved creating an extended version of the psychometric instrument Profile Of Mood States (POMS) \cite{POMS}, labelled POMS-ex, by seeding the original emotion adjectives describing aspects of mood using WordNet, extending the original 65 terms to nearly 800 terms. This was used for data mining on Twitter, only retaining tweets that match the regular expressions \textit{`feel'}, \textit{`I'm'}, \textit{`Im'}, \textit{`am'}, \textit{`being'}, and \textit{`be'}, since the only tweets of interest where those that represent an explicit expression of individual sentiment, or those that reflect an individual's present status \cite{gpoms}. This was improved two years later by extending the original 65 terms to a lexicon of over 950 associated terms by analyzing word co-occurrences in Google's Web 1T 5-Grams (Web1T5) database, creating Google Profile of Mood States (GPOMS) \cite{stock}. Initial findings show that the \textit{``happy"} dimension of GPOMS best approximates \textsc{OpinionFinder}'s trend, and the use of 4- and 5-grams observed in publicly accessible web pages makes the lexicon more organic compared to seeding using individual terms.

\subsection{Comparison with Sentiment Analysis}

\begin{figure*}
\centering
\includegraphics[height=85px]{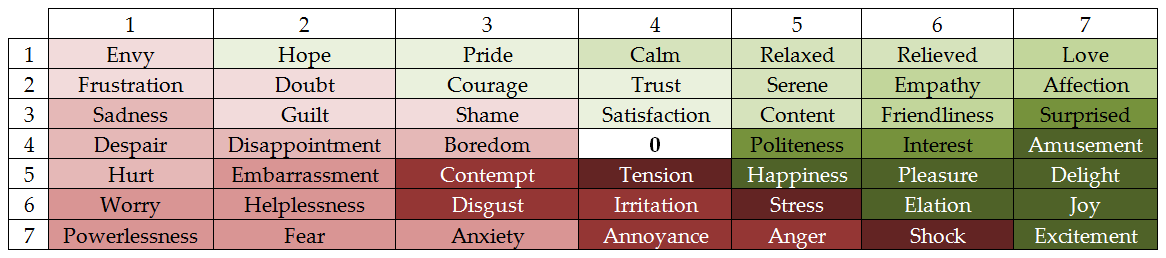}
\caption[Discrete Emotional Response Matrix]{Discrete emotional response matrix.}
\label{ECM}
\end{figure*}

Although sentiment is easier to classify --- indeed it is just polarity --- using specific emotions rather than a valence ratio of words provides a greater insight into the dimensions, and thus nature, of an expressed feeling. We conduct a brief preliminary comparison of basic emotion extraction using keyword spotting and sentiment extraction using \textsc{Twitter Sentiment} to reveal any similarities and discrepancies as a result of using the increased dimensionality of emotion. As our emotion keywords we use Humaine's Emotion Annotation and Representation Language (EARL) (2006)\footnote{\texttt{http://emotion-research.net/projects/humaine/earl}} which classifies 48 emotions in eight categories and correlates nicely with Watson and Tellegen's CTA in terms of matching descriptions of octants, illustrated in Figure~\ref{ECM}. Using these emotion keywords as a basis, we use WordNet \cite{wordnet} to find synonyms for each keyword, arriving at a set of 60 positive and 60 negative emotional keywords distributed among each of the tithes in our emotional space, as shown in Appendix~\ref{EARL}. Using this set of emotion keywords, we created a database of 1.5 million tweets in three days using the Twitter streaming API. Five random terms were chosen and results are shown in Table~\ref{sentiment}. 

\begin{table}
\centering
\begin{tabular}{l*{3}{c}r}
\textbf{Term}       & \textbf{Ave(TS)Pos\%} & \textbf{Ave(EK)Pos\%} & \textbf{Diff\%} \\
\hline
Chocolate  & 71 & 63 & 8 \\
Movie     & 66 & 64 & 2 \\
Dentist    & 33 & 55 & \textbf{22}  \\
Crying    & 21 & 32 & 11  \\
School    & 35 & 56 & \textbf{21} \\
\end{tabular}
\caption[Comparison of Twitter Sentiment and Emotion Keyword Extraction]{Mean average percentage of positive tweets using \textsc{Twitter Sentiment} (TS) and Emotion Keyword Extraction (EK) of five example keywords over a three day period.}
\label{sentiment}
\end{table}

Comparing polarity results from emotion keyword spotting and the software \textsc{Twitter Sentiment} (recall this uses a Maximum Entropy Classifier and considers all tweets as opposed to `explicitly emotional' tweets), it is safe to say that there is a significant correlation between the two. We look at two cases where the difference was significant: \textit{dentist} and \textit{school}. When the top two emotion keywords were analysed for \textit{dentist}, we found that our emotion keyword spotting method treated \textit{brave} as a positive emotion keyword, and 19\% hit this term, contrasted with 11\% for the term \textit{nervous}. \textsc{Twitter Sentiment} was unable to identify \textit{brave} as a positive. For \textit{school}, it treated \textit{glad} as a positive, which accounted for 15\%, whereas \textsc{Twitter Sentiment} would classify a tweet such as \textit{``I am glad that's over"} as a negative. In this example, the emotion elicited is indeed \textit{gladness}, however \textsc{Twitter Sentiment} fails to detect this.

We conclude, albeit crudely, that a `na\"{i}ve' emotion keyword spotting system to measure sentiment performs somewhat comparably with a trained classifier. Performance is dependent on the emotion keywords that are used and which are considered to be classified as positive and negative, but ultimately an increase in data granularity is achieved, concentrating on specific emotions. We hence presume that trained emotion classifiers implicitly trump trained sentiment classifiers, although this statement should be further investigated (see Chapter~\ref{further-research}).

\section{Discussion}

Many methods have been devised in attempt to capture emotion from text. Although a lot work has been done in the field of emotion extraction from arbitrary text, less focus has been drawn to analysing the nature of the emotions themselves. Basic emotions sets are used within emotion extraction as an attempt to define distinct clusters of discrete emotion terms with the aim of representing the entire emotion spectrum. Thus it is important to analyse existing basic emotion sets with the view to discern improved versions so that clusters are correctly labelled to mitigate any neglect of regions of the emotion spectrum, instigating an equally distributed seeding of synonyms for emotion classification. 

With regards to our selection of mechanism for emotion mining, since we are only interested in the words surrounding a specific emotion keyword, we need only use keyword spotting to extract tweets containing each specific emotion keyword. While simple statistical methods are often used with extracted emotion to infer trends and psychological analytics \citeaffixed{wefeelfine}{for example the \textit{We Feel Fine} emotion crawler and database,}, we require a more powerful form of statistics to analyse the meaning, or \textit{semantics}, of expressions at a more deeper level. We thus draw our attention to semantic analysis techniques.


\chapter{Semantic Analysis}

Over the past few decades there has been significant evidence that people's psychological aspects can be predicted through analysis of language style. One of the most notable examples of this is \possessivecite{schizo} work on verbal behavior and schizophrenia. They found that, while the speech of those diagnosed with schizophrenia did not differ with unaffected people on the structural level, it did differ on the semantic level, i.e. with regard to thematic concerns that were being addressed. It is this deviation from expected thematic concerns, which are linked to general and sex-specific social role expectations, that is associated with the diagnosis of schizophrenia \cite{schizo}. Analysis of language semantics has been used extensively in research, including discovering individual differences in personality \cite{personality}, lie detection \cite{lie} and discovering individual differences in beliefs \cite{bilovich}.

By analysing the semantics, specifically, co-occurrence statistics, of the language expressing individual emotion keywords, we can discern those emotions that are similar and those that are more distinct. We postulate that similar emotions are represented by similar semantics, and propose to cluster emotional documents based on the underlying meanings of each document. In order to analyse a corpus, we first require all target documents to be in the same Euclidean space, and thus introduce the concept of a semantic space.

\section{Semantic Space Theory}

The central assumption when creating a semantic space is that the context surrounding a given word provides important information about its meaning \cite{harris}. The semantic properties of words are represented by capturing statistical distributions of co-occurrence with neighboring words which is then projected into a multi-dimensional Euclidean space where each axis represents a context word. The semantic similarity of any two words can then be calculated, typically using the Euclidean distance between the two points or the cosine of the angles between them. Formally, a semantic space is a quadruple \texttt{$<$A, B, S, M$>$}, where \texttt{A} is the Lexical Association Function, \texttt{B} is the Basis, \texttt{S} is the Similarity Measure in $\Re^{| \texttt{B}|} \times \Re^{| \texttt{B}|}$ and \texttt{M} is an optional transformational Model \cite{lowe}.

Zipf's law states that in a (syntactically and semantically correct) textual corpus, the frequency of any word is roughly inversely proportional to its frequency rank \cite{zipf}. The lexical association function \texttt{A} is a normalising weighting scheme that is applied to the raw co-occurrence count matrix to mitigate resulting data sparsity and chance co-occurrences of words. Perhaps the most effective weighting function is the Pointwise Mutual Information method (PMI) \cite{PMI}. Indeed it has recently been proven to be more accurate than using \textit{idf} (inverse document frequency) or \textit{tf-idf} (term frequency-inverse document frequency) normalisation techniques for contextual similarity and feature selection \cite{idf}, and \citeasnoun{twitter1} uses PMI to successfully measure the association of an elicited emotion with a target-dependent features. Another weighting function, referred to as \textit{log-entropy}, is defined as:

\begin{equation}
G(i) = \frac{1 + (\sum_{j} p(i,j) \text{log} (p(i,j)))}{\text{log} (ndocs)}
\end{equation}

for term $i$ and document $j$ \cite{lsa-weight}. This is both the \textit{de facto} normalisation function for Latent Semantic Analysis (see section~\ref{sec:lsa}) and found to be much more accurate than \textit{idf} or \textit{gf-idf} by \citeasnoun{lsa-weight}.

The similarity measure \texttt{S}, as previously mentioned, is usually the Euclidean distance or the cosine of angles between two vectors in the semantic space. The cosine of the angle $\theta$ between the vectors $\textbf{x}$ and $\textbf{y}$ with \textit{n} elements, where $\textbf{x}=\langle x_{1},x_{2},...,x_{n} \rangle$ and $\textbf{y}=\langle y_{1},y_{2},...,y_{n} \rangle$ can be calculated as follows:
\begin{equation}\label{cosine-sim}
\cos(\textbf{x},\textbf{y}) = \frac{\sum^n_{i=1}x_{i} \cdot y_{i}}{\sqrt{\sum^n_{i=1}x^2_{i} \cdot \sum^n_{i=1} y^2_{i}}}
\end{equation}
It has been shown that the inner product is a more appropriate method for measuring the syntagmatic (`chunk-level') similarity, whereas the cosine was a better method for measuring the paradigmatic (`collection-level') similarity \cite{las-jap}. It has also been shown that the cosine of PMI vectors outperformed both the Euclidean distance of PMI vectors and the cosine of ratio vectors \cite{PPMI}. Kullback-Leibler divergence has also been shown to be less accurate than calculating the cosine \cite{PPMI,las-jap}. Topic synsets can be represented by summing up the vectors of individual terms contained within each topic, and then compared with individual documents or other synsets.

The basis of the semantic space, \texttt{B}, defines the dimensions of the semantic space, and is usually the vocabulary. The main issue with using a vector space model is its high dimensionality, resulting in difficulties in capturing the latent semantic information. Therefore \texttt{B} is usually truncated to the \textit{k} most frequent words. Other techniques such as porter stemming and word-stopping can be applied to \texttt{B} in an attempt to reduce the dimensionality of the semantic subspace, although by doing so a certain degree of semantic information is unrepresented in the semantic space.

A semantic space is fully functional when \texttt{A}, \texttt{B} and \texttt{S} have been specified. Indeed, \citeasnoun{bilovich} did not include \texttt{M}, instead truncating low frequency words directly. However it is possible to build a more structured mathematical or statistical model with the addition of \texttt{M} \cite{lowe} to increase the accuracy of the analysis. Latent Semantic Analysis overcomes the issues of data sparseness and high dimensionality through the use of dimensionality reduction methods.

\section{Latent Semantic Analysis}\label{sec:lsa}

Latent Semantic Analysis (LSA) \cite{lsa} is a variant of the vector space model that aims to create a semantic space by means of dimensionality reduction techniques and has been widely used in a variety of domains, from document indexing to essay grading. It has also been used in emotion classification of news headlines, performing better than Na\"{i}ve Bayes for recall but not as good as \textsc{WordNet} for precision \cite{headlines}.

Given a raw co-occurrence matrix \textbf{M} using the entire vocabulary as \texttt{B}, this is transformed by \texttt{A} (recall the documented function for LSA is log-entropy normalisation), and \texttt{M} is applied to reduce dimensionality. There are several techniques for \texttt{M} that reduce the dimensionality of words constituting the semantic space, the original method documented for LSA being Partial Singular Value Decomposition.

\subsection{Partial Singular Value Decomposition}

Partial Singular Value Decomposition (PSVD) is the \textit{de facto} method for dimensionality reduction in LSA. It uses Singular Value Decomposition (SVD) to decompose the data matrix \textbf{M} into the product of three matrices:
\begin{equation}
\textbf{M} = \textbf{T} \textbf{$\Sigma$} \textbf{D}^T
\end{equation}
where \textbf{T} is the term matrix, \textbf{D} is the document matrix and \textbf{$\Sigma$} is a diagonal matrix with singular values sorted in decreasing order that act as scaling factors that identify the varience with each dimension. LSA uses a truncated SVD, keeping only the \textit{k} largest singular values in \textbf{$\Sigma$} and their associated vectors: 
\begin{equation}
\textbf{M} \approx \textbf{M}_{k} = \textbf{T}_{k} \textbf{$\Sigma$}_{k} \textbf{D}_{k}^T
\end{equation}
This reduced-dimension SVD, or PSVD, $\textbf{M}_{k}$, is the best approximation to \textbf{M} with \textit{k} parameters, and is what LSA uses for its semantic space. The rows in $\textbf{D}_{k}$ are the document vectors and the rows in $\textbf{T}_{k}$ are the term vectors in LSA space. As \textbf{T} and \textbf{D} have orthogonal columns, the derived latent semantic space is orthogonal, and, combined with the fact that documents can take negative values in this space, makes the derived semantic space less likely to correspond to clusters that overlap \cite{nmf-doc}. Additionally, the complexity of SVD is $O(D^3)$ and so makes it unsuitable for very large datasets, although methods have been developed to append data without having to recompute, most notably \textit{folding-up} \cite{foldup}. PSVD has recently been successfully used to identify latent factors of political alignment using hashtags on Twitter \cite{twitter-lsa}. Traditional clustering techniques, such as K-means, can be applied in this space to derive semantic clusters.

\subsection{Comparison to Non-negative Matrix Factorisation}

Non-Negative Matrix Factorisation (NMF) \cite{nmf} is a multivariate analysis method originally developed for computer vision purposes, and increasingly being used within the field of dimensionality reduction, distinguished from other methods by its use of non-negativity constraints and a parts-based representation. With regard to document clustering, it treats documents as an additive combination of the base latent semantics and as a result, produces a semantic space that provides a direct indication of data partitions without the need for additional data clustering techniques \cite{nmf-doc}. Terms and documents are treated in an integrated fashion without the requirement of a diagonal matrix enabling natural \textit{soft clustering}, i.e. combining the term and document matrix for clustering. NMF finds the positive factorisation of the data matrix, \textbf{M}, such that
\begin{equation} 
\textbf{M} \approx \textbf{W} \textbf{H}^T
\end{equation}
NMF is typically faster than SVD; SVD is computationally expensive and does not produce intermediary results as it is not an optimisation procedure unlike NMF \cite{doc-clust}. \citeasnoun{nmf-doc} found NMF weighted by the normalised cut weighting scheme to be more accurate than SVD for document clustering although \citeasnoun{lsa-nmf} found negligible improvements over SVD when applied specifically for LSA. Recently, \citeasnoun{las-jap} found that SVD yielded better performance in both synonym judgement and word association over NMF. NMF is an example of a latent topic model, where a topic is represented by a distribution of words, and is equivalent to several other models in this field through parameter selection. NMF via KL divergence is equivalent to Probabilistic Latent Semantic Analysis (pLSA) \cite{plsa}, and pLSA using a uniform dirichlet prior distribution is equivalent to Latent Dirichlet Analysis (LDA) \cite{lda}, although these are generally regarded as topic modelling techniques. Latent Periodic Topic Analysis (LPTA) is a recent model that aims to represent \textit{periodic} topics, ideal for use with temporal data to distinguish between periodic, bursty and background topics \cite{lpta}.

\section{Algorithm Complexity and Human Cognition}\label{sec-moredata}

Although dimension reduction techniques such as LSA and NMF are effective, the question whether the assumed complexity of these algorithms best models human cognition must not be ignored. In a recent study, \citeasnoun{more-data} demonstrated that, although LSA was marginally more effective than PMI when using the same amount of data, PMI trained on large amounts of data far surpassed the effectiveness of LSA trained on less data. \citeasnoun{more-data} pose the theory that the requisite complexity of human behavior may already be present in the structure of language without the need to build this complexity into models such as LSA, if a realistic sample is taken.

PMI-IR, a version of PMI that uses Information Retrieval from the Internet, has also shown the effectiveness of PMI trained on more data over LSA, with just under a 10\% increase in precision in the Test of English as a Foreign Language (TOEFL) experiment, and a similar gain in the English as a Second Language (ESL) experiment \cite{pmi-ir}. Another approach that utilities massive amounts of world knowledge is Explicit Semantic Analysis (ESA), which uses Wikipedia to create a semantic space, resulting in an almost 20\% increase in recall for individual words, and just over a 10\% increase in recall for texts compared to LSA \cite{esa}. In an age where million-scale datasets are more frequently created and used, and until we have the processing power to use more complex algorithms such as LSA on these massive datasets, less complex algorithms such as PMI-IR and ESA may provide a viable alternative to approaching the human judgment problem, being relatively simple, scalable and easily incremental.

\section{Discussion}

There exist many algorithms that create and analyse high-dimensional linguistic spaces. The most widely used algorithms, LSA and NMF, focus on transforming word co-occurrence and document matrices in order to infer the underlying semantics of a particular corpus. Other techniques, such as PMI and ESA, use simpler mechanisms but compensate by throwing more data at the problem. For this study, however, we only need concentrate on clustering documents that do not exceed the computational limitations of LSA. Given that LSA is ultimately more accurate than other methods for inferring latent semantics, we select LSA as our method of semantic analysis, and install the Enthought Python Distribution (EPD) 7.2 on an x86 architecture to use the \textsc{gensim} framework \cite{radim} to create latent semantic clusters of an emotion corpus harvested from Twitter.


\chapter{The Emotional Twitter Corpus}

We are now in a position to discuss our data collection process. As previously mentioned, there are many basic emotion theories, each defining a distinct set of emotion keywords. We explain our selection of basic emotion sets and thus the emotion keywords used in this study, and the mechanisms we used to create a corpus of emotion expressions from Twitter.

\section{Basic Emotion Sets}

Our first task was to decide on the emotion sets we would evaluate. Originally starting with the list of basic emotion sets tabulated in \possessivecite{basic} study, we removed sets that have four or less emotion keywords, as these sets were regarded as not being detailed enough for significant analysis, more akin to a uni-dimensional valence-focused model. Sets that contain emotion keywords that have an unfeasibly low stream rate (see section \ref{sec-esr}) were also removed, as these were considered impractical to fully evaluate within the timescale and resources of this project. We additionally included Russell's dimensional categories from his CTA as a basic emotion set, resulting in a collection of six basic emotion theories as described in Table \ref{sets}.

\begin{table}
\centering

\begin{tabular}{l*{2}{p{7cm}}r}
\textbf{Basic Emotion Theory}       & \textbf{Identified Basic Emotions} \\
\hline
\citeasnoun{izard}  & Anger, Contempt, Disgust, Distress, Fear, Guilt, Interest, Joy, Shame, Surprise  \\
\possessivecite{russell} Categories  & Angry, Depressed, Distressed, Excited, Miserable, Pleased, Relaxed, Sleepy \\
\citeasnoun{plutchik}  & Acceptance, Anger, Anticipation, Disgust, Joy, Fear, Sadness, Surprise  \\
\citeasnoun{ekman-face}  & Anger, Disgust, Fear, Joy, Sadness, Surprise \\
\citeasnoun{tomkins}  & Anger, Interest, Contempt, Disgust, Distress, Fear, Joy, Shame, Surprise \\
\citeasnoun{johnson}  & Anger, Disgust, Anxiety, Happiness, Sadness \\
\end{tabular}

\caption[Basic Emotion Sets to be analysed]{Basic Emotion sets from the most notable Basic Emotion theories that were analysed.}
\label{sets}
\end{table}

\section{Emotion Keywords}

The extraction mechanism and the selection of keywords to be mined from Twitter would form the structure of our eventual emotion corpus. Taking the union of all the emotion sets identified for analysis, we obtained a set of 21 unique emotion keywords, which, theoretically, constitute the most distinct emotions. We extract unigrams created using the first person grammatical inflection of each keyword, similar to \citeasnoun{russell}, as most tweets will contain this type of inflection: \textit{``I am very \textbf{excited} today"} as opposed to \textit{``I am feeling \textbf{excitement} today"}. This was chosen as opposed to mining for bigrams, for example \textit{``feeling excited"}, \textit{``feel excited"} and \textit{``felt excited"}, as this resulted in far fewer tweets being returned due to Twitter's indexing focusing on single keywords. Moreover, tweets containing quantifiers would have been ignored if we chose to extract bigrams, for example \textit{``feeling \textbf{very} excited"}.

Contrary to \possessivecite{gpoms} work, we did not require tweets to contain the words \textit{`feel'}, \textit{`I'm'}, \textit{`Im'}, \textit{`am'}, \textit{`being'}, and \textit{`be'}, as an explicit mention of an emotion keyword would be sufficient to describe an experience of that emotion, reinforced by the fact that we will only be mining for the first person grammatical inflection of each keyword. We filtered out re-tweets --- minimising duplicates --- and negative tweets, because, for example, \textit{`happy'} $\neq$ \textit{`not happy'}; nor can we assume that \textit{`not sad'} $=$ \textit{`happy'}. tweets that include popular phrases including an emotion keyword such as \textit{``Happy Birthday"} and \textit{``Angry Birds"} were also filtered out from being written to the database.  Initially, @ tags were not filtered, but we quickly realised that these tweets refer to messages either closely relating to other people or as part of a thread of messages; thus we filtered them out as the emotion expressed within such tweets did not describe an atomic emotional experience. Table \ref{words} tabulates the complete set of emotion keywords and filtered phrases used for emotion extraction from Twitter.

We did not Porter stem collected words as \citeasnoun{kim} notes that this might hide important semantic differences, for example, conceptual differences between \textit{loved} and \textit{loving}. To optimally harvest emotions, we substituted \textit{fear} with \textit{scared} as it was proven to be the most popular keyword out of \textit{scared}, \textit{frightened} and \textit{afraid}. We also substituted \textit{distressed} with \textit{stressed}, due to an extremely low stream rate for this keyword (see section \ref{sec-esr}).

\begin{table}
\centering

\begin{tabular}{l*{3}{l}r}
\textbf{Term}       & \textbf{Filtered Phrases} & \textbf{Total tweets} \\
\hline
accepting  & not accepting, unaccepting & 1847 \\
angry  & not angry, angry birds & 7540 \\
anticipating  & not anticipating & 1493 \\
anxious  & not anxious & 1964 \\
ashamed  & not ashamed, unashamed & 2871 \\
contempt  & not contempt, in contempt, contempt of & 1287 \\
depressed  & not depressed & 4347 \\
disgusted  & not disgusted & 1607 \\
excited  & not excited, unexcited & 20629 \\
guilty  & not guilty, found guilty & 3450 \\
happy  & not happy, unhappy, happy birthday & 57631 \\
interested  & not interested, uninterested, disinterested & 4394 \\
joyful  & not joyful, unjoyful & 1249 \\
miserable  & not miserable & 3164 \\
pleased  & not pleased, displeased & 1872 \\
relaxed  & not relaxed, unrelaxed & 1817 \\
sad  & not sad & 26721 \\
scared  & not scared & 15273 \\
sleepy  & not sleepy & 9786 \\
stressed     & not stressed, unstressed & 4237 \\
surprised    & not surprised, unsurprised & 5625 \\
\textbf{ALL}    & \textbf{http://, RT, $<10$ words } & \textbf{178804}\\
\end{tabular}

\caption[Tracked Emotion Keywords and Filtered Phrases]{Emotion Keywords that were tracked on Twitter, with any tweets containing a filtered phrase not stored in the database. Total number of tweets are correct as of 28th April 2012 and reflects additional extraction carried out after the analysis for the current project was conducted.}
\label{words}
\end{table}

\section{Emotion Streaming}

The next stage involved mining and storing tweets. A PHP script was created that used the Gardenhose Level Twitter Streaming API --- a streaming sample of about 10\% of all public status updates on Twitter --- that allows tracking of up to 400 keywords. We collected tweets that contained each of the emotion keywords in Table \ref{words}, storing those which do not include a filtered phrase into a MySQL database. We programmed the PHP script to be cyclical in the sense that it streamed individual tweets, but changed emotion keywords every 5 minutes in order to collect the whole range of emotions without having to restart the script each time. Ten days of data collection resulted in a labelled Temporal Emotion Database containing six emotion theories totaling to 21 unique emotion keywords each with at least 1100 documents to base our analysis on. The PHP script can be found in Appendix~\ref{streamword}.

\subsection{Emotion Stream Rate} \label{sec-esr}

\begin{landscape}
\begin{figure*}
\centering
\includegraphics[width=555px]{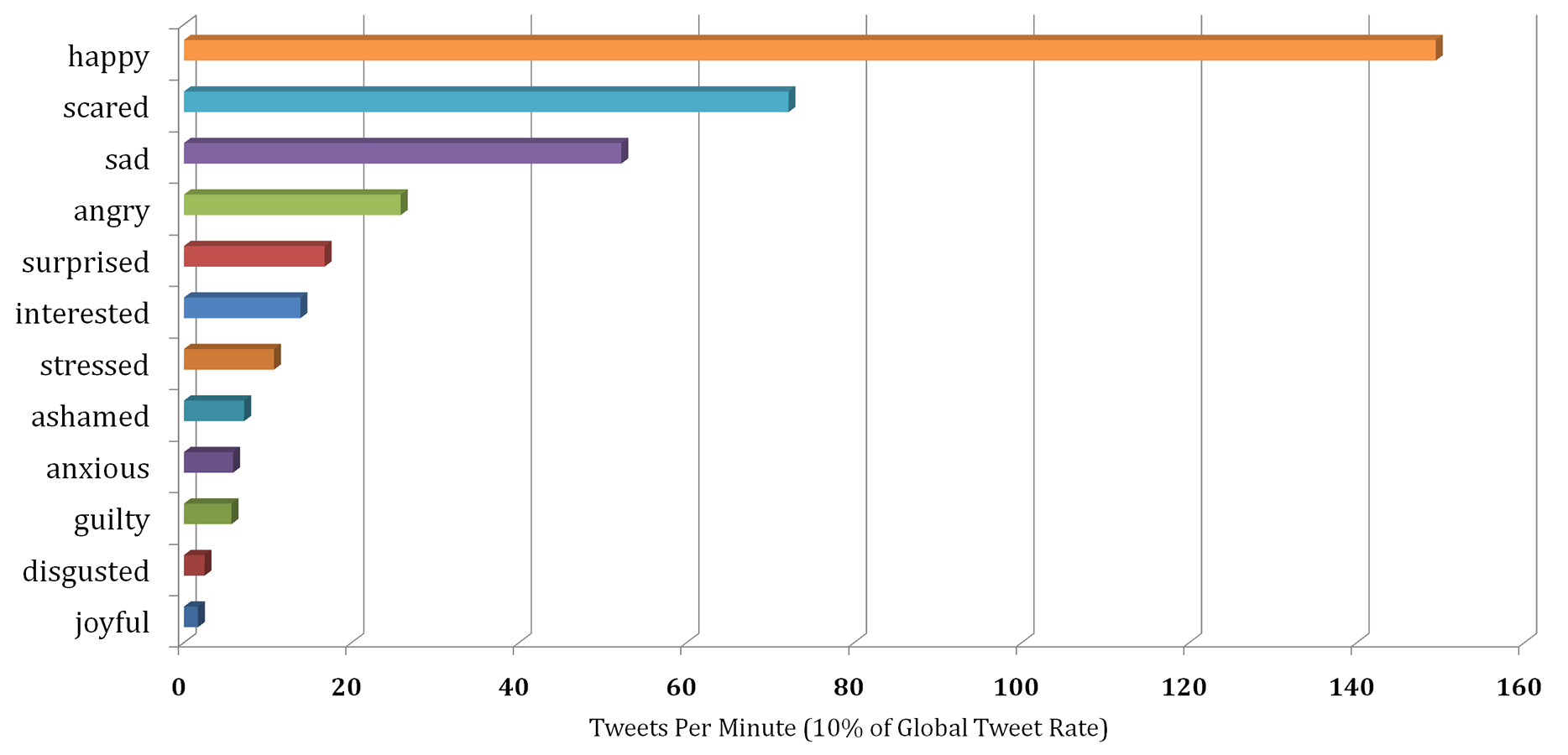}
\caption[Emotion Stream Rates for selected Emotion Keywords]{Emotion Stream Rates for selected Emotion Keywords.}
\label{esr}
\end{figure*}
\end{landscape}

We originally tracked emotions within each emotion set at the same time, however this produced biased results as the \textit{Emotion Stream Rate} (ESR) deviated the number of tweets proportionately to each emotion, and thus results were skewed in favour of more `popular' emotions. Figure~\ref{esr} illustrates this problem. A notable example was the keyword \textit{distressed}, with an ESR of less than 1 tweet per hour, so to collect 1000 tweets would take over 40 days. Seeing as we could only stream one emotion at a time due to Twitter API limitations, this was deemed unfeasible and so \textit{distressed} was substituted with \textit{stressed}.

We thus tracked each emotion keyword individually, which also created the advantage of being able to label tweets with their respective emotion in the database for efficient retrieval, in addition to the ability to measure each emotion's stream rate. It should be noted that by using \textsc{WordNet} \cite{wordnet} we could have expanded our initial list of 21 keywords by taking synonyms of each keyword and testing the ESR for each emotion, selecting the most popular keyword; however in order to fairly test each theory, we opted against this as the selected emotion keywords had been carefully chosen by each theorist.

\section{Discussion}

We have created a PHP script that streams a large number of emotional expressions from Twitter that can be executed from a UNIX terminal and run in the background. Using this script, we have created a corpus of emotional tweets, stored in a database, with each expression emotion-tagged and linked to additional information such as time (useful for temporal analysis) and nationality (useful for cultural analysis). In the process, we have discovered that the stream rates of emotions are highly irregular, ranging from 2 to 200 tweets per minute for the emotions analysed, including any word filters that were used. Using consistent quantities of tweets for each emotion was key to this research, so a natural limit of the number of documents we could analyse was imposed, this being the number of documents harvested from the emotion with the lowest stream rate. 

The primary limitation of this generated corpus is that it collects expressions that mention an emotion keyword without regard to whether the user actually felt the expressed emotion. However, we assume that the mere use of an emotion keyword still attributes to how people understand each emotion to be. Although we could have used many more popular emotion keywords and synonyms, we decided against this as the current study focuses on specific basic emotion sets, however this should be explored in further research (see Chapter~\ref{further-research}).

Our generated Twitter corpus is useful for those wanting to obtain raw explicit expressions of emotion. Each expression is labelled, rendering it especially useful for classification projects, although the data as it stands remains unparsed for duplicates or identified `junk' data. We can query our database to retrieve up to 1100 expressions for each of the 21 emotions tracked, which we incorporate into our Python code to perform semantic analysis.


\chapter{Semantic Emotion Analysis}\label{sea}

Having created an emotional Twitter corpus, we analyse this data in order to evaluate the semantic distinctiveness of existing basic emotion sets. We also develop an iterative latent semantic clustering algorithm to discern the optimal semantically irreducible basic emotion set from all 21 emotions collected. Latent Semantic Clustering (LSC) is a simple modification of the LSA algorithm which we base our DELSAR algorithm on. Given a labelled corpus $C$ with label set $K$, it calculates, using LSA, the semantic accuracy for each $label \in K$, thus providing an analysis of how distinct the labelling of $C$ and the selection of $K$ is. All analysis was performed on an Intel Core 2 U7700 CPU 2x1.33GHz with 2GB RAM. Unless specified, we tested dimensions of the LSA space in increments of 10 and selected the dimensionality that performed optimally for each task, similar to \citeasnoun{more-data}. For all tasks, we use Log-Entropy normalisation as our Association Function, found to generate optimal results by \citeasnoun{lsa-weight} and recommended by \citeasnoun{lsa}.

\section{DELSAR}

Document-Emotion Latent Semantic Algorithmic Reducer (DELSAR) takes an emotion set and clusters each document's emotion to the emotion of its closest document vector (excluding itself), calculating a clustering accuracy for each emotion. The closest document vector is calculated as the maximum cosine value of the angle between the current document and each other document in the subcorpus. The emotion keyword in each document is removed before the closest document vector is calculated, so we focus purely on the words surrounding the emotion keyword for each document. DELSAR operates in the LSA space created from the subcorpus of all documents matching all emotion keywords in the set being analysed, in which there are $(doc\_limit \times number\_of\_emotions)$ documents.

If a document expressing a certain emotion, $e$, is not clustered with a document of the same emotion, then the words surrounding $e$ is more similar to the words surrounding another emotion. Thus the clustering accuracy of an emotion set corresponds to how distinct that emotion set is; the more semantically accurate an emotion set is, the more distinct the language surrounding each emotion within the set is.

The reduction aspect of DELSAR initially starts with the set of all 21 emotions. After calculating the clustering accuracies for each emotion, it removes the least accurate emotion from the set and iterates until there are $n$ emotions remaining in the initial set, resulting in the optimal semantically distinct basic emotion set. The DELSAR algorithm is described in Algorithm \ref{DELSAR} and the Python code can be found in Appendix \ref{elsacode}. The raw Python output of the DELSAR algorithm can be found in Appendix~\ref{delsar-print}.

\begin{algorithm}
\caption{DELSAR}
\label{DELSAR}
\begin{algorithmic}

\REQUIRE Final keyword set size $reduceTo$, Corpus \textbf{C} and Keyword Set \textbf{K}, where $\forall document \in$ \textbf{C} $\exists document \rightarrow emotion \in$ \textbf{K}
\STATE calculate cosine document similarity matrix of LSC(\textbf{C}, \textbf{K})

\FOR{\textbf{each} \textit{document} $\in$ \textbf{C}} 
\STATE \textbf{delete} $emotion$ in \textit{document}
\STATE Find closest document vector $nearest$ where $nearest$ $\neq$ \textit{document}
\IF{
\STATE $nearest$(\textbf{K}) == $document$(\textbf{K})
}
\STATE $document$ is a hit
\ELSE
\STATE $document$ is a miss
\ENDIF
\ENDFOR \textbf{ each}

\FOR{\textbf{each} $emotion$ $\in$ \textbf{K}} 
\STATE calculate accuracy of $emotion$ using (total \textit{document} hits where $emotion$ in \textit{document}$/$total \textit{document} where $emotion$ in \textit{document})
\ENDFOR \textbf{ each}

\IF{
\STATE \textbf{length}(\textbf{K}) $>$ $reduceTo$
}
\STATE \textbf{delete} least accurate $word$ in \textbf{K}
\STATE DELSAR(reduced \textbf{K})
\ELSE
\RETURN \textbf{K}
\ENDIF

\end{algorithmic}
\end{algorithm}

\subsection{Analysis}

We performed DELSAR1000 on the corpus and various subcorpora and report the results in Table \ref{DELSAR1000}. Recall that DELSAR creates an LSA space of all documents within each emotion set; for each basic emotion set an LSA space of $(1000 \times number\_of\_emotions)$ documents is created. Evaluating all sets, our results show the accuracy of clustering each document to its nearest document, whether it is the same or another emotion. Out of the theories analysed, Ekman's set proved to be the most semantically distinct, with a 2.9\% increase in accuracy compared to the average of the remaining sets. Russell's categories performed worst, which is surprising seeing as these emotions were taken as the basis for representing the entire emotion spectrum as a whole.

\begin{table}
\centering
\scriptsize
\begin{tabular}{r|*{8}{c}r}
& \multicolumn{8}{c}{\textbf{Model}} \\
\hline
	&\textbf{Izard}	&\textbf{Russell}	&\textbf{Plutchik}	&\textbf{Ekman}	&\textbf{Tomkins}	&\textbf{Oatley}	&\textbf{All}	&\textbf{DELSAR}\\
\textbf{Dimension}	&\textbf{40}	&\textbf{30}	&\textbf{30}	&\textbf{30}	&\textbf{30}	&\textbf{30}	&\textbf{60}	&\textbf{40}\\
\hline
accepting	&	&	&0.583	&	&	&	&0.452	&0.553\\
angry	&0.390	&0.409	&0.400	&0.429	&0.391	&0.468	&0.248	&\\
anticipating	&	&	&0.455	&	&	&	&0.312	&\\
anxious	&	&	&	&	&	&0.535	&0.272	&\\
ashamed	&0.452	&	&	&	&0.467	&	&0.366	&0.534\\
contempt	&0.550	&	&	&	&0.575	&	&0.356	&0.574\\
depressed	&	&0.292	&	&	&	&	&0.193	&\\
disgusted	&0.364	&	&0.417	&0.484	&0.422	&0.527	&0.251	&\\
excited	&	&0.407	&	&	&	&	&0.227	&\\
guilty	&0.426	&	&	&	&	&	&0.339	&\\
happy	&	&	&	&	&	&0.411	&0.255	&\\
interested	&0.561	&	&	&	&0.560	&	&0.460	&0.603\\
joyful	&0.482	&	&0.518	&0.565	&0.507	&	&0.397	&0.519\\
miserable	&	&0.413	&	&	&	&	&0.272	&\\
pleased	&	&0.548	&	&	&	&	&0.359	&0.506\\
relaxed	&	&0.383	&	&	&	&	&0.245	&\\
sad	&	&	&0.388	&0.442	&	&0.424	&0.259	&\\
scared	&0.377	&	&0.456	&0.498	&0.396	&	&0.249	&\\
sleepy	&	&0.445	&	&	&	&	&0.332	&0.591\\
stressed	&0.454	&0.376	&	&	&0.481	&	&0.295	&0.502\\
surprised	&0.416	&	&0.491	&0.505	&0.414	&	&0.295	&\\
\hline
MEAN	&0.447	&0.409	&0.464	&\textbf{0.487}	&0.468	&0.473	&0.306	&\textbf{0.548}\\
\hline
STDEV	&0.068	&0.072	&0.065	&0.049	&0.069	&0.057	&0.072	&0.039\\
\end{tabular}
\caption[DELSAR1000]{DELSAR clustering accuracy of each basic emotion set using a corpus comprised of 1000 documents for each emotion within each set. Standard Deviation of all models is $\sigma = 0.027$.}
\label{DELSAR1000}
\end{table}

We performed DELSAR on the set of all 21 emotions, reducing the set to the eight most semantically distinct dimensions of emotion, these being:

\begin{center}
\textsc{accepting ashamed contempt interested joyful pleased sleepy stressed}
\end{center}

This set achieved a significant increase in accuracy over Ekman's set of 6.1\%. We could say that these emotions best represent the emotion spectrum in its entirety, or in other words, the remaining emotions could be expressed as a combination or a particular degree of intensity of these emotions.

In addition to performing DELSAR1000, we tested four subcorpora of varying document sizes to observe any temporal effects. We took document sizes of 500, 360, 280 and 100 and distributed equally across the whole corpus using a modulus function on the index. We found negligible temporal variance within our results, with a 0.7\% standard deviation of standard deviations for each test, illustrated in Table~\ref{DELSARMOD}. Ekman's set proved to be the best overall, in line with DELSAR1000, although it was outperformed by Oatley's set when lower amounts of documents were used. Again, in all cases, Russell's categories were outperformed by all other sets, and our DELSAR set outperformed each optimal set by as much as 8.8\%.

\begin{table}
\centering
\scriptsize
\begin{tabular}{r|*{8}{c}r}
& \multicolumn{8}{c}{\textbf{Model}} \\
\hline
	&\textbf{Izard}	&\textbf{Russell}	&\textbf{Plutchik}	&\textbf{Ekman}	&\textbf{Tomkins}	&\textbf{Oatley}		 &\textbf{STDEV} &\textbf{DELSAR}\\
\hline
D500 MOD2=0	&0.444	&0.411	&0.461	&\textbf{0.475}	&0.463	&0.472&0.024&\textbf{0.563}	\\
D360 MOD3=0	&0.416	&0.405	&0.440	&\textbf{0.456}	&0.427	&0.449&0.020&\textbf{0.532}	\\
D280 MOD4=0	&0.427	&0.403	&0.414	&0.430	&0.431	&\textbf{0.434}&0.012&\textbf{0.494}	\\
D100 MOD10=0	&0.376	&0.345	&0.396	&0.418	&0.382	&\textbf{0.422}&0.029&\textbf{0.470}	\\
\hline
MEAN	&0.416	&0.391	&0.428	&\textbf{0.445}	&0.426	&0.444&0.020		&\textbf{0.515}	\\
\hline
STDEV	&0.029	&0.031	&0.029	&0.025	&0.033	&0.022&0.007		&0.041	\\
\end{tabular}
\caption[DELSAR Subset Analysis]{DELSAR clustering accuracy using a dimension of 36. Each set uses a subcorpus comprised of 500, 360, 280 and 100 documents of each emotion in the set, selected using a modulus function on the index.}
\label{DELSARMOD}
\end{table}

\section{ELSA}

While DELSAR is highly effective, its analysis is relative to a subcorpus of documents that express all the emotions contained within a particular basic emotion set, so although it is a good measure of showing how distinct a particular emotion set is overall, it does not allow for each emotion to be mutually independent. This is important to take into consideration as it allows us to compare emotions without the constraint of it being in a set with other emotions --- for example an emotion within a set may be considered distinct only because other emotions within the set are not. Emotional Latent Semantic Analysis (ELSA) is a modified version of DELSAR, in which emotions are treated separately from one another. ELSA takes the set of all 21 emotions and, \textit{for each emotion}, creates an LSA space using documents matching only that particular emotion, in which there are $(doc\_limit)$ documents. For each ELSA space, the cosine value for the closest document vector to each document is determined, and an average of these is calculated. The higher this average value is for a specific emotion, the more similar the documents are for that emotion, in other words, the emotion cluster is tightly packed. Lower values mean less similar words being used in the expression of the same emotion --- the emotion cluster is more dispersed --- signifying a decrease in distinctiveness. The difference between ELSA and DELSAR, is that the latter evaluates whether a particular emotion \textit{set} is representative of the entire emotion spectrum, as opposed to seeing which \textit{emotions} are distinct.

Evaluating each basic emotion set according to ELSA is simply a matter of averaging the corresponding values of the constituent emotions, and discerning the most semantically distinct emotions is merely a case of selecting the emotions with maximum average values. The ELSA algorithm is described in Algorithm~\ref{ELSA}.

\begin{algorithm}
\caption{ELSA}
\label{ELSA}
\begin{algorithmic}

\REQUIRE Corpus \textbf{C} and Keyword Set \textbf{K}, where $\forall document \in$ \textbf{C} $\exists document \rightarrow emotion \in$ \textbf{K}

\FOR{\textbf{each} $emotion$ $\in$ \textbf{K}}
\FOR{\textbf{each} \textit{document} $\in$ \textbf{C}} 
\IF{\STATE $document$(\textbf{K}) == $emotion$}
\STATE \textbf{delete} $emotion$ in \textit{document}
\STATE calculate cosine document similarity matrix of LSA(\textit{document}, \textbf{C})
\STATE Find closest document vector $nearest$ where $nearest$ $\neq$ \textit{document}
\ENDIF
\ENDFOR \textbf{ each}
\RETURN average($nearest$)
\ENDFOR \textbf{ each}

\end{algorithmic}
\end{algorithm}

\subsection{Analysis}

We performed ELSA in a similar fashion to DELSAR --- testing the same copora --- and report the results in Table \ref{ELSA1000}. Out of all the basic emotion sets analysed, Tomkin's set proved to contain the most semantically concentrated emotions, although it must be pointed out that Tomkin's set is identical to Ekman's set without the emotion \textit{sad} and four other emotions added; by swapping \textit{disgusted} for \textit{contempt}, Ekman's set would have been optimal at $0.747$.

\begin{table}
\centering
\scriptsize
\begin{tabular}{r|*{8}{c}r}
& \multicolumn{8}{c}{\textbf{Model}} \\
\hline
	&\textbf{Izard}	&\textbf{Russell}	&\textbf{Plutchik}	&\textbf{Ekman}	&\textbf{Tomkins}	&\textbf{Oatley}	&\textbf{All}	&\textbf{ELSA}\\
\hline
accepting	&	&	&0.781	&	&	&	&0.781	&0.781\\
angry	&0.727	&0.727	&0.727	&0.727	&0.727	&0.727	&0.727	&\\
anticipating	&	&	&0.717	&	&	&	&0.717	&\\
anxious	&	&	&	&	&	&0.744	&0.744	&0.744\\
ashamed	&0.743	&	&	&	&0.743	&	&0.743	&0.743\\
contempt	&0.838	&	&	&	&0.838	&	&0.838	&0.838\\
depressed	&	&0.695	&	&	&	&	&0.695	&\\
disgusted	&0.708	&	&0.708	&0.708	&0.708	&0.708	&0.708	&\\
excited	&	&0.708	&	&	&	&	&0.708	&\\
guilty	&0.713	&	&	&	&	&	&0.713	&\\
happy	&	&	&	&	&	&0.694	&0.694	&\\
interested	&0.724	&	&	&	&0.724	&	&0.724	&\\
joyful	&0.761	&	&0.761	&0.761	&0.761	&	&0.761	&0.761\\
miserable	&	&0.744	&	&	&	&	&0.744	&0.744\\
pleased	&	&0.742	&	&	&	&	&0.742	&0.742\\
relaxed	&	&0.707	&	&	&	&	&0.707	&\\
sad	&	&	&0.713	&0.713	&	&0.713	&0.713	&\\
scared	&0.719	&	&0.719	&0.719	&0.719	&	&0.719	&\\
sleepy	&	&0.704	&	&	&	&	&0.704	&\\
stressed	&0.736	&0.736	&	&	&0.736	&	&0.736	&0.736\\
surprised	&0.723	&	&0.723	&0.723	&0.723	&	&0.723	&\\
\hline
MEAN	&0.739	&0.720	&0.731	&0.725	&\textbf{0.742}	&0.717	&0.731	&\textbf{0.761}\\
\hline
STDEV	&0.038	&0.019	&0.026	&0.019	&0.039	&0.019	&0.033	&0.034\\
\end{tabular}
\caption[ELSA1000]{ELSA average cosine values using dimensions 10, 20, 30, 40, 50, 60, 70, 80, 90 and 100. Each emotion uses a corpus of 1000 documents. Standard Deviation of all models is $\sigma = 0.010$.}
\label{ELSA1000}
\end{table}

We obtained a slightly different optimal set consisting of the eight most semantically distinct emotions compared to DELSAR, taking away \textit{interested} and \textit{sleepy} and adding \textit{anxious} and \textit{miserable}:

\begin{center}
\textsc{accepting anxious ashamed contempt joyful miserable pleased stressed}
\end{center}

This set achieved a 1.9\% increase in accuracy compared to Tomkin's set. We could say that these emotions are the most \textit{atomic} in the sense that the words surrounding these emotions are semantically concentrated; people using these emotions are more likely to be actually referring to these emotions due to the similarity of language across all documents. Take \textit{happy} as an example, which is the least atomic emotion: being the least semantically concentrated means that the language that people use when using the word \textit{happy} varies the most, either due to describing a great variety of things, being used in a great variety of contexts, or varying perceptions of what the emotion \textit{happy} actually means. The raw Python output of the ELSA1000 algorithm can be found in Appendix~\ref{elsa-print}.

Similarly to DELSAR, we tested the same four subcorpora to observe any temporal effects, and report the results in Table~\ref{ELSAMOD}. A standard deviation of 0.1\% suggests that there is negligible temporal effects on the corpus, with Tomkin's set again outperforming all other sets, in line with our results from ELSA1000, and our ELSA set outperforming Tomkin's set in each case.

\begin{table}
\centering
\scriptsize
\begin{tabular}{r|*{8}{c}r}
& \multicolumn{8}{c}{\textbf{Model}} \\
\hline
	&\textbf{Izard}	&\textbf{Russell}	&\textbf{Plutchik}	&\textbf{Ekman}	&\textbf{Tomkins}	&\textbf{Oatley}		 &\textbf{STDEV} &\textbf{ELSA}\\
\hline
E500 MOD2=0	&0.688	&0.675	&0.685	&0.678	&\textbf{0.690}	&0.672	&0.007	&\textbf{0.715}\\
E360 MOD3=0	&0.667	&0.653	&0.660	&0.654	&\textbf{0.669}	&0.651	&0.008	&\textbf{0.690}\\
E280 MOD4=0	&0.645	&0.641	&0.640	&0.633	&\textbf{0.648}	&0.632	&0.006	&\textbf{0.675}\\
E100 MOD10=0	&0.549	&0.539	&0.537	&0.536	&\textbf{0.553}	&0.527	&0.009	&\textbf{0.571}\\
\hline
MEAN	&0.637	&0.627	&0.631	&0.625	&\textbf{0.640}	&0.620	&0.008	&\textbf{0.663}\\
\hline
STDEV	&0.061	&0.060	&0.065	&0.063	&0.061	&0.064	&0.001	&0.063\\
\end{tabular}
\caption[ELSA Subset Analysis]{ELSA average cosine values using a dimension of 50. Each emotion uses a subcorpus comprised of 500, 360, 280 and 100 documents, selected using a modulus function on the index.}
\label{ELSAMOD}
\end{table}

For each set we calculated the optimal ELSA set, and found that in E280 the set does not change, in E500 \textit{stressed} is replaced with \textit{surprised}, in E360 \textit{ashamed} is replaced with \textit{surprised} and in E100 \textit{stressed} is replaced with \textit{excited}; negligible changes are observed with the optimal increasing accuracy by only 0.2\% for E360 and E100 and 0\% for E500 and E280.

\begin{landscape}
\begin{figure*}
\centering
\includegraphics[width=555px]{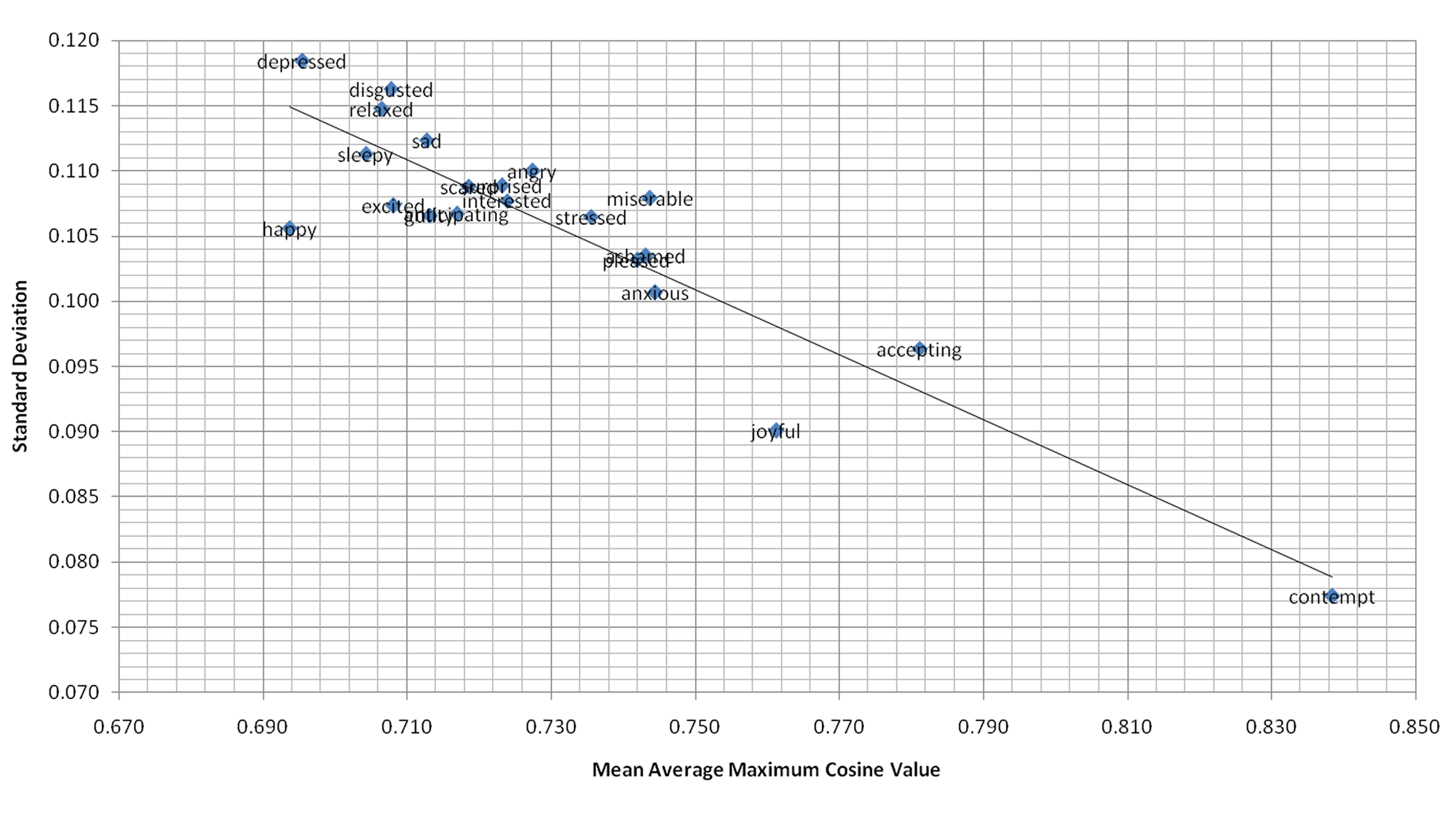}
\caption[Standard deviation correlated with maximum cosine values using ELSA]{Standard deviation plotted against the average maximum cosine value for each emotion of dimensions 10, 20, 30, 40, 50, 60, 70, 80, 90 and 100 using ELSA.}
\label{plot}
\end{figure*}
\end{landscape}

Plotting the standard deviation of ELSA1000 performances using a variety of dimensions against the average maximum cosine values creates an interesting correlation. By increasing the dimensionality, we increase the number of words that are included in the LSA space. Figure~\ref{plot} shows that the lower the standard deviation of data across all dimensions is, the more distinct the emotion is. This is actually intuitive as one would expect that the more variance an emotion has relative to the number of words that is used, the less accurate it would be in clustering.

\section{Testing Geographical Relativity}

We analysed subcorpora consisting of tweets originating from specific timezones in order to briefly assess whether there is any difference in the perception of emotion dimensions latent within the language of emotion expression across cultures. We select the following geographical areas (document limits in brackets): Pacific Time (US \& Canada) (77), Eastern Time (US \& Canada) (91), Central Time (US \& Canada) (92) and London, Edinburgh, Dublin (49). We designate these subcorpora the codes USP77, USE91, USC92 and UK49 receptively. The limits we tested each of these sets with were chosen as the minimum number of tweets any one emotion had for each timezone, in order for a fair test to be carried out; each timezone had consistent quantities of tweets for each emotion.

One concern was that such a small limit would render the analysis statistically irrelevant; however, we have mitigated this by showing a random ELSA calculation using a limit of 77, generated using a MOD 12 = 1 function on the index, which correlated to ELSA1000. We also recreated each ELSA set to the optimal. The results are shown in Table~\ref{ELSAGEO}.

Of all the current basic emotion theories Tomkin's set performs best, and Oatley's and Russell's sets perform worst overall, corresponding to ELSA1000. In most cases the ELSA set outperforms current basic emotion sets, although if the optimum is found an increase of up to 2.6\% is observed. While this alone somewhat diminishes the case for linguistic relativity, we found that in all cases the optimal set to be not only different to the best performing set, but in as much disarray as the original basic emotion sets as tabulated by \citeasnoun{basic} --- see Table~\ref{emocats}. As we have observed negligible changes in the optimal set when testing subcorpora of the same size, we must conclude that there is indeed linguistic relativity, at least with regards to emotion and geographical areas.

What we have determined is that the emotion \textit{disgusted} cannot be regarded as a semantically distinct emotion, as it does not appear in any emotion set derived in this research. This may come across as surprising, since all basic emotion theories analysed have included \textit{disgust} (with the exception of it not being one of Russell's categories of dimension), although recall in section~\ref{basic-section} we mentioned that \textit{disgust} is a `vague' emotion, with the possibility of explaining differing emotion qualia. In this sense, it is no surprise that is has been identified as an indistinct emotion.

\begin{table}
\centering
\scriptsize
\begin{tabular}{r|*{8}{c}r}
& \multicolumn{8}{c}{\textbf{Model}} \\
\hline
	&\textbf{Izard}	&\textbf{Russell}	&\textbf{Plutchik}	&\textbf{Ekman}	&\textbf{Tomkins}	&\textbf{Oatley}		 &\textbf{ELSA} &\textbf{Optimal}\\
\hline
USP77	&\textbf{0.605}	&0.569	&0.589	&0.594	&0.603	&0.590	&0.605	&0.621\\
USE91	&0.619	&0.601	&0.616	&0.615	&\textbf{0.622}	&0.596	&0.631	&0.636\\
USC92	&0.618	&0.610	&0.595	&0.599	&\textbf{0.624}	&0.583	&0.627	&0.637\\
UK49	&0.478	&0.476	&\textbf{0.484}	&0.480	&0.480	&0.459	&0.475	&0.501\\
RAND77	&0.593	&0.577	&0.584	&0.581	&\textbf{0.593}	&0.585	&0.601	&0.613\\
\hline
MEAN	&0.582	&0.567	&0.574	&0.574	&\textbf{0.584}	&0.563	&\textbf{0.588}	&\textbf{0.602}\\
\hline
STDEV	&0.060	&0.053	&0.052	&0.054	&0.060	&0.058	&0.065	&0.057\\
\end{tabular}
\caption[ELSA Geographical Analysis]{ELSA values for geographical subcorpora, using a dimension of 50.}
\label{ELSAGEO}
\end{table}

\section{Discussion}

A large majority of computer scientists tend to use Ekman's basic emotion set for emotion categorisation, and it appears that, semantically, it is the most distinct set. Using an iterative algorithm based on LSC, we have discerned a set of eight basic emotion keywords that have been calculated to be the most semantically distinct. This set performed better in all semantic tests than all of the basic emotion models analysed, with a 6.1\% increase in accuracy over Ekman's basic emotion set. Furthermore, the derived set satisfies the three laws of basic emotions (see Section~\ref{coreaffect}).

Emotions must be seen as relative to a specific domain. This research has not focused on any particular domain as it attempts to generalise the nature of emotions, however, as we have seen with the geographical testing, the composition of optimal sets are dynamic and highly dependent on region, even though negligible alterations were observed when testing a random sample --- if there is an underlying domain, then its semantically distinct set of emotion dimensions will differ to that of another corpus of a different domain. 

If we had to decide upon a universal set of basic emotions, it would be those emotions that appear thrice or greater in the derived basic emotion sets within the research thus far (ELSA, DELSAR and the optimal ELSA sets derived from USP77, USE91, USC92, UK49, RAND77):

\begin{center}
\textsc{accepting anxious ashamed contempt interested joyful pleased stressed}
\end{center}

We do not recommended the use of these emotion dimensions universally, although they are a much more accurate and complete set of basic emotions to base emotion classification with than existing basic emotion sets, and certainly more distinct than Ekman's set, at least in terms of the language used for each expression of emotion. Similar to \possessivecite{kim} point of view, we recommend that a different basic emotion set should be chosen dependent on the domain and corpus, suggesting that basic emotions are ultimately not universal and correlated with underlying thematic concerns within the corpus under analysis.

\begin{table}
\centering
\scriptsize
\begin{tabular}{r|*{8}{c}r}
& \multicolumn{8}{c}{\textbf{Basic Emotion Set}} \\
\hline
	&\textbf{USP77}	&\textbf{USE91}	&\textbf{USC92}	&\textbf{UK49}	&\textbf{RAND77}	&\textbf{ELSA}	&\textbf{DELSAR}	&\textbf{Category}\\
\hline
accepting	&	&x	&	&	&	&x	&x	&1\\
angry		&	&	&	&	&x	&	&	&3\\
anticipating	&	&	&	&x	&x	&	&	&2\\
anxious	&x	&x	&	&	&x	&x	&	&1\\
ashamed	&x	&	&x	&x	&	&x	&x	&1\\
contempt	&x	&x	&x	&	&x	&x	&x	&1\\
depressed	&	&	&x	&	&	&	&	&3\\
disgusted	&	&	&	&	&	&	&	&4\\
excited	&	&	&x	&x	&	&	&	&2\\
guilty		&x	&	&	&	&	&	&	&3\\
happy		&x	&	&	&	&	&	&	&3\\
interested	&	&x	&x	&x	&x	&	&x	&1\\
joyful		&x	&x	&x	&x	&x	&x	&x	&1\\
miserable	&	&	&	&	&	&x	&	&3\\
pleased	&	&x	&x	&x	&	&x	&x	&1\\
relaxed	&	&	&	&	&x	&	&	&3\\
sad		&x	&	&	&	&	&	&	&3\\
scared	&	&x	&	&	&x	&	&	&2\\
sleepy		&	&	&	&x	&	&	&x	&2\\
stressed	&x	&x	&x	&	&	&x	&x	&1\\
surprised	&	&	&	&x	&	&	&	&3\\
\end{tabular}
\caption[Optimal ELSA Emotion Sets]{Optimal ELSA emotion sets derived from various corpora.}
\label{emocats}
\end{table} 

We can define three more `categories' of emotions, according to the frequency of their use in derived basic emotion sets within the research thus far. Having just identified category~1 emotions, the next category consists of emotions that appear in multiple sets:

\begin{center}
\textsc{anticipating excited scared sleepy}
\end{center}

Category~2 emotions are also primary emotions, especially as some of them appear in our ELSA and DELSAR sets, although they have not been as strongly identified and thus not category~1 emotions. The next category refers to those emotions that could be considered as `secondary' emotions, similar in sense to that of Plutchik's definition:

\begin{center}
\textsc{angry depressed guilty happy miserable relaxed sad}
\end{center}

Secondary emotions, or category~3 emotions, only appear in one of the emotion sets derived in this research thus far and are the candidates for being the result of the fusion of multiple emotions, in addition to the final category of emotions:

\begin{center}
\textsc{disgusted}
\end{center}

Category~4 emotions are not basic emotions, and have not been included in any derived set thus far. We will be attempting to recreate secondary emotions as a result of fusing together permutations of the more distinct emotions identified in the next chapter. It should be noted that we have analysed a significantly smaller sample of the corpus for geographical testing, hence results are bound to be less accurate than if a larger sample was taken, as DELSAR and ELSA does.

\subsection{Fear}

Our basic set does not include \textit{fear} as a basic emotion. This will be sure to create much debate amongst researchers, as the majority of emotion research has identified \textit{fear} as a basic emotion; indeed, only five of the fourteen basic emotion sets as tabulated by \citeasnoun{basic} omitted \textit{fear}. There are two possible explanations as to why \textit{fear} may not be a basic emotion. The first, and the most obvious of the two, is that not enough synonymic emotion keywords, such as \textit{frightened}, \textit{afraid} and \textit{horrified}, were used in creating our Twitter corpus to be able to capture a complete signature of the emotion \textit{fear}. In our study, we have used the emotion keyword \textit{scared} due to its popularity of expression, however, this may have hidden underlying conceptualisations of the emotion \textit{fear}. For example, perhaps people use the word \textit{frightened} to describe the type of \textit{fear} experienced when, for example, facing a hungry lion about to attack them, and the word \textit{scared} to describe the type of \textit{fear} experienced when, for example, facing a break-up with their partner. Moreover, the vast majority of people are aware that their tweets are able to be read by the public, and we mentioned in Section~\ref{twitter-section} that we neglect public image considerations --- there may exist a tendency to omit certain emotions from the public. Would people really tweet true \textit{fear}?

The second explanation as to why \textit{fear} may not be a basic emotion, and arguably most controversial, is the proposition that \textit{fear} is in fact a compound emotion, comprised of an \textit{anxiety} component and a \textit{stress} component, both found to be more conceptually distinct than \textit{fear}. Although similar, \textit{fear} and \textit{anxiety} are said to be significantly distinct in duration and focus, specificity of the related threat and motivated direction of response. \textit{Fear} is defined as short lived, present focused, geared towards a specific threat, and facilitating escape from threat, while \textit{anxiety} is defined as long acting, future focused, broadly focused towards a diffuse threat, and promoting caution while approaching a potential threat \cite{fear}. However, we should attempt to distance ourselves from learned ideas and past experience of actions in response to these emotions. If we separate emotion from action, and focus on the expression, and thus the conceptualisation of emotions, we could say that these two emotions are similar. In this case, it is the emotional \textit{response} that is distinct, and not the emotion \textit{qualia}.

\citeasnoun{darwin-face} found that \textit{fear} had a characteristic facial expression, and coupled with its evolutionary role in responding to an immediate threat, posited \textit{fear} as a basic emotion. However, Darwin required \textit{fear} to be atomic (that is, as a single emotion as opposed to several mechanisms) in order to explain his theory, so, while the emotion that best describes the mechanism in Darwin's evolutionary theory may well be \textit{fear}, he did not need to consider the possibility that it could be a compound of several other emotional components. In \possessivecite{robots} work, an evaluation of the social robot Kismet's facial expressions found that \textit{fear} was the most ambiguous of all the basic emotions implemented, again casting doubt over its perceived distinctiveness compared to other basic emotions. Moreover, \citeasnoun{cannon} says that a \textit{stress} component, rather than \textit{fear}, is triggered in a `fight or flight' response. If this is compounded with a high \textit{anxiety} component, this could result in the manifestation of \textit{fear}. Whilst \textit{fear} may be facially distinct, it may not be as conceptually distinct as previously thought.

\subsection{Applications of LSC}

We have essentially ranked emotions and basic emotion sets using our algorithms according to a metric of semantic distinctiveness of extracted expressions from Twitter. Whilst this is useful for discovering psychological similarities of emotion expression, the real power lies in the application of the resulting clustering data. By visualising these matrices, we are able to compare the similarity of compound emotions, analyse the composite properties of emotions and highlight how specific emotions interact with each other.


\chapter{Visualising Emotion}\label{vis}

Until now, we have not analysed any relationships between specific emotions that contribute to defining their similarity, such as the relative levels of polarity and engagement. This is important as it could highlight specifically how emotions interact with each other, and easily analyse, relative to the corpus, the composite properties of each emotion. Traditionally, this has been illustrated using core affect models (see Section~\ref{coreaffect}) in which emotions are positioned in an emotional space as a single point (Russell's Theory of Affect), or designated a specific dimension (Watson's Circumplex Theory of Affect). Indeed Plutchik not only assigns emotions to bipolar dimensions, but does so using each dimension as a scale of intensity, and also incorporating mixtures of dimensions into his model. The key to illustrating emotions is, when calculating DELSAR clustering, instead of registering clustering hits and misses to obtain a final accuracy, we actually record the specific cluster labels of each miss, resulting in each emotion being a combination of all emotions. We visualise these semantic vectors of emotion in this chapter using two techniques: \textit{Multidimensional Scaling} and \textit{Emotion Profiling}.

\section{Emotion Space}

\begin{landscape}
\begin{figure*}
\centering
\includegraphics[width=555px]{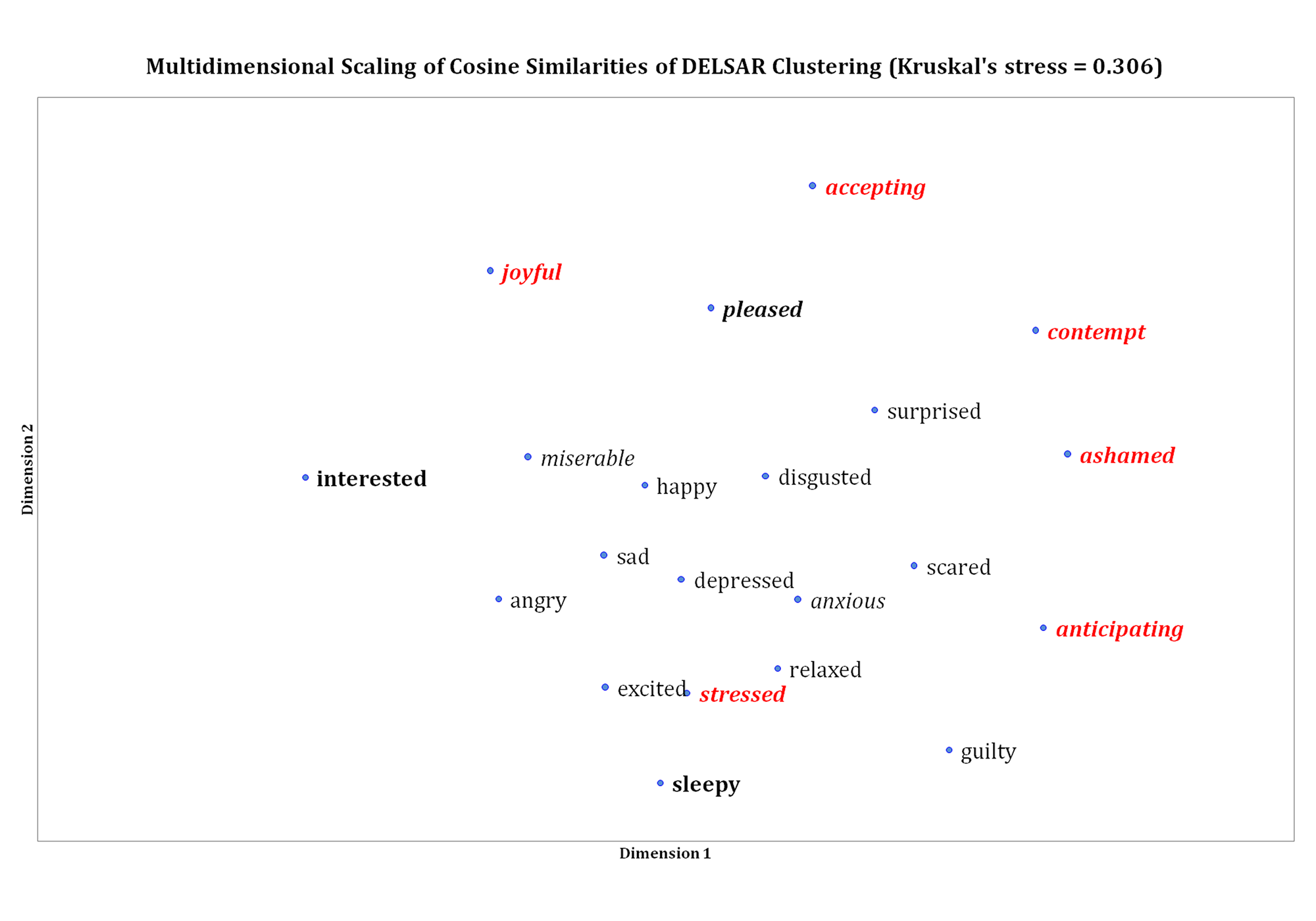}
\caption[Multidimensional Scaling of DELSAR1100 Clustering Emotion Vector Cosine Similarities]{Multidimensional Scaling of DELSAR1100 Clustering Emotion Vector Cosine Similarities --- red denotes our identified primary emotions; bold denotes our DELSAR set and italics denotes our ELSA set.}
\label{emospace}
\end{figure*}
\end{landscape}

\begin{figure*}
\centering
\includegraphics[height=360px]{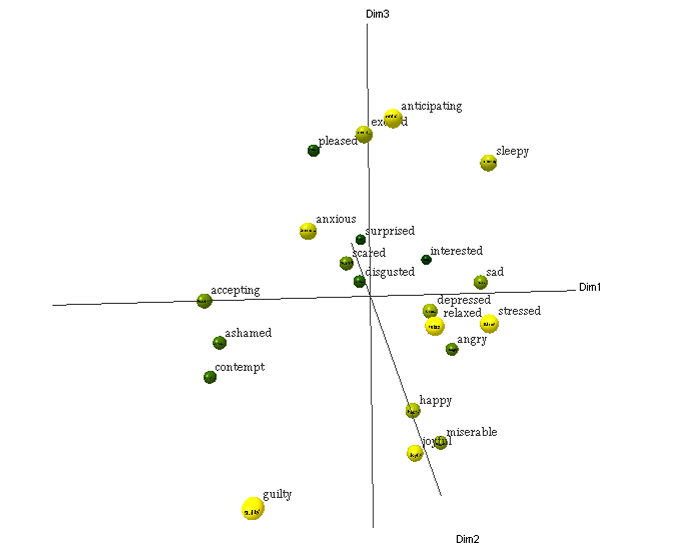}
\caption[Multidimensional Scaling of DELSAR1100 Clustering Emotion Vector Cosine Similarities --- 3D Visualisation]{Multidimensional Scaling of DELSAR1100 Clustering Emotion Vector Cosine Similarities --- 3D Visualisation. Yellow to Green indicates the depth of dimension 2.}
\label{xlstat}
\end{figure*}

Firstly, we visualise what a two and three-dimensional emotion space looks like. We create an emotion matrix of DELSAR1100 clustering, shown in Table~\ref{delsar-clust}, using 1100 documents for each emotion. We can see that this matrix illustrates the notion that each emotion is in fact a combination of all other emotions. This property has also been shown in \possessivecite{russell} and more recently \possessivecite{robots} work. The Cosine Similarity is then calculated for each pairwise semantic emotion vector using Equation~\ref{cosine-sim}, creating a symmetrical similarity matrix of cosine similarities for all emotions, shown in Table~\ref{delsar-cosine}. We use the XLSTAT Excel plug-in to perform multidimensional scaling of these cosine similarities, and plot a two-dimensional representation of this space, shown in Figure~\ref{emospace}, and a three-dimensional representation, shown in Figure~\ref{xlstat}.

\begin{landscape}
\begin{table}
\centering
\tiny
\begin{tabular}{r|*{21}{c}r}
& \multicolumn{11}{c}{\textbf{Emotion}} \\
\hline
	&accepting	&angry	&anticipating	&anxious	&ashamed	&contempt	&depressed	&disgusted	&excited	&guilty	&happy\\
\hline
accepting	&471	&19	&21	&13	&19	&62	&14	&20	&11	&24	&28	\\
angry		&29	&300	&24	&45	&36	&27	&74	&51	&20	&26	&50	\\
anticipating	&24	&18	&325	&31	&13	&16	&21	&11	&33	&8	&13	\\
anxious	&32	&57	&64	&262	&42	&55	&56	&56	&110	&34	&60	\\
ashamed	&39	&32	&24	&84	&398	&171	&23	&44	&28	&39	&42	\\
contempt	&52	&14	&31	&27	&23	&362	&17	&33	&22	&39	&17	\\
depressed	&41	&99	&58	&61	&57	&36	&199	&79	&63	&63	&93	\\
disgusted	&46	&67	&44	&39	&59	&44	&64	&241	&43	&60	&50	\\
excited	&22	&17	&59	&71	&19	&19	&33	&18	&261	&16	&25	\\
guilty		&12	&23	&20	&36	&50	&28	&40	&48	&25	&372	&27	\\
happy		&24	&40	&27	&31	&34	&16	&70	&19	&32	&26	&246	\\
interested	&41	&28	&36	&39	&21	&40	&34	&44	&29	&26	&21	\\
joyful		&30	&17	&28	&24	&22	&25	&27	&19	&27	&24	&28	\\
miserable	&32	&48	&33	&25	&28	&16	&47	&48	&22	&33	&68	\\
pleased	&61	&33	&50	&29	&37	&44	&25	&83	&54	&30	&45	\\
relaxed	&30	&52	&57	&57	&40	&38	&65	&52	&69	&79	&51	\\
sad		&28	&63	&31	&42	&33	&28	&64	&47	&35	&37	&56	\\
scared	&18	&40	&26	&25	&32	&13	&23	&29	&29	&36	&30	\\
sleepy		&20	&50	&53	&78	&42	&14	&86	&55	&99	&42	&62	\\
stressed	&21	&45	&45	&53	&30	&19	&69	&33	&45	&45	&38	\\
surprised	&27	&38	&44	&28	&65	&27	&49	&70	&43	&41	&50	\\
\hline
Theoretical Positivity	&0.24	&0.22	&0.57	&0.28	&0.23	&0.20	&0.29	&0.29	&0.50	&0.23	&0.44	\\
\hline
Vector Magnitude	&494.38	&361.34	&373.51	&333.17	&432.38	&425.29	&299.13	&320.71	&340.64	&412.33	&323.15	\\
\end{tabular}
\end{table} 
\begin{table}
\centering
\tiny
\begin{tabular}{r|*{21}{c}r}
& \multicolumn{10}{c}{\textbf{Emotion}} \\
\hline
		&interested	&joyful	&miserable	&pleased	&relaxed	&sad	&scared	&sleepy	&stressed	&surprised\\
\hline
accepting		&10	&43	&25	&23	&21	&20	&24	&7	&9	&9\\
angry			&32	&23	&46	&35	&31	&43	&46	&35	&54	&33\\
anticipating		&11	&11	&18	&26	&33	&22	&17	&12	&24	&21\\
anxious		&41	&30	&49	&60	&52	&54	&44	&66	&60	&36\\
ashamed		&28	&23	&33	&31	&27	&31	&59	&25	&22	&50\\
contempt		&36	&37	&27	&26	&23	&30	&17	&5	&16	&18\\
depressed		&38	&46	&85	&44	&68	&91	&71	&89	&114	&62\\
disgusted		&54	&32	&50	&77	&63	&59	&59	&33	&38	&80\\
excited		&19	&32	&12	&28	&21	&43	&45	&36	&48	&29\\
guilty			&16	&28	&28	&21	&55	&28	&34	&24	&31	&20\\
happy			&18	&60	&61	&36	&43	&57	&29	&33	&39	&35\\
interested		&489	&22	&31	&32	&42	&32	&25	&18	&25	&46\\
joyful			&24	&432	&37	&31	&32	&23	&17	&11	&10	&22\\
miserable		&29	&48	&300	&22	&31	&44	&33	&22	&47	&35\\
pleased		&44	&59	&22	&387	&32	&43	&30	&28	&21	&64\\
relaxed		&52	&61	&62	&54	&280	&52	&49	&71	&77	&48\\
sad			&40	&38	&56	&30	&55	&258	&27	&46	&48	&40\\
scared		&28	&11	&35	&25	&24	&21	&269	&36	&28	&30\\
sleepy			&27	&18	&44	&36	&79	&64	&92	&407	&84	&50\\
stressed		&19	&19	&45	&27	&62	&44	&39	&54	&274	&25\\
surprised		&45	&27	&34	&49	&26	&41	&74	&42	&31	&347\\
\hline
Theoretical Positivity		&0.64	&0.64	&0.25	&0.58	&0.46	&0.28	&0.26	&0.23	&0.25	&0.56\\
\hline
Vector Magnitude		&510.87	&461.99	&357.48	&423.27	&343.36	&328.50	&339.24	&445.97	&350.18	&393.32\\
\end{tabular}
\caption[DELSAR1100 Clustering Matrix]{DELSAR1100 Clustering Matrix.}
\label{delsar-clust}
\end{table} 
\end{landscape}

\begin{landscape}
\begin{table}
\centering
\tiny
\begin{tabular}{r|*{21}{c}r}
& \multicolumn{11}{c}{\textbf{Emotion}} \\
\hline
	&accepting	&angry	&anticipating	&anxious	&ashamed	&contempt	&depressed	&disgusted	&excited	&guilty	&happy\\
\hline
accepting	&1	&0.240	&0.229	&0.242	&0.214	&0.340	&0.275	&0.304	&0.222	&0.193	&0.288\\
angry	&0.240	&1	&0.345	&0.514	&0.347	&0.273	&0.716	&0.582	&0.401	&0.332	&0.549\\
anticipating	&0.229	&0.345	&1	&0.477	&0.261	&0.272	&0.478	&0.410	&0.490	&0.261	&0.384\\
anxious	&0.242	&0.514	&0.477	&1	&0.497	&0.411	&0.646	&0.556	&0.715	&0.390	&0.552\\
ashamed	&0.214	&0.347	&0.261	&0.497	&1	&0.529	&0.410	&0.456	&0.326	&0.340	&0.408\\
contempt	&0.340	&0.273	&0.272	&0.411	&0.529	&1	&0.326	&0.399	&0.296	&0.294	&0.306\\
depressed	&0.275	&0.716	&0.478	&0.646	&0.410	&0.326	&1	&0.707	&0.607	&0.490	&0.761\\
disgusted	&0.304	&0.582	&0.410	&0.556	&0.456	&0.399	&0.707	&1	&0.503	&0.485	&0.556\\
excited	&0.222	&0.401	&0.490	&0.715	&0.326	&0.296	&0.607	&0.503	&1	&0.335	&0.496\\
guilty	&0.193	&0.332	&0.261	&0.390	&0.340	&0.294	&0.490	&0.485	&0.335	&1	&0.379\\
happy	&0.288	&0.549	&0.384	&0.552	&0.408	&0.306	&0.761	&0.556	&0.496	&0.379	&1\\
interested	&0.178	&0.269	&0.242	&0.319	&0.201	&0.251	&0.340	&0.377	&0.270	&0.206	&0.261\\
joyful	&0.238	&0.253	&0.244	&0.290	&0.211	&0.243	&0.361	&0.308	&0.302	&0.242	&0.371\\
miserable	&0.259	&0.497	&0.356	&0.453	&0.331	&0.284	&0.633	&0.533	&0.384	&0.354	&0.615\\
pleased	&0.268	&0.353	&0.355	&0.394	&0.290	&0.291	&0.410	&0.580	&0.419	&0.272	&0.420\\
relaxed	&0.253	&0.481	&0.454	&0.557	&0.350	&0.316	&0.680	&0.584	&0.533	&0.504	&0.552\\
sad	&0.267	&0.557	&0.410	&0.558	&0.368	&0.337	&0.736	&0.604	&0.531	&0.389	&0.631\\
scared	&0.247	&0.495	&0.373	&0.503	&0.432	&0.295	&0.591	&0.552	&0.509	&0.391	&0.501\\
sleepy	&0.149	&0.398	&0.332	&0.523	&0.278	&0.172	&0.625	&0.439	&0.541	&0.297	&0.464\\
stressed	&0.204	&0.535	&0.424	&0.579	&0.322	&0.257	&0.760	&0.512	&0.552	&0.396	&0.546\\
surprised	&0.202	&0.399	&0.357	&0.399	&0.405	&0.274	&0.535	&0.601	&0.422	&0.316	&0.474\\
\hline
\end{tabular}
\end{table} 
\begin{table}
\centering
\tiny
\begin{tabular}{r|*{21}{c}r}
& \multicolumn{10}{c}{\textbf{Emotion}} \\
\hline
		&interested	&joyful	&miserable	&pleased	&relaxed	&sad	&scared	&sleepy	&stressed	&surprised\\
\hline
accepting	&0.178	&0.238	&0.259	&0.268	&0.253	&0.267	&0.247	&0.149	&0.204	&0.202\\
angry	&0.269	&0.253	&0.497	&0.353	&0.481	&0.557	&0.495	&0.398	&0.535	&0.399\\
anticipating	&0.242	&0.244	&0.356	&0.355	&0.454	&0.410	&0.373	&0.332	&0.424	&0.357\\
anxious	&0.319	&0.290	&0.453	&0.394	&0.557	&0.558	&0.503	&0.523	&0.579	&0.399\\
ashamed	&0.201	&0.211	&0.331	&0.290	&0.350	&0.368	&0.432	&0.278	&0.322	&0.405\\
contempt	&0.251	&0.243	&0.284	&0.291	&0.316	&0.337	&0.295	&0.172	&0.257	&0.274\\
depressed	&0.340	&0.361	&0.633	&0.410	&0.680	&0.736	&0.591	&0.625	&0.760	&0.535\\
disgusted	&0.377	&0.308	&0.533	&0.580	&0.584	&0.604	&0.552	&0.439	&0.512	&0.601\\
excited	&0.270	&0.302	&0.384	&0.419	&0.533	&0.531	&0.509	&0.541	&0.552	&0.422\\
guilty	&0.206	&0.242	&0.354	&0.272	&0.504	&0.389	&0.391	&0.297	&0.396	&0.316\\
happy	&0.261	&0.371	&0.615	&0.420	&0.552	&0.631	&0.501	&0.464	&0.546	&0.474\\
interested	&1	&0.179	&0.270	&0.258	&0.335	&0.312	&0.268	&0.189	&0.249	&0.310\\
joyful	&0.179	&1	&0.346	&0.301	&0.352	&0.321	&0.239	&0.178	&0.239	&0.246\\
miserable	&0.270	&0.346	&1	&0.302	&0.496	&0.540	&0.446	&0.356	&0.510	&0.383\\
pleased	&0.258	&0.301	&0.302	&1	&0.390	&0.402	&0.348	&0.277	&0.319	&0.416\\
relaxed	&0.335	&0.352	&0.496	&0.390	&1	&0.572	&0.483	&0.523	&0.616	&0.405\\
sad	&0.312	&0.321	&0.540	&0.402	&0.572	&1	&0.464	&0.478	&0.564	&0.453\\
scared	&0.268	&0.239	&0.446	&0.348	&0.483	&0.464	&1	&0.511	&0.491	&0.498\\
sleepy	&0.189	&0.178	&0.356	&0.277	&0.523	&0.478	&0.511	&1	&0.532	&0.355\\
stressed	&0.249	&0.239	&0.510	&0.319	&0.616	&0.564	&0.491	&0.532	&1	&0.378\\
surprised	&0.310	&0.246	&0.383	&0.416	&0.405	&0.453	&0.498	&0.355	&0.378	&1\\
\hline
\end{tabular}
\caption[Symmetrical Matrix of DELSAR1100 Clustering Emotion Vector Cosine Similarities]{Symmetrical Matrix of DELSAR1100 Clustering Emotion Vector Cosine Similarities.}
\label{delsar-cosine}
\end{table} 
\end{landscape}


We can clearly see that our identified basic emotions are in fact distributed nicely around the outside of the space, while other emotions are located in the middle of the space, which correlates to our basic emotion set being the most representative of distinct dimensions of emotion space. The three-dimensional plot shows the same space with a reduction of stress, or in other words, a more accurate representation of the similarity distance between emotions; clusters such as [\textsc{surprised}, \textsc{scared}, \textsc{disgusted}], [\textsc{accepting}, \textsc{ashamed}, \textsc{contempt}] and [\textsc{happy}, \textsc{joyful}, \textsc{miserable}] are clearly identified. Whilst visualising emotion space can be beneficial to illustrate the relationship between a set of emotions and help identify such clusters, these models do not illustrate the nature of individual emotions, but rather their relativity to other emotions. Having derived the dimensions of the specific emotion spectrum of the corpus we have analysed, we propose a novel technique to visualising emotions called an \textit{Emotion Profile}, in which an emotion is plotted as an area contained within a radar diagram, similar to circumplex models of affect.

\section{Emotion Profiling}

\begin{figure*}
\centering
\includegraphics{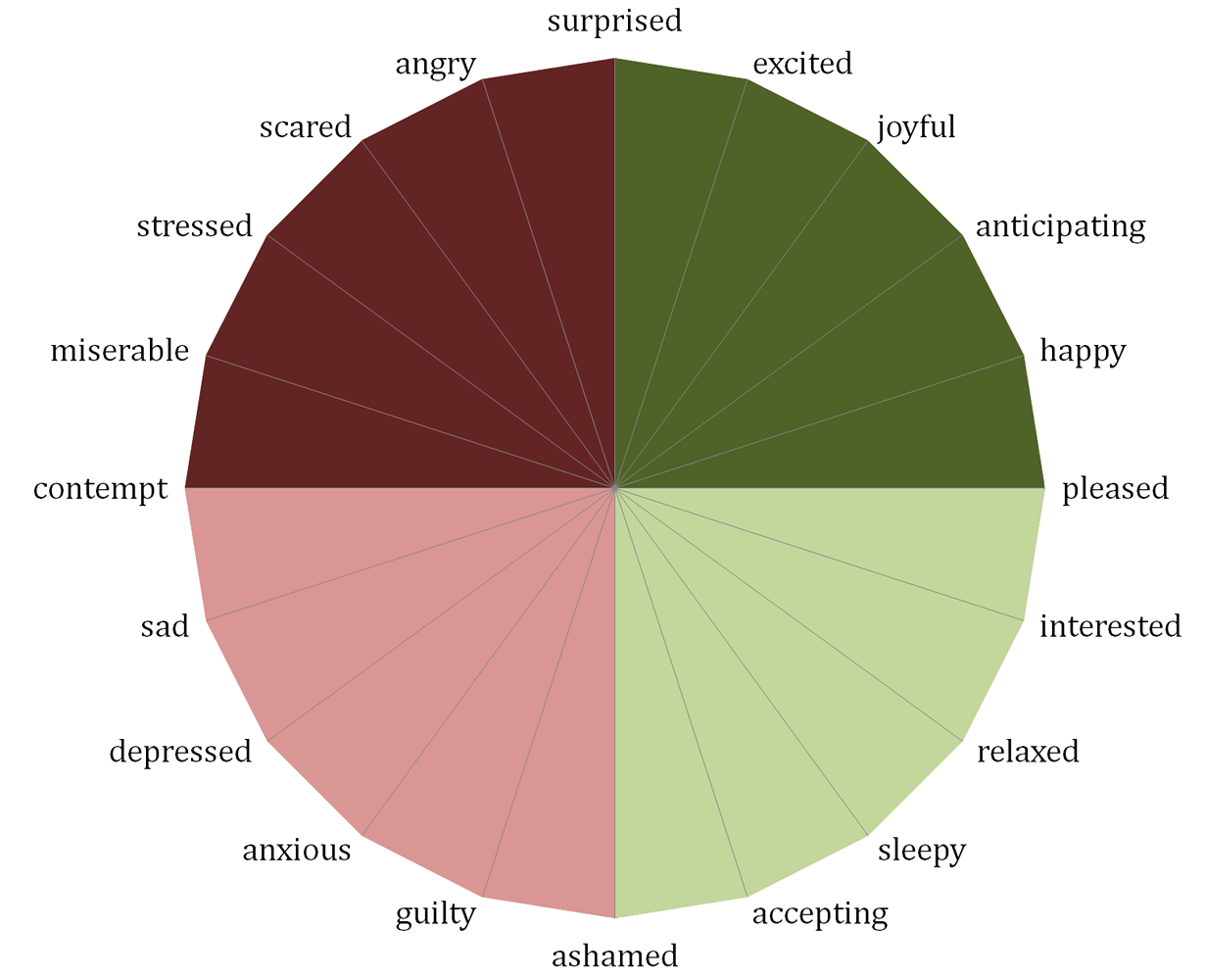}
\caption[Emotion Profile Space]{Emotion Profile Space --- red (left half) denotes negative emotions, green (right half) denotes positive emotions, dark denotes high arousal and light denotes low arousal.}
\label{circumplex-shades}
\end{figure*}

Typically, emotions are represented as single points within some space, as shown in the previous section. We can illustrate the nature of emotions by using an \textit{Emotion Profile}: a plot modelling each emotion's semantic ratio using a radar diagram. Circumplex properties are defined by the areas of the profile in specific segments of the diagram; each octant, quadrant and half has a specific property. To effectively represent core affect properties, we need to create the space theoretically based on Russell's circular scaling of emotions to position all 21 emotions, with the exception of \textit{disgusted} to preserve octal segmentation (chosen for being the least distinct emotion), in order around a circumplex, according to the two dimensional axis of polarity (valence) and engagement (arousal). Figure~\ref{circumplex-shades} shows the properties of specific regions of this emotion space according to each emotion's theoretical position around the circumplex, adapted from Russell's CMA. 

Generally speaking, the primary benefit of a graph is to quickly identify correlations and anomalies. Plotting emotion profiles provides us with a graphic visualisation of emotion, so we can easily identify key differences in the composition of emotions. We could represent this information numerically of course; the theoretical positivity of each emotion is calculated and appended to Table~\ref{delsar-clust} to illustrate this, and a similar measure can be obtained for arousal levels. Where the emotion element is equal to the emotion being profiled, the average of adjacent cells is plotted to avoid skewing.

Figure~\ref{circumplex-depressed} shows the emotion profiles of the emotions \textit{depressed}, \textit{anxious}, \textit{relaxed} and \textit{stressed}. We can see that all four emotions are more negative than positive (even with \textit{sleepy} as a positive emotion) and that \textit{relaxed} is clearly a disengaged emotion. They are all have a similar shape of profile, or in other words, people talk about these emotions in much the same way. This is also highlighted by their closeness in Figure~\ref{xlstat}.

What can these profiles tell us about these emotions? The key to interpreting emotion profiles, apart from the similarity of profile shape, is identifying anomalous spikes. We can see that the profile of \textit{anxious} has a spike in the dimension of \textit{ashamed}; if one is uses similar language as used by people talking about the emotion \textit{ashamed} then they are more likely to feel \textit{anxious} as opposed to any of the other three emotions. It could also be said that \textit{depressed} people are more likely than those who are \textit{anxious} or \textit{stressed} to `hide' their emotion by using language similar to those expressing \textit{happiness}. We can also see that \textit{stressed} people talk similarly to those speaking about \textit{depression}; based on their language, this is possibly an indication that these people are more likely to become \textit{depressed}.

\begin{landscape}
\begin{figure*}
\centering
\includegraphics[width=555px]{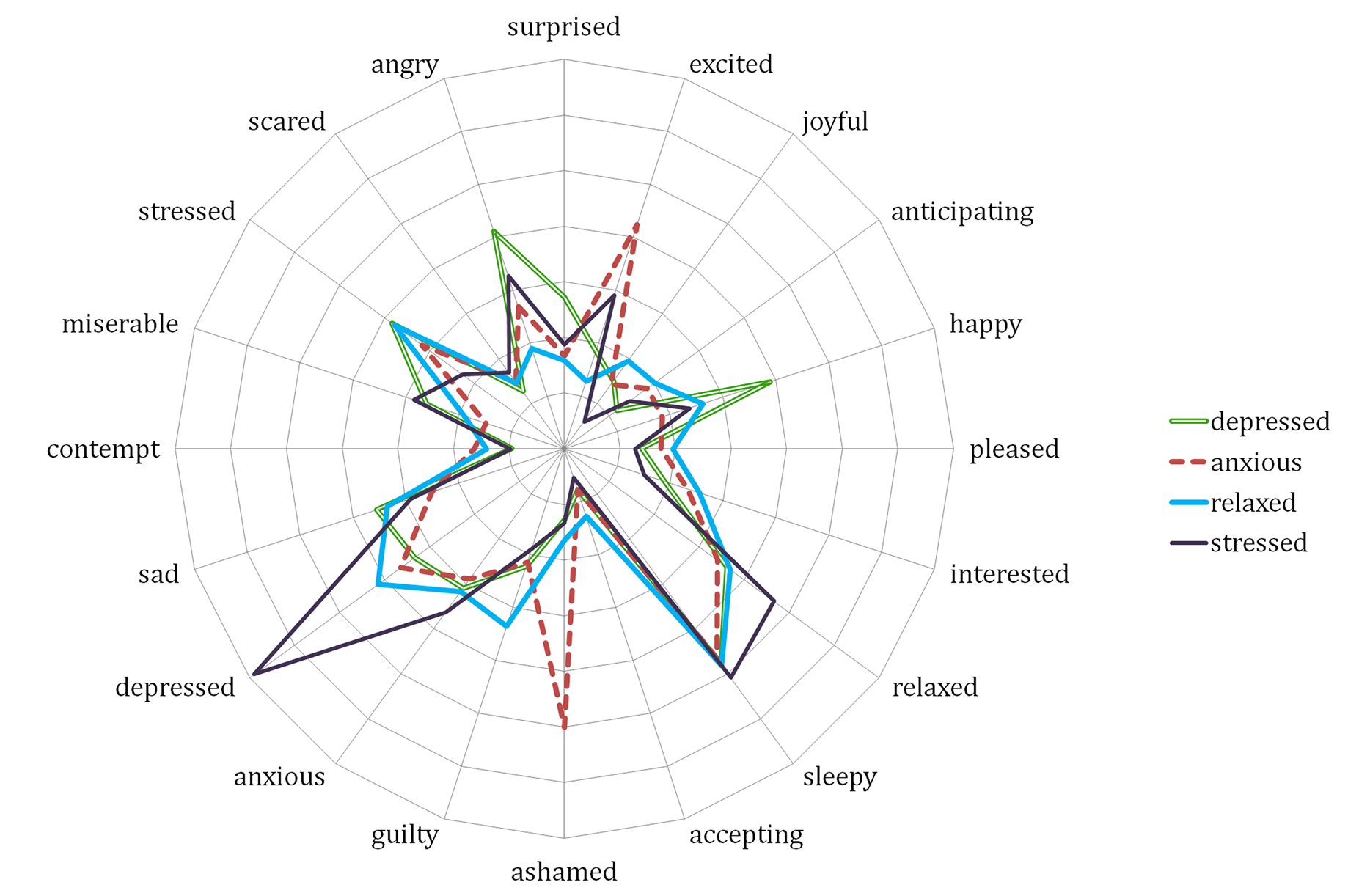}
\caption[Emotion Profile of Depressed, Anxious, Relaxed and Stressed]{Emotion Profile of Depressed, Anxious, Relaxed and Stressed.}
\label{circumplex-depressed}
\end{figure*}

\begin{figure*}
\centering
\includegraphics[width=555px]{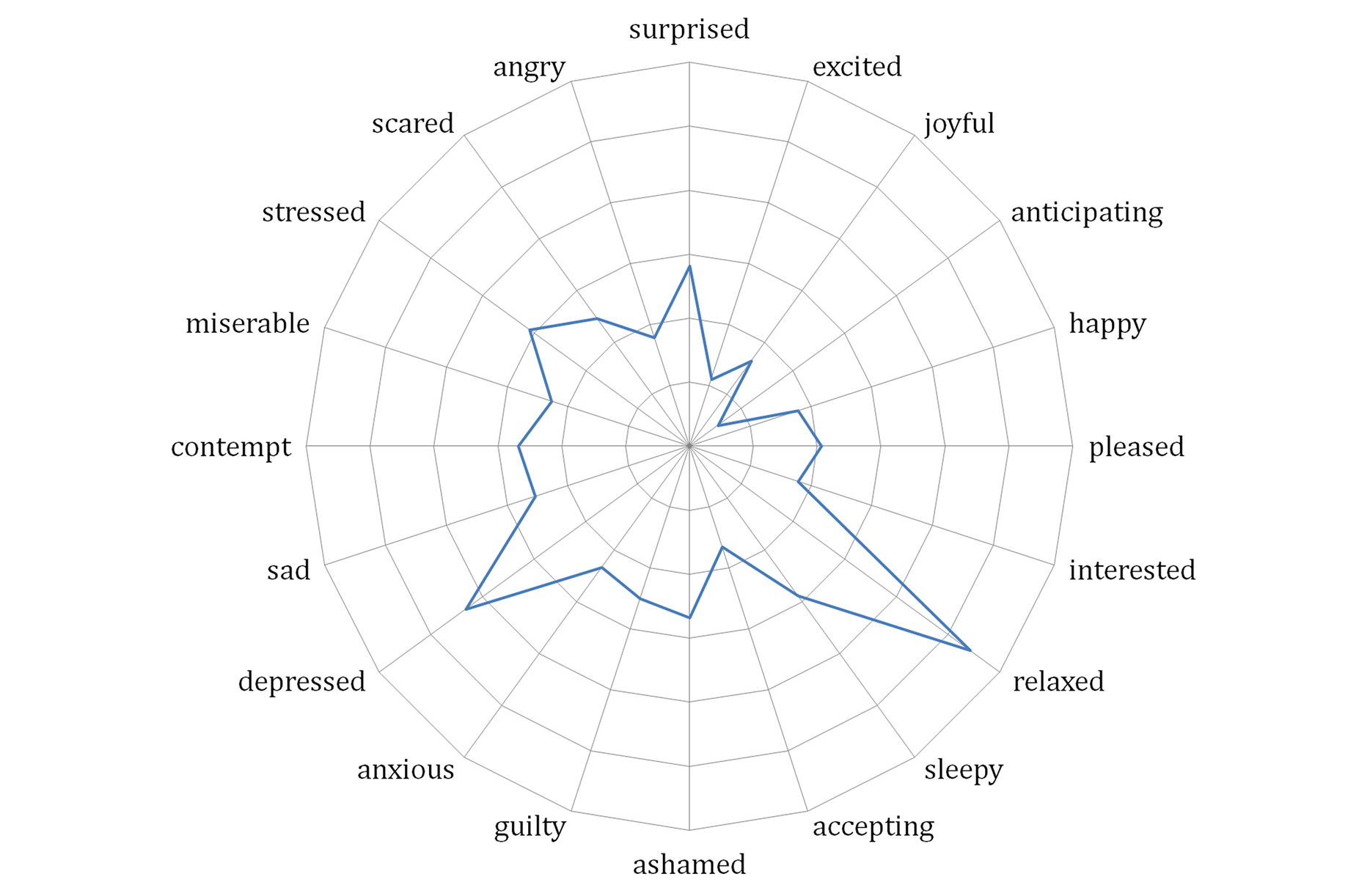}
\caption[Emotion Profile of Guilty]{Emotion Profile of Guilty.}
\label{guilty-circ}
\end{figure*}
\end{landscape}

We also show the emotion profile of \textit{guilty} in Figure~\ref{guilty-circ}. We can see that this is a more disengaged emotion, and more negative than positive, based on the properties of the emotion profile's area as determined by Figure~\ref{circumplex-shades}. The spike in the \textit{relaxed} dimension means that people are more likely to use similar language to those expressing the emotion \textit{relaxed} when talking about \textit{guilt}, possibly because it is a disengaged feeling, or due to a self-defence mechanism that reduces the impact of feeling \textit{guilty}. This implies that, compared to other emotions, \textit{guilt} is somewhat harder to identify through latent semantics. It also seems that \textit{depression} and \textit{guilt} are linked.

It must be remembered that all the documents of expressed emotions used for our analysis do not describe what people understand each emotion to mean; it merely represents the natural usage of these emotion keywords in language.

\subsection{Emotion Wave}

Specifically for temporal data visualisation, using an array of emotion profiles to model temporal data generates a signal, or wave form, of emotion that can be used to measure the conceptualisation of emotions over time. Instead of a graph that shows the peak and troughs of a variable in a single dimension, using a time-lapse of emotion profiles enables us to visualise how the spikes of emotion profiles morph over time, identifying volatility and trends in the properties of each dimension and area covered by the profile.

\section{Emotion Equations}\label{equations}

In this section we briefly introduce the concept of emotion equations. The objective is to see if we can create a selection of the secondary emotions identified in Chapter~\ref{sea} by combining particular primary emotions. For this, we again use DELSAR1100 clustering data (see Table~\ref{delsar-clust}) as our emotion vectors. While Plutchik theorises on which combinations of emotions create another emotion, we attempt to provide the first significant empirical evidence that emotions can in fact be combined to create other emotions, to some extent.

\subsection{Method}

Each primary emotion vector is firstly zero-weighted --- we set the element $V_{e}$ in the vector of emotion $e$ to 0 --- and then normalised by dividing each element by the sum of the vector's elements. We then combine each of the emotions by adding together their vectors to create 78 pairwise emotion combination vectors. For each emotion combination vector we calculate the cosine similarity between it and each of the secondary emotion vectors, zero-weighted and normalised in the same way, using Equation~\ref{cosine-sim}.

\subsection{Analysis}

Preliminary results show that some emotions are more accurately represented as a combination of two emotions rather than the most similar emotion. We present graphs for three of the most interesting secondary emotions: \textit{depressed}, \textit{disgusted} and \textit{guilty}.

\begin{landscape}
\begin{figure*}
\centering
\includegraphics[width=555px]{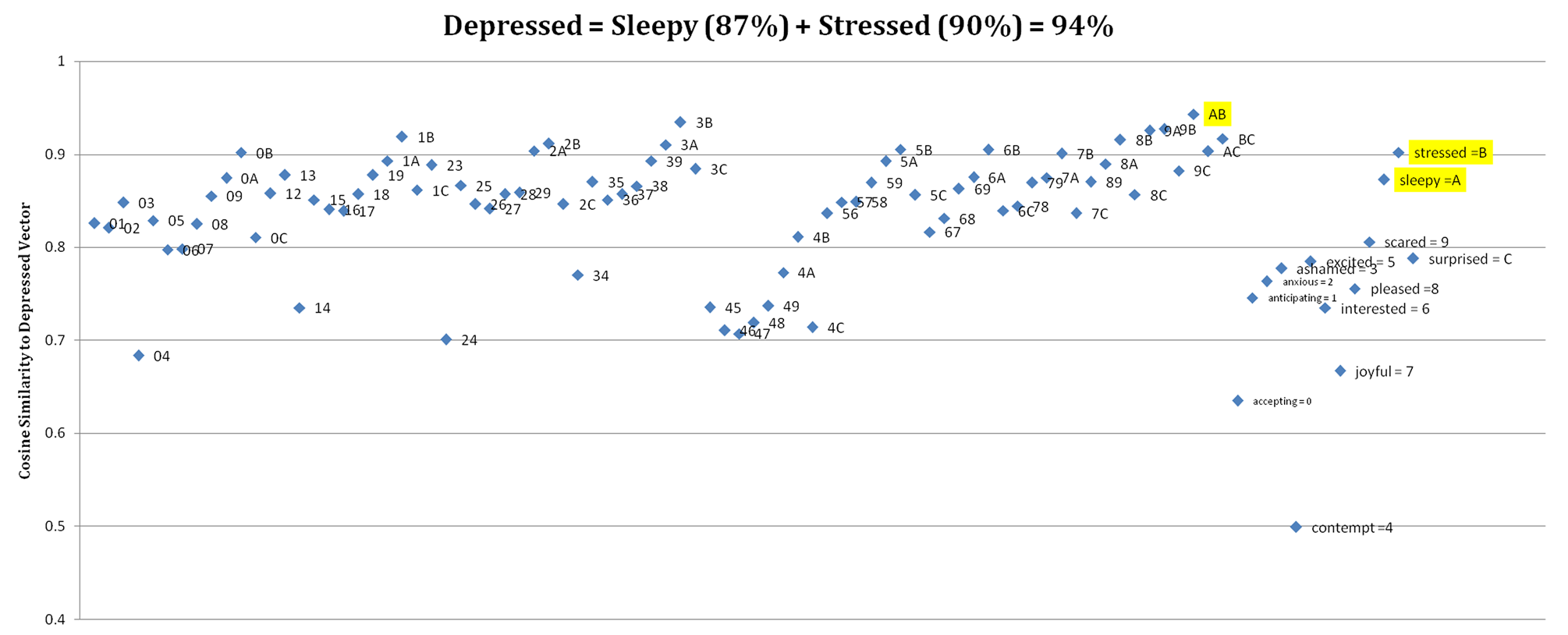}
\caption[Similarity of primary emotion combinations to \textit{depressed}]{Cosine similarity of primary emotion combinations to \textit{depressed}.}
\label{depressed}
\end{figure*}

\begin{figure*}
\centering
\includegraphics[width=555px]{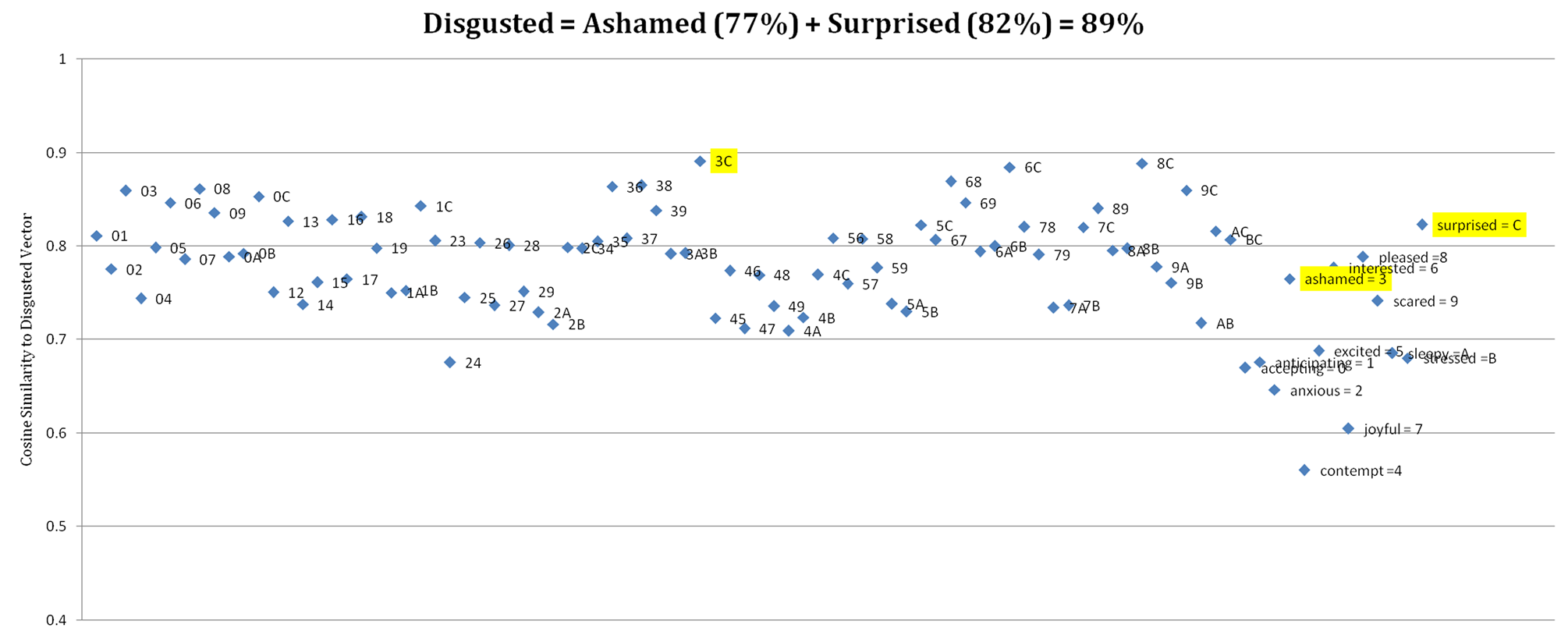}
\caption[Similarity of primary emotion combinations to \textit{disgusted}]{Cosine similarity of primary emotion combinations to \textit{disgusted}.}
\label{disgusted}
\end{figure*}

\begin{figure*}
\centering
\includegraphics[width=555px]{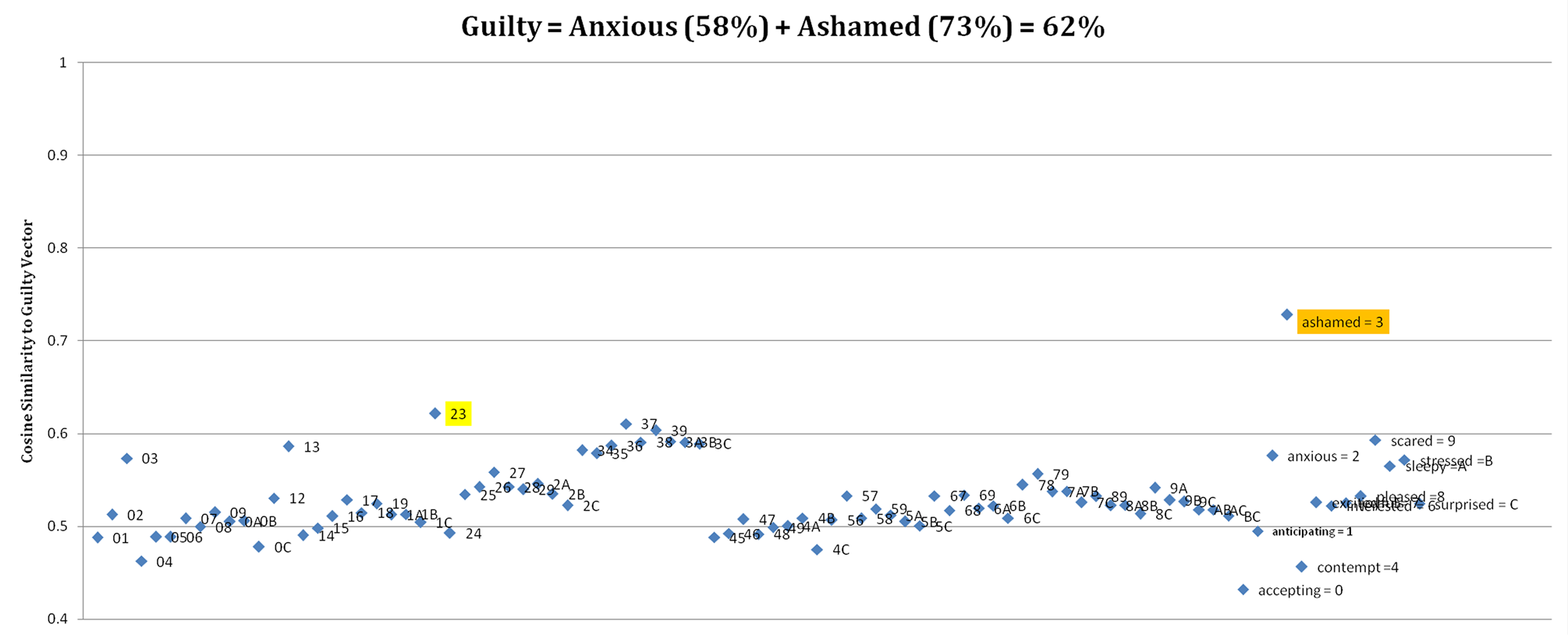}
\caption[Similarity of primary emotion combinations to \textit{guilty}]{Cosine similarity of primary emotion combinations to \textit{guilty}.}
\label{guilty}
\end{figure*}
\end{landscape}

\subsubsection{Depressed}

The similarity of emotion combination vectors to the emotion vector of \textit{depressed} is shown in Figure~\ref{depressed}. We can see that overall, emotion combinations that include the emotion \textit{stressed} are most similar to \textit{depressed}. Indeed, the emotion vector for \textit{stressed} is the most similar to \textit{depressed} out of all the secondary emotions; however, by combining it with the second most similar emotion \textit{sleepy}, we increase the similarity from an average of 88\% to 94\%. This is important to consider, as we identify other emotional factors contributing to \textit{depression} other than the primary `cause' of the emotion, this being \textit{stress}.

\subsubsection{Disgusted}

The similarity of emotion combination vectors to the emotion vector of \textit{disgusted} is shown in Figure~\ref{disgusted}. Again, we can see the most similar emotion (this time being \textit{surprised}) is dominant in all the most similar emotion combination vectors, and by combining it with the emotion \textit{ashamed} we increase the similarity from an average of 79\% to 89\%. We should note that the emotions \textit{pleased} and \textit{interested} are also similar, however not only do their combinations score fractionally less than \textit{ashamed}, it doesn't seem psychologically correct to use them as part of the emotion \textit{disgust}. Their similarity is due to the similarity of the language people use, so we could say that people using similar language as those expressing the emotions \textit{pleased} and \textit{interested} are, statistically speaking, more likely to feel \textit{disgusted}.

\subsubsection{Guilty}

The similarity of emotion combination vectors to the emotion vector of \textit{guilty} is shown in Figure~\ref{guilty}. This is an example where the most similar emotion of a particular emotion is a single emotion rather than a combination of emotions, namely \textit{ashamed}, which is 11\% more similar to \textit{guilty} than the most similar emotion combination vector. This provides evidence that \textit{guilty} is more likely to be a basic emotion than other secondary emotions we have analysed.

The similarity of emotion combination vectors to the emotion vector of \textit{happy} can be seen in Chapter~\ref{more-figures}.

\section{Discussion}

Emotions are notoriously difficult to visualise. The predominant method for showing how emotions change over time is by measuring the quantity of expressions at each time point without any regard for dimensionality, possibly due to a lack of post-processing of collected emotion data. Sentiment is usually visualised in this way, however, when applied to emotions, each emotion is determined as being either positive or negative. In our model, we propose that emotions are a ratio of both positive and negative, and both engaged and disengaged properties that morph according to time, relative to the particular topic or corpus being analysed. Emotion waves and profiles can be used as signatures in which pattern matching techniques can be applied for predictive analysis over a time series. Using these signatures as definitions of emotions can generate a more accurate analysis of emotion or mood (for example, calmness) compared to more theoretical definitions.

Emotion equations is an area which we will be studying at greater depth. We have only taken combinations of secondary emotions and a more concentrated study will involve combining a much greater number of both primary and secondary emotions to create axiomatic equations of emotions. Other basic operators will also be investigated, such as taking away emotions and adding varying intensities of emotions, for example adding a \textit{fraction} of one emotion to another emotion. Having said that, we have demonstrated that both our clustering algorithm and method of combining emotions provide a fairly accurate meaning to certain emotions, which is somewhat impressive seeing as we have only analysed 1100 documents extracted from a micro-blogging site with little parsing of the data nor strict constraints to data collection. Ideally, we would have liked to test Plutchik's theory of emotion combinations, however we do not currently hold the data for his specific emotion keywords, and therefore have presented our method and analysis for a few other basic emotions.

\subsection{Emotion Classification and Discovery}

The most obvious application of this research is to the mining, classification and identification of emotions in text. As we have discussed in Chapter~\ref{emo-extract}, many computer science studies of emotion include the use of a set of basic emotions, and Ekman's set is most commonly used. However, we should not assume that Ekman's basic set of emotions is the most ideal to use in all situations. Any studies involving seeding terms from a set of emotions should consider the dimensions of emotion relative to the topic before seeding to ensure a representative spectrum of emotion is taken into account. In fact, the keywords may not even be traditional emotions at all. Emotions mean different things dependent on the thematic concerns of a corpus. For example, the emotion spectrum will look very different when classifying political news headlines compared to classifying fairy tales --- the same emotion words can be used differently in these contexts due to the variance of representative emotion spectrums.

We have highlighted the fact that emotions should not be considered as strictly positive or negative, but as a ratio of these. This affects sentiment research, which has mainly focused on mining emoticons --- future research could involve the use of emotion keywords instead, increasing data granularity, and perhaps accuracy.

\subsection{Psychological Applications}\label{psych-app}

Arguably, the most interesting aspect of this research, at least with regard to the emotion domain, is its direct psychological applications. An interesting idea arising from accessibility to co-occurrence data is the potential ability to subliminally infer specific emotions by carefully constructing sentences that convey an arbitrary point across using specific words and phrases that are more commonly expressed when feeling a particular emotion. In other words, we could create `emotional dialects', which could prove to be an interesting tool for reassurance or, on the other end of the scale, manipulation.

\subsubsection{Identification of Underlying Emotional Conditions}

Emotion Profiles can be generated using any form of natural text, from emails to transcripts of interrogations. It is particularly useful to find links between specific emotions, for example, \textit{depression}, to other underlying emotions that may not have been directly targeted, in this case, \textit{sleepiness}. The important aspect to take into consideration when analysing emotion profiles is that fact that it is a visualisation of LSC data, as opposed to a general view of the emotion spectrum, so it defines individual emotions as a combination of other emotions.

\subsubsection{Emotional Engineering}

Deriving psychological emotion equations, specifically relative to a particular domain, will enable psychological researchers to be able to break down feelings to their constituent emotions. This will be useful to understand which emotions to add or take away from an existing emotional condition to create another, more preferable emotion. This could lead to the creation of `new emotions'; existing emotions could be manipulated, combined and `re-packaged' in order to explain types of behaviour, not traditionally thought to be a particular emotion. This mechanism is not dissimilar to how people of different cultures divide the affective world into different basic emotion categories. \citeasnoun{cultures} describes several emotion words in other languages for which no word exists in English. An example from German is the word \textit{schadenfreude}, which refers to pleasure derived from another's displeasure. An example from Japanese is \textit{itoshii}, which refers to longing for an absent loved one; another is \textit{ijirashii}, which refers to a feeling associated with seeing someone praiseworthy overcoming an obstacle. There may be many more different types of emotions to those we are conceptually aware of, and if this is the case, we may need to approach them in different ways.

\subsubsection{Detecting Unconscious Stress Factors}

We have seen that emotion conceptualisation is dependent on geographical region --- similar to dialects of a language --- but why could this be? We have shown, using emotion combinations, that \textit{stress} is highly correlated with \textit{depression}, and it is also correlated with \textit{anxiety}. We could use our data to identify correlations between properties of geographical regions, such as density of roads, number of factories and variety of entertainment venues. Instead of assuming that specific properties contribute to \textit{stress}, we can actually determine what factors are indeed contributing; it could be the case that seemingly negative properties could in fact have a negligible role in causing \textit{stress}, and the opposite could also be true.

\subsubsection{Clinical Assessments}

The disadvantage with our current emotion extraction method is that not everyone will describe experiences using emotion keywords --- we have aggregated many people's responses in the present study. A more powerful method of identifying emotions is by using self-report psychometric instruments that involve participants completing a written or oral questionnaire regarding their emotional experiences in order to analyse their responses taking into account a number of factors. Examples of such instruments include the Depression Anxiety Stress Scale \cite{DASS}, the Beck Hopelessness Scale \cite{BHS} and Profile Of Mood States \cite{POMS}. Indeed \citeasnoun{gpoms1} modified Profile Of Mood States to infer the current global mood on Twitter, as opposed to mining for phrases such as \textit{``I feel calm"}. Using LSC data, however, we are able to identify otherwise unknown unconscious factors that contribute to specific emotions, which could lead to more accurate questionnaires being devised.


\chapter{Conclusion}

A vast majority of computer scientists tend to use Ekman's basic emotion set for emotion categorisation, and it appears that, semantically, it is the most distinct set, with a 2.9\% increase in accuracy compared to the average of the remaining sets. Using an iterative algorithm based on LSC, we have discerned a set of eight (rather than Ekman's six) basic emotion keywords that have been calculated to be the most semantically distinct. This set performed better in all semantic tests than all of the basic emotion models analysed, with a 6.1\% increase in accuracy over Ekman's basic emotion set, providing evidence that by carefully selecting emotion keywords, more of the emotion spectrum can be accounted for.

Emotions must be seen as relative to a specific domain. This research has not focused on any particular domain as it attempts to generalise the nature of emotions, however, as we have seen with the geographical testing, the composition of optimal sets were highly dynamic and dependent on region, even though negligible alterations were observed when testing a random sample --- if there is an underlying domain, then its semantically distinct set of emotion dimensions will differ to those of another corpus of a different domain. If we had to decide upon a universal set of basic emotions, it would be those emotions that appear thrice or greater in the derived basic emotion sets within the analysis (ELSA, DELSAR and the optimal ELSA sets derived from USP77, USE91, USC92, UK49, RAND77):

\begin{center}
\textsc{accepting anxious ashamed contempt interested joyful pleased stressed}
\end{center}

We do not recommended the use of these emotion dimensions universally, although they are a much more accurate and complete set of basic emotions to base emotion classification with than existing basic emotion sets, and certainly more distinct than Ekman's set, at least in terms of the language used for each expression of emotion. Similar to \possessivecite{kim} point of view, we recommend that a different basic emotion set should be chosen dependent on the domain and corpus, suggesting that basic emotions are ultimately not universal and correlated with underlying thematic concerns within the corpus under analysis. The semantic nature of our analysis means that our algorithms can achieve similar results in any language, so long as the language within the corpus under analysis is consistent.

This research has been limited to the corpus used; while extremely relevant, the extent in which temporal effects were mitigated may not have been sufficient. We have based our research on 21 thousand tweets collected over ten days, and the same amount of data over several months may or may not produce different results. The primary limitation of our emotional Twitter corpus is that it consists of expressions that mention an emotion keyword without regard to whether the user actually felt the expressed emotion. However, we assume that the mere use of an emotion keyword still attributes to how people understand each emotion to be. Additionally, adding more emotion keywords may change derived sets, but although we could have used many more popular emotion keywords and synonyms, we decided against this as the current study focuses on specific basic emotion sets. We also noted that adding an additional six basic emotions from Russel's set negligibly altered our results. The Emotional Twitter Corpus is useful for those wanting to obtain raw explicit expressions of emotion. Each expression is labelled, rendering it especially useful for classification projects, although the data as it stands remains unparsed for duplicates or identified `junk' data.

We have essentially ranked emotions and basic emotion sets using our algorithms according to a metric of semantic distinctiveness of extracted expressions from Twitter. Whilst this is useful for discovering psychological similarities of emotion expression, the real power lies in the application of the resulting clustering data. By visualising LSC data matrices, we are able to compare the similarity of compound emotions, analyse the composite properties of emotions and highlight how specific emotions interact with each other, with applications ranging from clinical assessments to applied psychological research.

Emotion may contribute to evolution on a much grander scale than previously thought. Indeed, \citeasnoun{izard-09} suggests that the main component in evolution could be Emotion Schemas, that is, evolution of actions through imitative learning of specific emotions. Mapping such processes could shed light on an updated and, combined with genetic algorithms, a more complete model of human evolution.

\section{Summary of Contributions}

\textbf{An Emotional Twitter Corpus.} We have created a PHP script that streams a large number of emotional expressions from Twitter that can be executed from a UNIX terminal and run in the background. Using this script, we have created a corpus of emotional tweets, stored in a database, with each expression emotion-tagged and linked to additional information such as time (useful for temporal analysis) and nationality (useful for cultural analysis). In the process, we have discovered that the stream rates of emotions are highly irregular, ranging from 2 to 200 tweets per minute for the emotions analysed, including any word filters that were used. Using consistent quantities of tweets for each emotion was key to this research, so a natural limit of the number of documents we could analyse was imposed, this being the number of documents harvested from the emotion with the lowest stream rate.

\textbf{Semantic Distinctiveness Evaluation of Basic Emotion Sets.} We evaluate six basic emotion sets on a scale of \textit{semantic distinctiveness}, based on the theory that the more distinct the language is used to express a certain emotion, then conceptually, i.e. what we understand that emotion keyword to mean, the more psychologically irreducible that emotion is. The less semantically accurate a set of emotions is, then the more similar these emotions are to each other, or in other words, if similar words are used when expressing two different emotions, then these emotions are, in theory, conceptually, and thus psychologically, similar. A large majority of computer scientists tend to use Ekman's basic emotion set for emotion categorisation, and it appears that, semantically, it is the most distinct set.

\textbf{A Semantically Irreducible Emotion Set.} Using DELSAR, an iterative algorithm based on LSC, we derive an optimal basic emotion set consisting of eight, rather than six, emotions. This set performed better in all semantic tests than all of the basic emotion models analysed, with a 6.1\% increase in distinctiveness over Ekman's basic emotion set. Furthermore, the derived set satisfies \textit{the three laws of basic emotions}. We also use our ELSA algorithm to derive optimal basic emotion sets using a variety of geographically specific subcorpora, and present a Semantically Irreducible Emotion Set based on the combination of these derived basic emotion sets and the derived sets using DELSAR and ELSA.

\textbf{The Theory of Emotional Relativity.} The Theory of Emotional Relativity is a fusion of the Linguistic Relativity Hypothesis and \possessivecite{cultures} Cultural Relativity proposition. We propose the theory that each person's conceptualisation of their emotion spectrum is unique; throughout this project, we have aggregated these individualised emotion spaces, although we began to test this theory by analysing subsets of the corpus containing tweets from specific geographical areas. We find that, while the majority of areas are still best modeled by the optimal basic emotion set, this is not true for all areas; in some cases we find that a different set best models the dimensions of their emotion space. Furthermore, we discern the optimal basic emotion set for each geographical area, and find that constituent emotions are as disorderly as the original collection of basic emotion sets.

\textbf{Emotion Visualisation Techniques.} We have demonstrated four key techniques of visualising emotions that illustrate specific properties of emotions, making the LSC data more accessible and easier to navigate: Multi-Dimensional Scaling that highlights properties of emotions relative to each other, Emotion Profiling that highlights individual properties of specific emotions, Emotion Waves that aim to highlight emotional trends and correlations over time by visualising a time-lapse of emotion profiles, and Emotion Equations that aim to illustrate compounded emotions and their similarity to individual emotions.

\textbf{Applications of Emotion Data.} We have discussed a variety of important applications of LSC data, including determining unconscious factors contributing to experienced emotions and improvement and enhancement of clinical assessments. We have also introduced the notion of \textit{Emotional Engineering}. This could lead to the creation of `new emotions'; existing emotions could be manipulated, combined and `re-packaged' in order to explain types of behaviour, not traditionally thought to be a particular emotion. There may be different types of emotions to those we are conceptually aware of, and if this is the case, we may need to approach them in different ways.

Focusing on expressions in text allows our methods to be applied to the wealth of data on the Internet, and text in general. 
We are slowing beginning acquire a full understanding of the mechanisms of emotion, increasing our awareness of originally unconscious factors, potentially up to a point in which we will be able to predict both individual and global emotions in response to arbitrary scenarios. Emotion theory remains an area of active Computer Science research, and there are several matters yet to be thoroughly researched.


\chapter{Future Research}\label{further-research}

Apart from expanding on the applications as discussed in Chapter~\ref{vis}, there are many areas that require extensive research to fully understand the nature of emotion. Does anonymity increase negativity? Does the negation of an emotion equate to its psychological opponent? How closely linked are emotion and sentiment? How does the time of day affect our emotions? In addition to these questions, a number of research projects have been conceptualised by the author.

\section{Current Project Extensions}

The accuracy of our data collection could be improved. We would like to extend our streaming script to a list of all emotions by using an extensive list of keywords seeded with synonyms from WordNet, and also taking into consideration colloquialisms of emotion keywords, such as \textit{jeals} (meaning \textit{jealous}). Another method we would like to integrate is using LDA to populate the initial list of keywords from a corpus, thus automating the keyword selection process and identifying underlying thematic concerns for a more accurate seeding of keywords.

Applying our algorithms to Facebook profile data, we could be able to detect the individualisation of emotion by discerning emotional spaces right down to individual people (in other words, \textit{attitudes}), which could have applications in consumer profiling and online dating. It may be the case that using Facebook data would produce significantly different results to our Twitter-based analysis, possibly due to both the increase in blog space and the privatised nature of posting. Since the majority of profiles are limited to friends, there may be a considerable lack of public image concerns, and analysing this data could prove to create a more accurate representation of emotional mechanisms. There could, however, be \textit{social} image concerns.

We would also like to find some sort of measurement for determining the level of bias resulting from public image considerations. By comparing public emotions, for example from Twitter, with emotions thought to have been expressed in private, for example the Enron internal email data set, we could be able to detect a semantic measurement of publicity bias.

\section{Economic Prediction}

There has been a growing amount of research involving stock market prediction techniques by analysing social network data, specifically Twitter. As we have previously discussed, \citeasnoun{stock} used a modified version of Profile Of Mood States to analyse the current mood on Twitter, and found that the \textit{calmness} dimension predicted the movement of the Dow Jones Industrial Average Index four days in advance, with an accuracy of 87\%. More recently, \citeasnoun{stock-predict} created trading strategies that outperformed baseline strategies by extracting connected components and interaction nodes from tweets. By extending such research to cover a wide range of sources and combining it with our mechanisms of emotion analysis, we would be able to more accurately predict the economy, possibly right down to individual stock. As decisions are heavily influenced by emotions, the use of real-time emotion semantic analysis could be considered for outperforming more traditional trading strategies.

\section{General Principle of Emotion}

We present the \textit{General Principle of Emotion}: the theory that differential emotion values may prove to be more accurate in defining emotion qualia than the experience of emotion qualia at each point in time. These `Delta Emotion' values infer the emotional response of a given time point (i.e. event) by capturing and analysing the emotional responses at either side of the time point, and could provide a more genuine emotion signal for predictive analysis. Delta emotions are geared towards long-term analysis; specifically, it could be thought of as measuring shifts in perception. Using this data, we could be able to measure what we call \textit{Emotion Injection}, or in other words, measuring to what extent does perception change according to which, why, how and when emotions are experienced. If we can understand someone else's emotional perception, we will better be able to manipulate their emotions using language, a position not to be taken lightly without a vast array of opportunities and consequences. 

\section{Machine Consciousness}\label{memes}

There is a strong link between emotion conceptualisation and \textit{memetics}. The term meme was introduced and defined by \citeasnoun{selfish-gene} as ``the basic unit of cultural transmission, or imitation", and in the English Oxford Dictionary as ``an element of culture that may be considered to be passed on by non-genetic means". Emotion may contribute to evolution on a much grander scale than previously thought. Indeed, \citeasnoun{izard-09} suggests that the main component in evolution could be Emotion Schemas, that is, evolution of actions through imitative learning of specific emotions. Mapping such processes could shed light on an updated and, combined with genetic algorithms, a more complete model of human evolution. Memetic theory states that the ability to imitate is the only requirement for language to occur in evolution, and it has been shown in several studies that syntax and semantics emerge spontaneously \citeaffixed{meme-c}{for a discussion, see}. Thus, by analysing language we should be able to reverse-engineer the imitative mechanisms of humans. It may be the case that we cannot simply build the most accurate `emotion brain' and hard code it into a machine; an emotional brain would need to experience and evolve using emotion-based memetic algorithms. Using such a process would enable us to create an emotional projection of ourself with the potential for acquiring a human-like illusion of consciousness; indeed, it may hold key to simulating minds, and could provide the first step towards testing \possessivecite{simulation} simulation hypothesis.

\section{Epilogue}

We have conducted thorough research and discovered emotional mechanisms that advance our understanding of the individual conceptualisation of emotion qualia through analysis of social language. We have explored what emotional semantic analysis data could reveal about the nature of emotions and have discussed its potential applications to both clinical and general psychological research. We have not, however, spoken of the emotion \textit{love}. It could be argued that no amount of language could describe what true \textit{love} is. Or, equally as plausible, that true \textit{love} is in fact an illusion, which is not hard to believe if consciousness itself is an illusion. Perhaps an emotional simulation of ourselves could tell us the answer. Whichever direction our research into emotions take us, there is still much to be learned.


\bibliography{ExampleBibFile}


\appendix

\chapter{EARL Emotion WordNet Synonyms}\label{EARL}

\includepdf[noautoscale=true,offset=25 0]{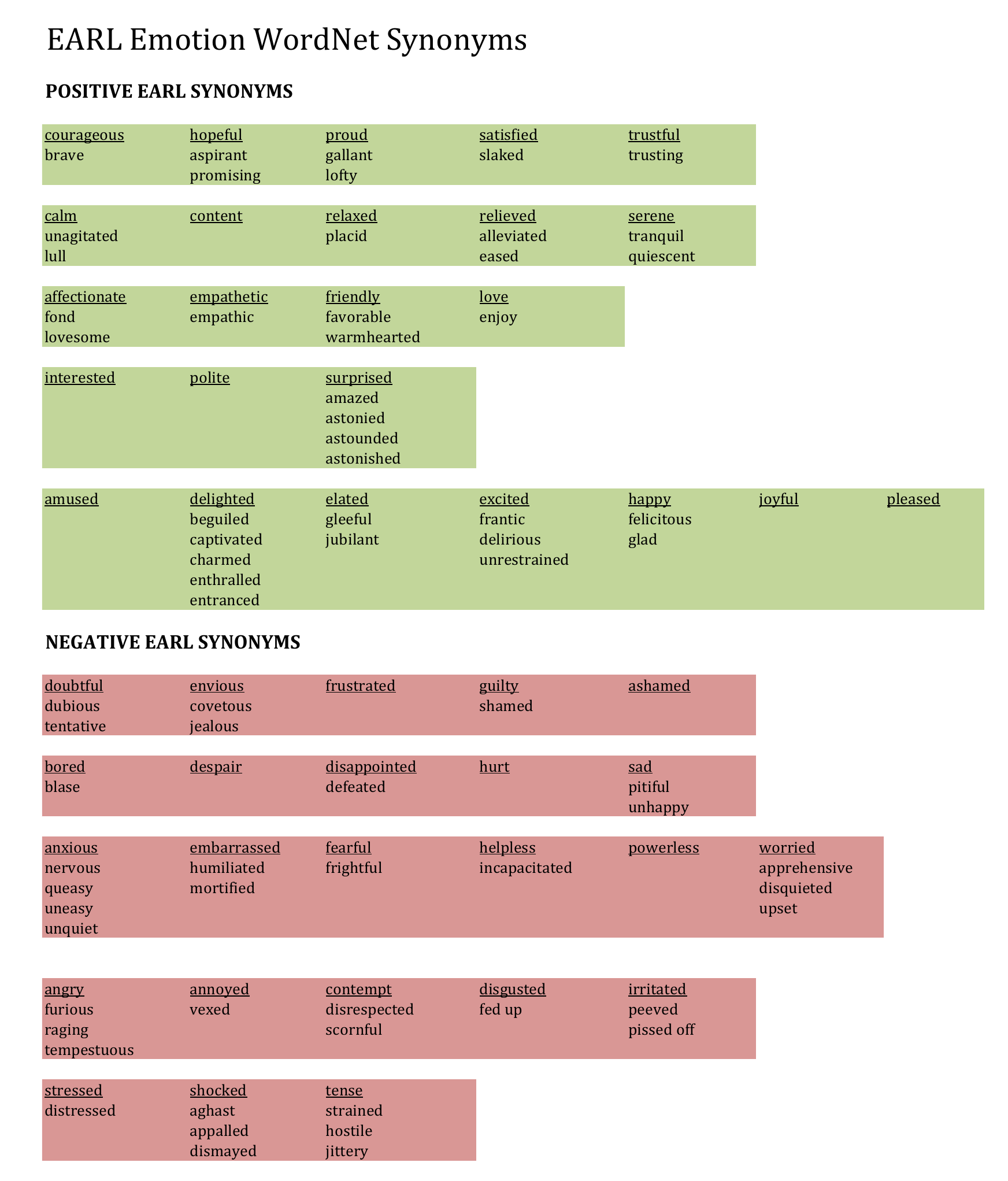}

\chapter{Raw results output}

\begin{lstlisting}[breaklines=true,showstringspaces=false,title=DELSAR1000 Algorithm,label=delsar-print]
>>> 
DELSAR1000 testing ALL with 60 dimensions.

21 emotions, 0 reductions remaining.
Creating dictionary...
Dictionary(13532 unique tokens)
Creating corpus object...
Done.
Generating LSA Space...
<gensim.interfaces.TransformedCorpus object at 0x0373B1B0>
Indexing LSA Space...

Warning (from warnings module):
  File "C:\Python27\lib\site-packages\scipy\sparse\compressed.py", line 122
    % self.indices.dtype.name )
UserWarning: indices array has non-integer dtype (float64)
Done.
Generating document similarity matrix...
['accepting', 0.452]
['angry', 0.248]
['anticipating', 0.312]
['anxious', 0.272]
['ashamed', 0.366]
['contempt', 0.356]
['depressed', 0.193]
['disgusted', 0.251]
['excited', 0.227]
['guilty', 0.339]
['happy', 0.255]
['interested', 0.46]
['joyful', 0.397]
['miserable', 0.272]
['pleased', 0.359]
['relaxed', 0.245]
['sad', 0.259]
['scared', 0.249]
['sleepy', 0.332]
['stressed', 0.295]
['surprised', 0.295]
TOTAL ACCURACY: 0.306381
Time Taken:
27.1112883866

>>> 

DELSAR1000 testing ALL with 40 dimensions.

21 emotions, 13 reductions remaining.
Creating dictionary...
Dictionary(13532 unique tokens)
Creating corpus object...
Done.
Generating LSA Space...
<gensim.interfaces.TransformedCorpus object at 0x038011B0>
Indexing LSA Space...

Warning (from warnings module):
  File "C:\Python27\lib\site-packages\scipy\sparse\compressed.py", line 122
    % self.indices.dtype.name )
UserWarning: indices array has non-integer dtype (float64)
Done.
Generating document similarity matrix...
Removed depressed (0.187000)
20 emotions, 12 reductions remaining.
Creating dictionary...
Dictionary(13170 unique tokens)
Creating corpus object...
Done.
Generating LSA Space...
<gensim.interfaces.TransformedCorpus object at 0x03A3FD10>
Indexing LSA Space...
Done.
Generating document similarity matrix...
Removed disgusted (0.217000)
19 emotions, 11 reductions remaining.
Creating dictionary...
Dictionary(12666 unique tokens)
Creating corpus object...
Done.
Generating LSA Space...
<gensim.interfaces.TransformedCorpus object at 0x038BB370>
Indexing LSA Space...
Done.
Generating document similarity matrix...
Removed sad (0.235000)
18 emotions, 10 reductions remaining.
Creating dictionary...
Dictionary(12178 unique tokens)
Creating corpus object...
Done.
Generating LSA Space...
<gensim.interfaces.TransformedCorpus object at 0x038CC090>
Indexing LSA Space...
Done.
Generating document similarity matrix...
Removed happy (0.246000)
17 emotions, 9 reductions remaining.
Creating dictionary...
Dictionary(11698 unique tokens)
Creating corpus object...
Done.
Generating LSA Space...
<gensim.interfaces.TransformedCorpus object at 0x0393B430>
Indexing LSA Space...
Done.
Generating document similarity matrix...
Removed excited (0.274000)
16 emotions, 8 reductions remaining.
Creating dictionary...
Dictionary(11230 unique tokens)
Creating corpus object...
Done.
Generating LSA Space...
<gensim.interfaces.TransformedCorpus object at 0x03A87D90>
Indexing LSA Space...
Done.
Generating document similarity matrix...
Removed scared (0.290000)
15 emotions, 7 reductions remaining.
Creating dictionary...
Dictionary(10774 unique tokens)
Creating corpus object...
Done.
Generating LSA Space...
<gensim.interfaces.TransformedCorpus object at 0x03A8FA10>
Indexing LSA Space...
Done.
Generating document similarity matrix...
Removed relaxed (0.314000)
14 emotions, 6 reductions remaining.
Creating dictionary...
Dictionary(10241 unique tokens)
Creating corpus object...
Done.
Generating LSA Space...
<gensim.interfaces.TransformedCorpus object at 0x039DFA90>
Indexing LSA Space...
Done.
Generating document similarity matrix...
Removed anxious (0.337000)
13 emotions, 5 reductions remaining.
Creating dictionary...
Dictionary(9759 unique tokens)
Creating corpus object...
Done.
Generating LSA Space...
<gensim.interfaces.TransformedCorpus object at 0x05E7BD70>
Indexing LSA Space...
Done.
Generating document similarity matrix...
Removed angry (0.362000)
12 emotions, 4 reductions remaining.
Creating dictionary...
Dictionary(9241 unique tokens)
Creating corpus object...
Done.
Generating LSA Space...
<gensim.interfaces.TransformedCorpus object at 0x074DBB30>
Indexing LSA Space...
Done.
Generating document similarity matrix...
Removed anticipating (0.375000)
11 emotions, 3 reductions remaining.
Creating dictionary...
Dictionary(8652 unique tokens)
Creating corpus object...
Done.
Generating LSA Space...
<gensim.interfaces.TransformedCorpus object at 0x06D4E590>
Indexing LSA Space...
Done.
Generating document similarity matrix...
Removed miserable (0.392000)
10 emotions, 2 reductions remaining.
Creating dictionary...
Dictionary(8058 unique tokens)
Creating corpus object...
Done.
Generating LSA Space...
<gensim.interfaces.TransformedCorpus object at 0x0B154B70>
Indexing LSA Space...
Done.
Generating document similarity matrix...
Removed surprised (0.417000)
9 emotions, 1 reductions remaining.
Creating dictionary...
Dictionary(7479 unique tokens)
Creating corpus object...
Done.
Generating LSA Space...
<gensim.interfaces.TransformedCorpus object at 0x039731B0>
Indexing LSA Space...
Done.
Generating document similarity matrix...
Removed guilty (0.444000)
8 emotions, 0 reductions remaining.
Creating dictionary...
Dictionary(6819 unique tokens)
Creating corpus object...
Done.
Generating LSA Space...
<gensim.interfaces.TransformedCorpus object at 0x0BBDF0D0>
Indexing LSA Space...
Done.
Generating document similarity matrix...
['accepting', 0.553]
['ashamed', 0.534]
['contempt', 0.574]
['interested', 0.603]
['joyful', 0.519]
['pleased', 0.506]
['sleepy', 0.591]
['stressed', 0.502]
TOTAL ACCURACY: 0.547750
Time Taken:
258.695950863
>>>
\end{lstlisting}
\begin{lstlisting}[breaklines=true,showstringspaces=false,title=ELSA1000 Algorithm (10 Dimensions),label=elsa-print]
>>> 

ELSA1000 testing accepting with 10 dimensions.
1 emotions, 0 reductions remaining.
Creating dictionary...
Dictionary(1249 unique tokens)
Creating corpus object...
Done.
Generating LSA Space...
<gensim.interfaces.TransformedCorpus object at 0x03990E10>
Indexing LSA Space...
Done.
Generating document similarity matrix...
['accepting', 0.9675678052306175]
ELSA1000 testing angry with 10 dimensions.
1 emotions, 0 reductions remaining.
Creating dictionary...
Dictionary(1203 unique tokens)
Creating corpus object...
Done.
Generating LSA Space...
<gensim.interfaces.TransformedCorpus object at 0x039904B0>
Indexing LSA Space...

Warning (from warnings module):
  File "C:\Python27\lib\site-packages\scipy\sparse\compressed.py", line 122
    % self.indices.dtype.name )
UserWarning: indices array has non-integer dtype (float64)
Done.
Generating document similarity matrix...
['angry', 0.9511218096613884]
ELSA1000 testing anticipating with 10 dimensions.
1 emotions, 0 reductions remaining.
Creating dictionary...
Dictionary(1233 unique tokens)
Creating corpus object...
Done.
Generating LSA Space...
<gensim.interfaces.TransformedCorpus object at 0x03992310>
Indexing LSA Space...
Done.
Generating document similarity matrix...
['anticipating', 0.9295840657949448]
ELSA1000 testing anxious with 10 dimensions.
1 emotions, 0 reductions remaining.
Creating dictionary...
Dictionary(829 unique tokens)
Creating corpus object...
Done.
Generating LSA Space...
<gensim.interfaces.TransformedCorpus object at 0x0392F870>
Indexing LSA Space...
Done.
Generating document similarity matrix...
['anxious', 0.9886916783452034]
ELSA1000 testing ashamed with 10 dimensions.
1 emotions, 0 reductions remaining.
Creating dictionary...
Dictionary(1271 unique tokens)
Creating corpus object...
Done.
Generating LSA Space...
<gensim.interfaces.TransformedCorpus object at 0x0392FC70>
Indexing LSA Space...
Done.
Generating document similarity matrix...
['ashamed', 0.9471849942207337]
ELSA1000 testing contempt with 10 dimensions.
1 emotions, 0 reductions remaining.
Creating dictionary...
Dictionary(1468 unique tokens)
Creating corpus object...
Done.
Generating LSA Space...
<gensim.interfaces.TransformedCorpus object at 0x038CEDD0>
Indexing LSA Space...
Done.
Generating document similarity matrix...
['contempt', 0.977734048485756]
ELSA1000 testing depressed with 10 dimensions.
1 emotions, 0 reductions remaining.
Creating dictionary...
Dictionary(1158 unique tokens)
Creating corpus object...
Done.
Generating LSA Space...
<gensim.interfaces.TransformedCorpus object at 0x039A5CF0>
Indexing LSA Space...
Done.
Generating document similarity matrix...
['depressed', 0.9502798588275909]
ELSA1000 testing disgusted with 10 dimensions.
1 emotions, 0 reductions remaining.
Creating dictionary...
Dictionary(1270 unique tokens)
Creating corpus object...
Done.
Generating LSA Space...
<gensim.interfaces.TransformedCorpus object at 0x038CEDD0>
Indexing LSA Space...
Done.
Generating document similarity matrix...
['disgusted', 0.9326055783033371]
ELSA1000 testing excited with 10 dimensions.
1 emotions, 0 reductions remaining.
Creating dictionary...
Dictionary(1121 unique tokens)
Creating corpus object...
Done.
Generating LSA Space...
<gensim.interfaces.TransformedCorpus object at 0x0399E5D0>
Indexing LSA Space...
Done.
Generating document similarity matrix...
['excited', 0.9424464861154557]
ELSA1000 testing guilty with 10 dimensions.
1 emotions, 0 reductions remaining.
Creating dictionary...
Dictionary(1306 unique tokens)
Creating corpus object...
Done.
Generating LSA Space...
<gensim.interfaces.TransformedCorpus object at 0x038CEC10>
Indexing LSA Space...
Done.
Generating document similarity matrix...
['guilty', 0.925568910241127]
ELSA1000 testing happy with 10 dimensions.
1 emotions, 0 reductions remaining.
Creating dictionary...
Dictionary(1216 unique tokens)
Creating corpus object...
Done.
Generating LSA Space...
<gensim.interfaces.TransformedCorpus object at 0x039A7F90>
Indexing LSA Space...
Done.
Generating document similarity matrix...
['happy', 0.9486168209314346]
ELSA1000 testing interested with 10 dimensions.
1 emotions, 0 reductions remaining.
Creating dictionary...
Dictionary(1398 unique tokens)
Creating corpus object...
Done.
Generating LSA Space...
<gensim.interfaces.TransformedCorpus object at 0x039A64F0>
Indexing LSA Space...
Done.
Generating document similarity matrix...
['interested', 0.9563596816658974]
ELSA1000 testing joyful with 10 dimensions.
1 emotions, 0 reductions remaining.
Creating dictionary...
Dictionary(1329 unique tokens)
Creating corpus object...
Done.
Generating LSA Space...
<gensim.interfaces.TransformedCorpus object at 0x0398DC70>
Indexing LSA Space...
Done.
Generating document similarity matrix...
['joyful', 0.9408138270378112]
ELSA1000 testing miserable with 10 dimensions.
1 emotions, 0 reductions remaining.
Creating dictionary...
Dictionary(1263 unique tokens)
Creating corpus object...
Done.
Generating LSA Space...
<gensim.interfaces.TransformedCorpus object at 0x039AC3F0>
Indexing LSA Space...
Done.
Generating document similarity matrix...
['miserable', 0.9543267293572426]
ELSA1000 testing pleased with 10 dimensions.
1 emotions, 0 reductions remaining.
Creating dictionary...
Dictionary(1220 unique tokens)
Creating corpus object...
Done.
Generating LSA Space...
<gensim.interfaces.TransformedCorpus object at 0x039AC8B0>
Indexing LSA Space...
Done.
Generating document similarity matrix...
['pleased', 0.9491549980044365]
ELSA1000 testing relaxed with 10 dimensions.
1 emotions, 0 reductions remaining.
Creating dictionary...
Dictionary(1267 unique tokens)
Creating corpus object...
Done.
Generating LSA Space...
<gensim.interfaces.TransformedCorpus object at 0x039AC630>
Indexing LSA Space...
Done.
Generating document similarity matrix...
['relaxed', 0.938998939216137]
ELSA1000 testing sad with 10 dimensions.
1 emotions, 0 reductions remaining.
Creating dictionary...
Dictionary(1367 unique tokens)
Creating corpus object...
Done.
Generating LSA Space...
<gensim.interfaces.TransformedCorpus object at 0x039A4AD0>
Indexing LSA Space...
Done.
Generating document similarity matrix...
['sad', 0.9406832031607628]
ELSA1000 testing scared with 10 dimensions.
1 emotions, 0 reductions remaining.
Creating dictionary...
Dictionary(1227 unique tokens)
Creating corpus object...
Done.
Generating LSA Space...
<gensim.interfaces.TransformedCorpus object at 0x039A4930>
Indexing LSA Space...
Done.
Generating document similarity matrix...
['scared', 0.9354943093657494]
ELSA1000 testing sleepy with 10 dimensions.
1 emotions, 0 reductions remaining.
Creating dictionary...
Dictionary(1019 unique tokens)
Creating corpus object...
Done.
Generating LSA Space...
<gensim.interfaces.TransformedCorpus object at 0x0399D230>
Indexing LSA Space...
Done.
Generating document similarity matrix...
['sleepy', 0.9391740371584892]
ELSA1000 testing stressed with 10 dimensions.
1 emotions, 0 reductions remaining.
Creating dictionary...
Dictionary(1124 unique tokens)
Creating corpus object...
Done.
Generating LSA Space...
<gensim.interfaces.TransformedCorpus object at 0x0399EDD0>
Indexing LSA Space...
Done.
Generating document similarity matrix...
['stressed', 0.9501295542120933]
ELSA1000 testing surprised with 10 dimensions.
1 emotions, 0 reductions remaining.
Creating dictionary...
Dictionary(1228 unique tokens)
Creating corpus object...
Done.
Generating LSA Space...
<gensim.interfaces.TransformedCorpus object at 0x0392FAB0>
Indexing LSA Space...
Done.
Generating document similarity matrix...
['surprised', 0.9372707696557045]
0.967567805231
0.951121809661
0.929584065795
0.988691678345
0.947184994221
0.977734048486
0.950279858828
0.932605578303
0.942446486115
0.925568910241
0.948616820931
0.956359681666
0.940813827038
0.954326729357
0.949154998004
0.938998939216
0.940683203161
0.935494309366
0.939174037158
0.950129554212
0.937270769656
Time Taken:
5.45335914796
\end{lstlisting}

\chapter{Code}

Up-to-date code is available at http://www.aeir.co.uk.


\begin{landscape}
\section{Database Schema (MySQL)}
\begin{lstlisting}[language=SQL, breaklines=true,showstringspaces=false]
----------------------------------
-- TWEETS.sql                   --
--                              --
-- Creates the TWEETS table     --
--                              --
-- Compatibility: MySQL         --
--                              --
----------------------------------
-- Host: mysql5host.bath.ac.uk  --
-- Database: `PROJtwitterEDB`   --
----------------------------------

--
-- Table structure for table `TWEETS`
--

CREATE TABLE `TWEETS` (
  `id` int(20) UNSIGNED NOT NULL AUTO_INCREMENT COMMENT 'Table key',
  `tweet_id` bigint(20) UNSIGNED NOT NULL COMMENT 'Twitter tweet ID for lookup',
  `text` varchar(150) NOT NULL,
  `emotion` varchar(150) NOT NULL,
  `screen_name` varchar(255) NOT NULL,
  `retweet_count` int(11) NOT NULL,
  `created_at` datetime NOT NULL COMMENT 'Timestamp of database insert',
  `timezone` varchar(255) NOT NULL COMMENT 'Twitter-defined, see seperate SQL file for datetime conversion',
  `flag` tinyint(4) NOT NULL COMMENT 'Generic Flag',
  PRIMARY KEY (`id`)
) ENGINE=InnoDB  DEFAULT CHARSET=utf8;
\end{lstlisting}
\newpage
\section{Twitter Stream (PHP)}
\begin{lstlisting}[language=PHP, breaklines=true,showstringspaces=false,label=streamword]
<?php
/*  streamword.php
 *
 *  Creates a twitter stream for a single keyword
 *
 *  Filters out tweets that contain specific phrases
 *
 *  Saves matching tweets into the database
 *  
 *  Changes emotion every 5 minutes  
 *
 */ 
  set_time_limit(0);
  include("JSON.php");
  include("timer.php");
  $currentEmotionCode = 0;
  $filter = array();
  $filter[0] = array(0 => 'RT ','not accepting', 'not really accepting', 'not that accepting', 'wasn\'t that accepting', 'wasn\'t very accepting', 'unaccepting', 'disaccepting', 'http://', ' RT', '@');
  $filter[1] = array(0 => 'RT ','not angry', 'not really angry', 'not that angry', 'wasn\'t that angry', 'wasn\'t very angry', 'unangry', 'angry birds', 'http://', ' RT', '@');
  $filter[2] = array(0 => 'RT ','not anticipating', 'not really anticipating', 'not that anticipating', 'wasn\'t that anticipating', 'wasn\'t very anticipating', 'unanticipating', 'disanticipating', 'http://', ' RT', '@');
  $filter[3] = array(0 => 'RT ','not anxious', 'not really anxious', 'not that anxious', 'wasn\'t that anxious', 'wasn\'t very anxious', 'unanxious', 'disanxious', 'http://', ' RT', '@');
  $filter[4] = array(0 => 'RT ','not ashamed', 'not really ashamed', 'not that ashamed', 'wasn\'t that ashamed', 'wasn\'t very ashamed', 'unashamed', 'disashamed', 'http://', ' RT', '@');
  $filter[5] = array(0 => 'RT ','not contempt', 'not really contempt', 'not that contempt', 'wasn\'t that contempt', 'wasn\'t very contempt', 'uncontempt', 'discontempt', 'in contempt', 'contempt of', 'http://', ' RT', '@');
  $filter[6] = array(0 => 'RT ','not depressed', 'not really depressed', 'not that depressed', 'wasn\'t that depressed', 'wasn\'t very depressed', 'undepressed', 'disdepressed', 'http://', ' RT', '@');
  $filter[7] = array(0 => 'RT ','not disgusted', 'not really disgusted', 'not that disgusted', 'wasn\'t that disgusted', 'wasn\'t very disgusted', 'undisgusted', 'disdisgusted', 'http://', ' RT', '@');
  $filter[8] = array(0 => 'RT ','not excited', 'not really excited', 'not that excited', 'wasn\'t that excited', 'wasn\'t very excited', 'unexcited', 'disexcited', 'http://', ' RT', '@');
  $filter[9] = array(0 => 'RT ','not guilty', 'not really guilty', 'not that guilty', 'wasn\'t that guilty', 'wasn\'t very guilty', 'unguilty', 'found guilty', 'http://', ' RT', '@');
  $filter[10] = array(0 => 'RT ','not happy', 'not really happy', 'not that happy', 'wasn\'t that happy', 'wasn\'t very happy', 'unhappy', 'happy birthday', 'http://', ' RT', '@');
  $filter[11] = array(0 => 'RT ','not interested', 'not really interested', 'not that interested', 'wasn\'t that interested', 'wasn\'t very interested', 'uninterested', 'disinterested', 'http://', ' RT', '@');
  $filter[12] = array(0 => 'RT ','not joyful', 'not really joyful', 'not that joyful', 'wasn\'t that joyful', 'wasn\'t very joyful', 'unjoyful', 'disjoyful', 'http://', ' RT', '@');
  $filter[13] = array(0 => 'RT ','not miserable', 'not really miserable', 'not that miserable', 'wasn\'t that miserable', 'wasn\'t very miserable', 'unmiserable', 'dismiserable', 'http://', ' RT', '@');
  $filter[14] = array(0 => 'RT ','not pleased', 'not really pleased', 'not that pleased', 'wasn\'t that pleased', 'wasn\'t very pleased', 'unpleased', 'displeased', 'http://', ' RT', '@');
  $filter[15] = array(0 => 'RT ','not relaxed', 'not really relaxed', 'not that relaxed', 'wasn\'t that relaxed', 'wasn\'t very relaxed', 'unrelaxed', 'disrelaxed', 'http://', ' RT', '@');
  $filter[16] = array(0 => 'RT ','not sad', 'not really sad', 'not that sad', 'wasn\'t that sad', 'wasn\'t very sad', 'unsad', 'dissad', 'http://', ' RT', '@');
  $filter[17] = array(0 => 'RT ','not scared', 'not really scared', 'not that scared', 'wasn\'t that scared', 'wasn\'t very scared', 'unscared', 'disscared', 'http://', ' RT', '@');
  $filter[18] = array(0 => 'RT ','not sleepy', 'not really sleepy', 'not that sleepy', 'wasn\'t that sleepy', 'wasn\'t very sleepy', 'unsleepy', 'dissleepy', 'http://', ' RT', '@');
  $filter[19] = array(0 => 'RT ','not stressed', 'not really stressed', 'not that stressed', 'wasn\'t that stressed', 'wasn\'t very stressed', 'unstressed', 'disstressed', 'http://', ' RT', '@');
  $filter[20] = array(0 => 'RT ','not surprised', 'not really surprised', 'not that surprised', 'wasn\'t that surprised', 'wasn\'t very surprised', 'unsurprised', 'dissurprised', 'http://', ' RT', '@'); 
  $searchword = array();
  $searchword[0] = "accepting"; 
  $searchword[1] = "angry"; 
  $searchword[2] = "anticipating"; 
  $searchword[3] = "anxious"; 
  $searchword[4] = "ashamed"; 
  $searchword[5] = "contempt"; 
  $searchword[6] = "depressed"; 
  $searchword[7] = "disgusted"; 
  $searchword[8] = "excited"; 
  $searchword[9] = "guilty"; 
  $searchword[10] = "happy"; 
  $searchword[11] = "interested"; 
  $searchword[12] = "joyful"; 
  $searchword[13] = "miserable"; 
  $searchword[14] = "pleased"; 
  $searchword[15] = "relaxed"; 
  $searchword[16] = "sad"; 
  $searchword[17] = "scared"; 
  $searchword[18] = "sleepy"; 
  $searchword[19] = "stressed"; 
  $searchword[20] = "surprised";
  
  function harvest() {
    global $currentEmotionCode;
    global $filter;
    global $searchword;    
    $query_data = array('track' => $searchword[$currentEmotionCode]);
    $user = 'USERNAME';	
    $pass = 'PASSWORD'; 
    $fp = fsockopen("ssl://stream.twitter.com", 443, $errno, $errstr, 720);
    if(!$fp){
      echo "$errstr ($errno)\n";
    } 
    else {
      $mysqli = new mysqli('HOSTNAME', 'USERNAME', 'PASSWORD', 'DATABASE');
      if (mysqli_connect_errno()) { 
         printf("Can't connect to MySQL Server. Errorcode: %s\n", mysqli_connect_error());  
      }
    	$request = "GET /1/statuses/filter.json?".http_build_query($query_data)." HTTP/1.1\r\n";
    	$request .= "Host: stream.twitter.com\r\n";
    	$request .= "Authorization: Basic ".base64_encode($user.":".$pass)."\r\n\r\n";
    	fwrite($fp, $request);
    	// Start a timer, end streaming the current emotion after 5 minutes
      $timer = new timer(1); // constructor starts the timer
      while(!feof($fp) && $timer->get() < 300) {
    		$json = fgets($fp);
    		$tweet = json_decode($json, true);
    		if($tweet) {
          $in = true;		
          foreach($filter[$currentEmotionCode] as $word) {
            if(stripos($tweet["text"],$word) !== false) {$in = false; echo "SKIPPED ".$word."\n\n";} 
          }
          if(str_word_count($tweet["text"]) < 10) {$in = false; echo "Less than 10 words SKIPPED\n\n";} 
          //if(!preg_match('~[^A-Za-z0-9 ]~', $tweet["text"])) {$in = false; echo "Non alphanumeric SKIPPED\n\n";} 
          if($in) {     
      		  $text = $mysqli->real_escape_string($tweet["text"]);
      		  $tweet_id = $mysqli->real_escape_string($tweet["id_str"]);
      		  $screen_name = $mysqli->real_escape_string($tweet["user"]["screen_name"]);
      		  $retweet_count = $mysqli->real_escape_string($tweet["retweet_count"]);
            $time_zone = $mysqli->real_escape_string($tweet["user"]["time_zone"]);
            if($result = $mysqli->query("INSERT INTO `PROJtwitterEDB`.`ETWEETS`(`tweet_id`,`text`,`emotion`,`screen_name`,`retweet_count`,
              `created_at`,`timezone`) VALUES($tweet_id,'$text','$searchword[$currentEmotionCode]','$screen_name',$retweet_count,NOW(),
              '$time_zone')")) {
              echo $tweet["text"]." FROM ".$tweet["user"]["time_zone"]." SAVED TO DATABASE AT ".date("Y-m-d H:i:s")."\n";
            }
            else {
              echo $mysqli->error;
            }		
    		  }
        }
      }
    	$mysqli->close();
    	fclose($fp);                   
    }
    // Increment up to 21 the loop round to 1
    if($currentEmotionCode + 1 == 21) {
      $currentEmotionCode = 0;
    }
    else {
      $currentEmotionCode += 1;
    }
    harvest();
  }
  
  harvest();    
?>
\end{lstlisting}
\newpage
\section{DELSAR/ELSA (Python)}
\begin{lstlisting}[language=Python, breaklines=true,showstringspaces=false,label=elsacode]
###################################################################
# DELSAR.py                                                       #
#                                                                 # 
# Algorithms:                                                     #
#                                                                 #
# DELSAR                                                          #
# Documented-Emotion Latent Semantic Algorithmic Reducer          #
# Calculates semantic distinctiveness of a corpus, given each     #
# document is labeled using LSC                                   #
# Reduces an initial keyword (label) set using LSC                #
#                                                                 #
# ELSA                                                            #
# Emotional Latent Semantic Analysis                              #
# Calculates the accuracy of emotion-specific sub-corpora, using  #
# the average of maximum cosine similarity between documents      #
#                                                                 #
# @requires MySQLdb, gensim                                       #
# @author Eugene Yuta Bann                                        #
# @version 28/03/12                                               #
#                                                                 #
###################################################################

from gensim import corpora, models, similarities
import MySQLdb, time, itertools
from collections import Counter

########## VARIABLES ##############################################

ELSA = False
printDELSAR = True # print array for Excel matrix input

# Uncomment emotion set to test:

#emotionTest = "IZARD"
#emotionTest = "RUSSELL"
#emotionTest = "PLUTCHIK"
emotionTest = "EKMAN"
#emotionTest = "TOMKINS"
#emotionTest = "JOHNSON" #OATLEY
#emotionTest = "ALL"
#emotionTest = "DELSAR"
#emotionTest = "BANN"

limit = 100 # Document limit per emotion
dimension = 40 # LSA number of dimensions

# Uncomment additional parameters for MySQL query:

sqlHaving = ""
#sqlHaving = "HAVING idx MOD 4 = 0"
#sqlHaving = "HAVING idx MOD 13 = 1"
#sqlHaving = "HAVING idx MOD 3 = 2"
#sqlHaving = "HAVING (idx MOD 11 = 7 AND idx > 280)"
#sqlHaving = "AND timezone = 'London'"
#sqlHaving = "AND NOT timezone = 'Mountain Time (US & Canada)' AND NOT timezone = 'Pacific Time (US & Canada)' AND NOT timezone = 'Eastern Time (US & Canada)' AND NOT timezone = 'Central Time (US & Canada)' AND NOT timezone = '' AND NOT timezone = 'Quito' AND NOT timezone = 'Atlantic Time (Canada)'"
#sqlHaving = "AND (timezone = 'London' OR timezone = 'Edinburgh' OR timezone = 'Dublin')"
#sqlHaving = "AND (timezone = 'Mountain Time (US & Canada)' OR timezone = 'Pacific Time (US & Canada)' OR timezone = 'Eastern Time (US & Canada)' OR timezone = 'Central Time (US & Canada)')"
#sqlHaving = "AND timezone = 'Pacific Time (US & Canada)'"
#sqlHaving = "AND timezone = 'Eastern Time (US & Canada)'"
#sqlHaving = "AND timezone = 'Central Time (US & Canada)'"
#sqlHaving = "AND timezone = 'Quito'"


# Possible Asia group: Beijing Tokyo Hong Kong Jakarta Kuala Lumpur Singapore

################################################################## 

# Create emotion term arrays
# Change below for where the emotion lists are stored
# .e files are text files, each emotion on a new line
emotionFile = "eLists/" + emotionTest + ".e"
emotionTerms = [emotion.lower().strip() for emotion in open(emotionFile)]
# Database variables
db = MySQLdb.connect("HOSTNAME", "USERNAME", "PASSWORD", "DATABASE")
cursor = db.cursor()
          
###################################################################         

## For DELSA (without reduction):
## set reduceTo to a number higher than the initial set
def DELSAR(initial, reduceTo):
     # Corpus class to load each document iteratively from database
     class MyCorpus(object):
          def __iter__(self):
               for emotion in range(len(emotionTerms)):
                    try:
                         sql = "SELECT text, @idx:=@idx+1 AS idx FROM ETWEETS WHERE emotion = '%s' %s LIMIT %d" % (emotionTerms[emotion], sqlHaving, limit)
                         cursor.execute("SELECT @idx:=0;")
                         cursor.execute(sql)
                         results = cursor.fetchall()
                         for row in results:
                              yield dictionary.doc2bow(row[0].lower().split())
                    except:
                         print "Error: unable to fetch data"
     emotionTerms = initial
     if (len(initial) > reduceTo and not ELSA):
          print str(len(initial)) + " emotions, " + str(len(initial)-reduceTo) + " reductions remaining."
     else:
          print str(len(initial)) + " emotions, 0 reductions remaining."
     print "Creating dictionary..."
     # Create a dictionary
     dictionary = corpora.Dictionary()
     for emotion in range(len(emotionTerms)):
          try:
               sql = "SELECT text, @idx:=@idx+1 AS idx FROM ETWEETS WHERE emotion = '%s' %s LIMIT %d" % (emotionTerms[emotion], sqlHaving, limit)
               cursor.execute("SELECT @idx:=0;")
               cursor.execute(sql)
               results = cursor.fetchall()
               dictionary.add_documents(row[0].lower().split() for row in results)
          except MySQLdb.Error, e:
               print "Error %d: %s" % (e.args[0], e.args[1])
     # We don't use a stoplist
     stoplist = set("a".split())
     stop_ids = [dictionary.token2id[stopword] for stopword in stoplist
                 if stopword in dictionary.token2id]
     # We get rid of words that only occur once in the entire corpus
     once_ids = [tokenid for tokenid, docfreq in dictionary.dfs.iteritems() if docfreq == 1]
     dictionary.filter_tokens(stop_ids + once_ids)
     # Remove gaps in id sequence after words that were removed
     dictionary.compactify()
     # Print dictionary information
     print dictionary

     ###############################################################

     print "Creating corpus object..."
     # Create corpus object from database iteratively, doesn't load into memory
     corpus_memory_friendly = MyCorpus()
     print "Done."

     ###############################################################

     print "Generating LSA Space..."     
     # use a log-entropy model to weight terms
     logent = models.LogEntropyModel(corpus_memory_friendly)
     # initialize an LSI transformation
     corpus_logent = logent[corpus_memory_friendly]
     lsi = models.LsiModel(corpus_logent, id2word=dictionary, num_topics=dimension)
     # create a double wrapper over the original corpus: bow->logent->fold-in-lsi
     corpus_lsi = lsi[corpus_logent]
     print corpus_lsi
     
     print "Indexing LSA Space..."
     index = similarities.Similarity("index.e", corpus_lsi, num_features=dimension)     
     print "Done."

     # Save Emotive Brain Model (for future work)
     # We can have a new brain model each month for example
     #index.save("EB0")
     # Load Emotive Brain Model
     #index = similarities.MatrixSimilarity.load("EB0")

     ###############################################################

     print "Clustering Documents..."     
     mapEmotion = [] # index, word
     queryMatch = [] # word, max cosine (ELSA)/index (DELSAR)
     sequentialCount = 0 # We need this to keep track of streaming
     # Stream back all documents in order
     for emotion in range(len(emotionTerms)):
          try:
               sql = "SELECT text, @idx:=@idx+1 AS idx FROM ETWEETS WHERE emotion = '%s' %s LIMIT %d" % (emotionTerms[emotion], sqlHaving, limit)
               cursor.execute("SELECT @idx:=0;")
               cursor.execute(sql)
               results = cursor.fetchall()
               for row in results:
                    # For each document, a is actual term, b is most similar term using LSA
                    a = emotionTerms[emotion]
                    # Delete the emotion keyword from the document
                    tweet = row[0].replace(emotionTerms[emotion], "")
                    # Convert the document to logent LSA space
                    vec_bow = dictionary.doc2bow(tweet.lower().split())
                    query_lsi = lsi[logent[vec_bow]]
                    # Compute document similarity
                    sims = index[query_lsi]
                    sims = sorted(enumerate(sims), key=lambda item: item[0])
                    # Delete the current document from the array
                    del sims[sequentialCount]
                    if(ELSA):
                         # ELSA uses maximum cosine value
                         b = max(sims, key=lambda x: x[1])
                    else:
                         # DELSAR just wants to know what emotion is the most similar document
                         b = sims.index(max(sims, key=lambda x: x[1]))
                    queryMatch.append([a,b]) 
                    mapEmotion.append(emotionTerms[emotion]) 
                    sequentialCount += 1
          except MySQLdb.Error, e:
               print "Error %d: %s" % (e.args[0], e.args[1])      
     accuracy = []
     totalHit = 0
     totalMiss = 0
     if(printDELSAR):
          clusters = []
     if(ELSA):
          # For each emotion we get an average maximum cosine value
          floatNums = [float(vec[1][1]) for vec in queryMatch]
          average = sum(floatNums) / len(queryMatch)
          accuracy.append([emotionTerms[0], average])
          for vec in accuracy:
               numbers.append(vec[1])
               print vec
     else: 
          # DELSAR
          # This bit determines accuracy of clustering (labelling)
          # If the nearest document is the same emotion, it is a hit,
          # otherwise it is a miss. In both occassions, we record
          # the actual emotion in the clusters array if required
          # for printing to Excel
          for a in range(len(emotionTerms)):
               hitRate = 0
               hit = 0
               miss = 0
               for vec in queryMatch:
                    if vec[0] == emotionTerms[a]:
                         if vec[0] == mapEmotion[vec[1]]:
                              hit += 1
                              totalHit += 1
                              if(printDELSAR):
                                   clusters.append((vec[0],mapEmotion[vec[1]]))
                         else:
                              miss += 1
                              totalMiss += 1
                              if(printDELSAR):
                                   clusters.append((vec[0],mapEmotion[vec[1]]))
               c = float(hit)
               d = float(miss)
               if c+d > 0:
                    hitRate = c/(c+d)
               # Accuracy of hit rate for each emotion gets appended to this array
               accuracy.append([emotionTerms[a], hitRate])
          totalHit = float(totalHit)
          totalMiss = float(totalMiss)
          if((totalHit+totalMiss) > 0):
               total = (totalHit/(totalHit+totalMiss))
          else:
               total = 0
          # Reduce emotion set and rerun DELSAR if we need to reduce more emotions
          if(len(initial) > reduceTo):
               print "Removed %s (%f)" % (min(accuracy, key=lambda x: x[1])[0], min(accuracy, key=lambda x: x[1])[1])
               del initial[initial.index(min(accuracy, key=lambda x: x[1])[0])]
               DELSAR(initial, reduceTo)
          # Otherwise we're done
          else:
               for vec in accuracy:
                    print vec
               print "TOTAL ACCURACY: %f" % total  
          if(printDELSAR):
               # Print DELSAR clustering values in order for copy&paste into Excel
               c = Counter(clusters)
               for i in range(len(emotionTerms)):
                    for j in range(len(emotionTerms)):
                         for vec in c:
                              if vec[0] == emotionTerms[i]:
                                   if vec[1] == emotionTerms[j]:
                                        print c[vec]


###############################################################

# START PROGRAM HERE:

###############################################################

# Start the timer
t0 = time.clock()
if(ELSA):
     numbers = []
     for a in emotionTerms:
          print "ELSA" + str(limit) + " testing " + a + " with " + str(dimension) + " dimensions."
          print sqlHaving
          DELSAR([a], 88) # Make sure the number here is greater than the emotion set size
     for vec in numbers:
          print vec
else: # DELSAR
     print "DELSAR" + str(limit) + " testing " + emotionTest + " with " + str(dimension) + " dimensions."
     print sqlHaving
     DELSAR(emotionTerms, 88) # Change the number here to the final reduced keyword set size
# Close database and print total time taken
db.close()
print "Time Taken:"
print str((time.clock() - t0)/60)

# # # # # # # # -------- END PYTHON -------- # # # # # # # #

\end{lstlisting}
\end{landscape}

\chapter{Appended Figures}\label{more-figures}

\begin{landscape}
\begin{figure*}
\centering
\includegraphics[width=555px]{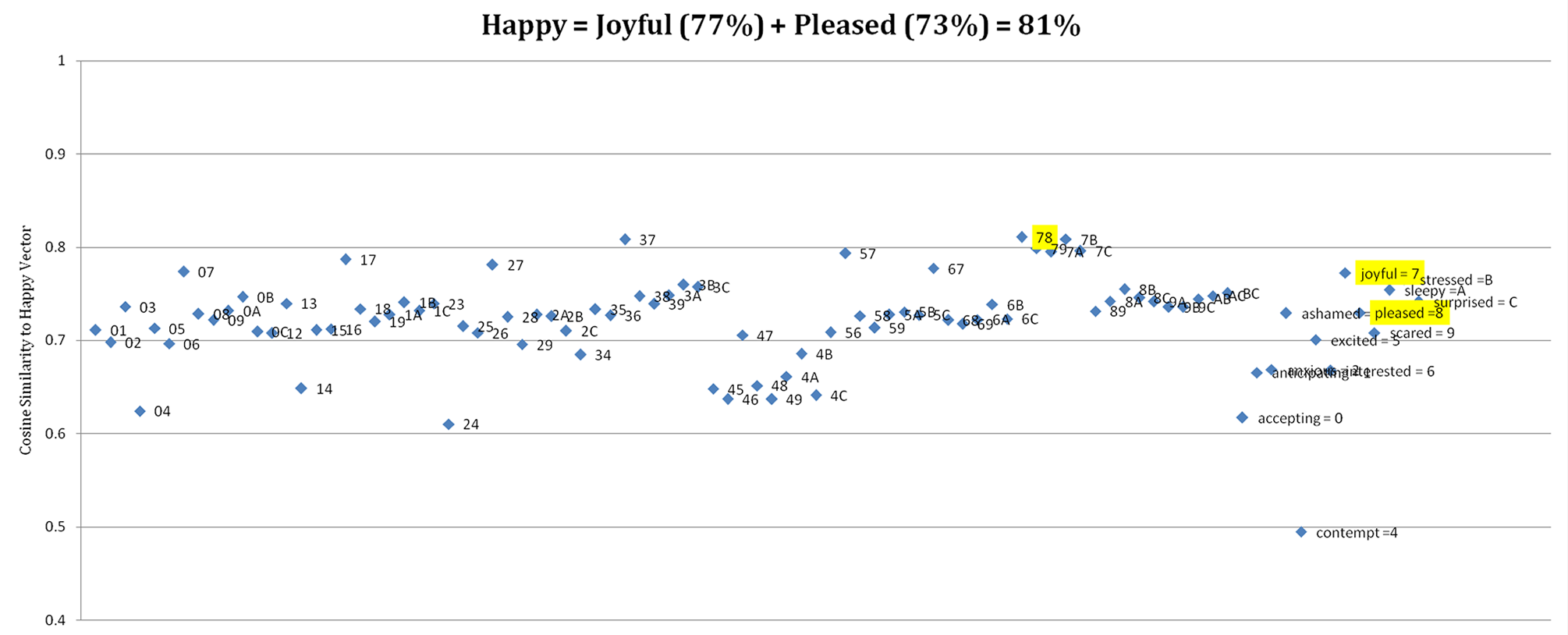}
\caption[Similarity of primary emotion combinations to \textit{happy}]{Cosine similarity of primary emotion combinations to \textit{happy}.}
\label{happy}
\end{figure*}
\end{landscape}

\newpage

\newpage

\end{document}